\PassOptionsToPackage{table}{xcolor}
\RequirePackage{fix-cm}
\documentclass[11pt]{sjtu_sai}
\usepackage{amsmath,amssymb,mathtools}
\usepackage{tabularx,array,xspace}
\usepackage{tikz}
\usetikzlibrary{arrows.meta,positioning,calc,fit}
\usepackage{needspace}
\usepackage{flafter}
\usepackage{float}
\usepackage{wrapfig}
\usepackage{xurl}
\usepackage{fontawesome5}
\newcommand{\system}{A\textsuperscript{2}-MEAS\xspace}
\newcommand{\wla}{A\textsuperscript{2}-WLA\xspace}
\DeclareRobustCommand{\metabrain}{\textnormal{MetaBrain}\xspace}
\newcommand{\sg}{\operatorname{sg}}

\newcommand{\topicmarker}{\textcolor{saiblue}{\ensuremath{\blacktriangleright}}}
\definecolor{RevisionBlue}{HTML}{0057B8}

\newenvironment{revision}{%
  \begingroup
  \hypersetup{linkcolor=saiblue,citecolor=black,urlcolor=black}%
}{\par\endgroup}
\AtBeginEnvironment{table}{\hypersetup{linkcolor=black,citecolor=black,urlcolor=black}}
\newcolumntype{Y}{>{\raggedright\arraybackslash}X}
\definecolor{Ablue}{HTML}{234E70}
\definecolor{Ateal}{HTML}{13765C}
\colorlet{OurMethodBg}{saiblue!15}
\definecolor{Aorange}{HTML}{B5672B}
\definecolor{Apale}{HTML}{F1F5F7}
\definecolor{Arule}{HTML}{CAD4DA}
\renewcommand{\topfraction}{0.9}
\renewcommand{\textfraction}{0.08}
\renewcommand{\floatpagefraction}{0.75}
\setcitestyle{numbers,square,comma}
\usepackage{xcolor}
\usepackage{booktabs,tabularx,multirow,array}

\definecolor{MetaBg}{HTML}{EDF3FA}
\definecolor{DiagBg}{HTML}{EDF5F1}
\definecolor{ActionBg}{HTML}{FAF4E9}

\usepackage{tcolorbox}
\tcbuselibrary{listings,skins}

\definecolor{contextbg}{HTML}{EDF7F4}
\definecolor{contextborder}{HTML}{B8DAD0}

\DeclareRobustCommand{\namedref}[2]{\hyperref[#2]{#1~\ref*{#2}}}
\DeclareRobustCommand{\equationref}[1]{\hyperref[#1]{Equation~(\ref*{#1})}}
\hypersetup{pdftitle={Toward the Era of Agentic Embodied Intelligence},pdfauthor={}}
\title{Toward the Era of Agentic Embodied Intelligence}
\makeatletter
\renewcommand{\@toptitlebar}{%
  {\color{saiblue}\hrule height 0.5pt}%
  \nointerlineskip
  \hbox to \textwidth{%
    \smash{\raisebox{3.5pt}{%
      \includegraphics[height=25pt,trim=6bp 4bp 4bp 6bp,clip]{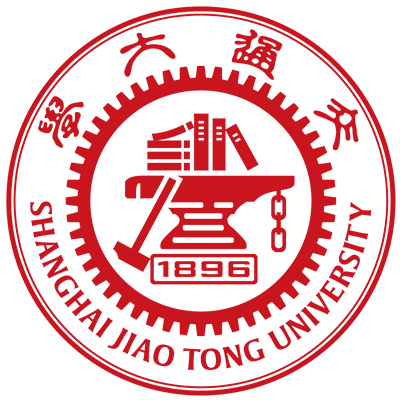}%
      \hspace{6pt}%
      \includegraphics[height=25pt,trim=110bp 149bp 93bp 91bp,clip]{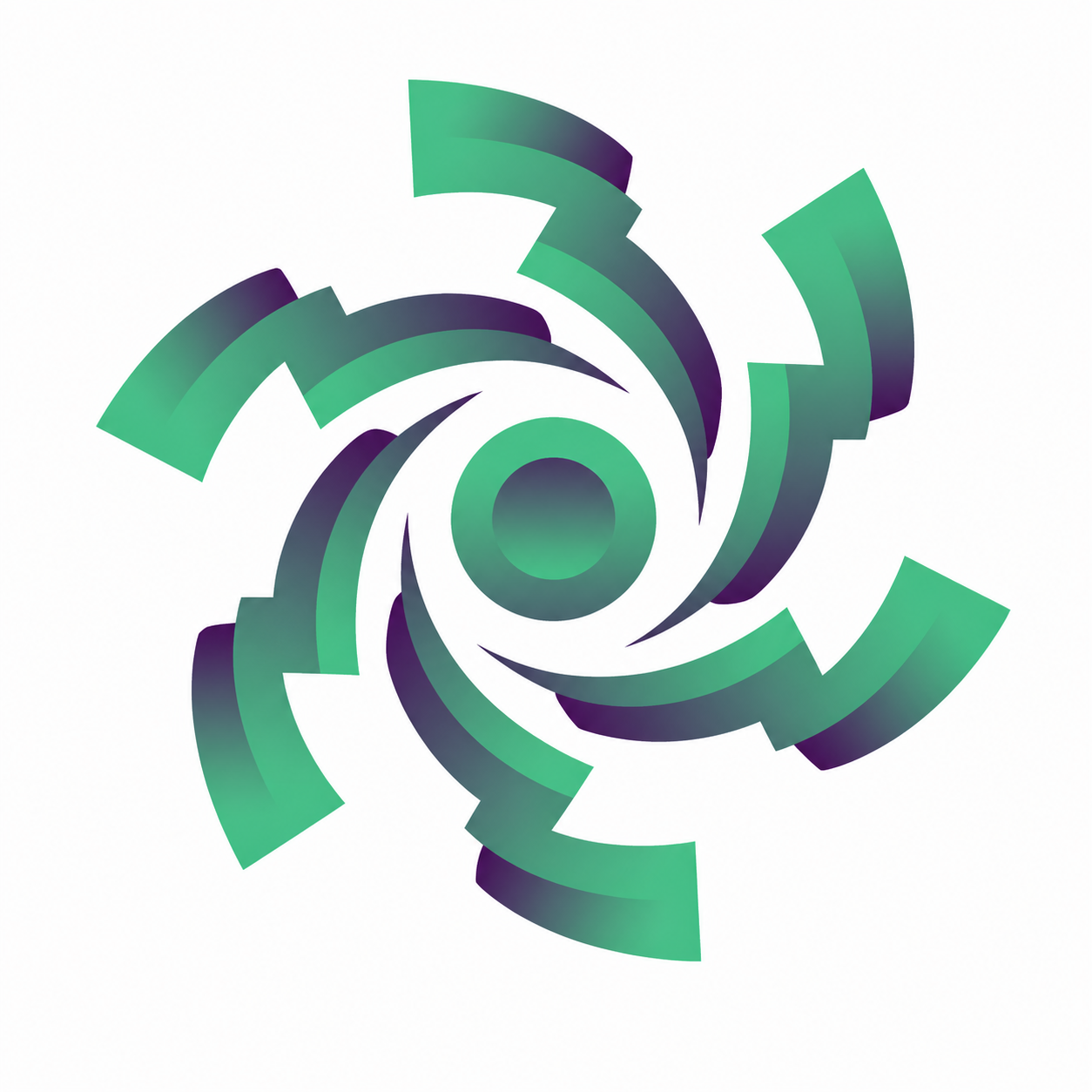}%
      \hspace{6pt}%
      \includegraphics[height=25pt,trim=19bp 17bp 1085bp 26bp,clip]{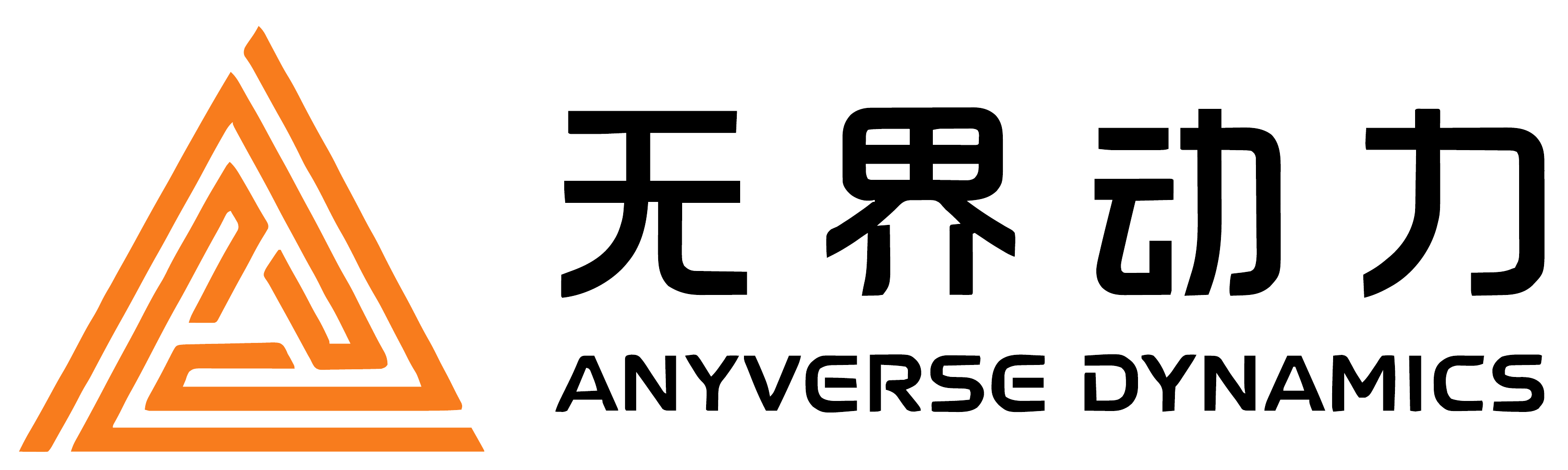}%
    }}\hfil
  }%
  \nointerlineskip
  \vskip 6mm
}
\makeatother

\renewcommand{\affiliationlist}{{\affiliationfont\rmfamily
AutoLab, School of Artificial Intelligence, Shanghai Jiao Tong University\\[3pt]
Research Lab, Anyverse Dynamics}}
\abstract{
Frontier foundation models can increasingly guide robot actions, but sustained physical autonomy also requires efficient control and recovery from failures. We introduce \system, an embodied multi-agent framework built around \emph{code as a scaling interface}. Here, a correction code is a learned representation of corrective language, grounded spatial evidence, execution history, and an action proposal. \metabrain supplies corrective supervision during training and grounded guidance during execution. The correction encoder and adapter learn to turn this guidance into actions, absorbing corrective knowledge into trainable action-policy weights. At deployment, fresh codes adjust behavior while all weights remain fixed. This couples learning from foundation models with using their reasoning when new failures arise. The underlying action model, \wla, augments a vision-language-action policy with motion-aware history and a latent world model. Surprisingly, even before online correction, this simple architecture with action pretraining and task specialization reaches 0.5755 Q-score on the 50-task BEHAVIOR 2026 public subset, approximately $2.21\times$ the 2025 winner's reported public score of 0.2605 across different benchmark editions and training configurations. On the full 100-task BEHAVIOR 2026 public set, the complete system combines \metabrain, diagnostic agents, and learned correction to achieve 0.6411 Q-score and 45.40\% complete-task success. The same action design also supports short-horizon bimanual manipulation, achieving 93.98\% average success on RoboTwin 2.0, or 95.05\% with the complete system. Real-robot evaluation spans 12 tabletop tasks across five household settings, where \wla achieves 57.92\% average success compared with 26.67\% for $\pi_{0.5}$. Long-horizon pour-over coffee preparation tests recovery during sustained execution. At two filter-handling steps, a history-conditioned Adapter raises post-failure completion from 14.00\% to 94.00\% and from 8.00\% to 82.00\%, relative to retries by \wla without the Adapter. To support development at this scale, our training infrastructure reaches 427.1 samples per second, approximately $4.03\times$ the throughput of a $\pi_{0.5}$ reference implementation on matched hardware. We will release the system, models, training scripts, and infrastructure as an efficient, usable \mbox{framework for embodied learning.}

}
\renewcommand{\checkdatalist}{%
  {\small\begin{tabular}{@{}c@{\hspace{6pt}}l@{\hspace{10pt}}l@{}}
    \raisebox{-1pt}{\fontsize{11}{11}\selectfont\faGithub} &
    {\checkdatafont\sffamily\bfseries Code and releases} &
    \url{https://github.com/AutoLab-SAI-SJTU/A2MEAS} \\[3pt]
    \raisebox{-3pt}{\includegraphics[height=15pt]{logo/autolab_logo.png}} &
    {\checkdatafont\sffamily\bfseries Project page} &
    \url{https://a2-meas.github.io/}
  \end{tabular}}%
}
\begin{document}
\maketitle
Recent evaluations show that general-purpose models can generate robot actions without task-specific parameter updates. On the 42-task RoboDojo evaluation, GPT-6 Astra ranks above the 40 public policies included in the comparison, although precision and dynamic control remain limitations~\cite{gpt6robot}. Specialized action models can benefit from this progress by absorbing foundation-model knowledge into their weights and consulting these models during execution. Learning recurring decisions into a compact policy offers efficient control, while access to a capable foundation model can help address unfamiliar failures. The challenge is to combine these routes to sustain coherent behavior throughout an activity.

A robot must do more than choose a plausible next action. Each interaction changes the conditions for what follows. A recovered grasp may secure an object in an orientation that prevents its later placement, and a successful local correction may leave the original goal unfinished. Long-horizon autonomy therefore requires reasoning about both the immediate physical state and the consequences of changing it. This motivates our focus on how learned control, task reasoning, and diagnostic evidence cooperate across complete activities, including failures and the attempts to recover from them.

Recovery exposes the need to connect task reasoning directly to action generation. Language-guided skill selection and feedback-driven replanning help a robot decide what to attempt next~\cite{saycan,innermonologue,reflect}. After a failed grasp, however, choosing to retry does not determine how the next attempt should differ. The controller must adjust its approach, target alignment, or gripper timing using evidence from the failed interaction. Language-conditioned robot policies and vision-language-action (VLA) models learn to map observations and instructions to control~\cite{rt1,rt2,openvla,pi0,pi05,grootn1}. Predictive approaches, from GR-1 to world-action models (WAMs), also connect action generation with future visual dynamics~\cite{gr1,dreamzero,lawam}. The interface must make newly inferred corrections usable within this control process. Spatial planning, joint reasoning-and-action models, and learned recovery offer ways to connect decisions to execution~\cite{voxposer,g05,flare}. We focus on a learned interface that carries corrective intent together with its spatial and temporal evidence into the action policy, so that a diagnosis can change the next trajectory while preserving the task goal.

We introduce \system, an embodied multi-agent framework that coordinates reasoning, diagnostic grounding, and continuous control through a shared execution record. Its action model, \wla, learns from long-horizon demonstrations and uses recent visual history and predicted scene features to generate actions. A vision-language coordinator, \metabrain, maintains the task goal and examines unexpected outcomes. Diagnostic agents provide perceptual, spatiotemporal, and interaction evidence, using object localization, segmentation, tracking, available depth, and robot-state signals. These are complementary diagnostic functions rather than a fixed list of mandatory perception models. \metabrain combines their findings with executed action history to formulate a correction. The action policy then attempts the correction, and the observed result becomes evidence for the next decision.

Our central principle is \emph{code as a scaling interface}. The code is a learned conditioning representation that brings together a recovery instruction, grounded spatial evidence, executed history, and a fresh action proposal. Language expresses the intended change, while the other inputs specify the physical situation in which it must be made. A residual adapter uses the resulting code to modify the action expert's hidden states throughout trajectory generation. This gives heterogeneous agents a common route for influencing continuous control without requiring them to share an architecture. During training, \metabrain-generated instructions and expert continuations teach the correction encoder and adapter how corrective reasoning should change actions. This absorbs corrective knowledge into trainable action-policy parameters while leaving the \wla backbone frozen. During deployment, new guidance changes the code and the generated behavior, with all parameters fixed. The framework therefore links knowledge learned into the policy with reasoning supplied when execution needs it.

A strong action model is the starting point for this cooperation. \wla augments a vision-language-action policy with motion-aware history and a latent world model that predicts future visual features. Surprisingly, this simple architecture, together with action pretraining and task specialization, already yields strong long-horizon performance before adding online correction. On the 50-task public subset of BEHAVIOR 2026, one shared \wla model reaches 0.4522 Q-score, and task-specialized policies reach 0.5755. The latter is approximately $2.21\times$ RLC's public score of 0.2605 in the 2025 competition~\cite{behavior2025board}. RLC, the 2025 winner, combines learned control with explicit stage-transition logic, a gripper-reopening heuristic, and a radio-specific rollback rule~\cite{rlc2025}. The strong result from our comparatively simple action architecture highlights how much competence can be learned within the policy before adding agent-guided intervention. 


We then ask what grounded intervention adds to that competence. Across all 100 BEHAVIOR 2026 tasks, the multi-model \wla baseline obtains 0.6040 Q-score and 40.60\% complete-task success. Adding \metabrain, diagnostic evidence and learned correction raises these public-set results to 0.6411 and 45.40\%, with at least one successful episode on 88 tasks. Fixed-policy Adapter ablations and a disjoint set of task instances test the correction mechanism separately from model selection.


We also test the action model beyond long-horizon whole-body control. On RoboTwin 2.0, \wla achieves 93.98\% average success across Clean and Randomized conditions, and the complete system reaches 95.05\%. Real-robot experiments span 12 tabletop tasks across five household settings, with 20 trials per task and configuration. Without high-level correction, the full \wla recipe reaches 57.92\% macro success, compared with 26.67\% for $\pi_{0.5}$~\cite{pi05}. These tasks cover storage, retrieval, food preparation, pushing and tool use. Long-horizon pour-over coffee preparation takes approximately four minutes per run. At two filter-handling steps, a lightweight history-conditioned Adapter completes 94/100 and 82/100 post-failure trials, compared with 14/100 and 8/100 for retries by the same frozen \wla policy with the Adapter disabled. These real-robot recovery tests complement the full correction-interface experiments in simulation.

Long-horizon experiments also make iteration expensive. We therefore develop training and evaluation infrastructure alongside the models. Our implementation processes 427.1 demonstration samples per second on eight GPUs, approximately $4.03\times$ the throughput of a $\pi_{0.5}$ reference implementation on the same hardware. The reference uses matching image resolution and action horizon, with each implementation's batch size chosen for resource utilization. For evaluation, ideal linear scaling of the estimated 1,275-hour serial runtime gives a naive eight-GPU baseline of approximately 159.4 hours. Our eight-Pro6000 simulation and policy worker pool completes the 1,000-episode suite in 18.1 hours, approximately $8.8\times$ faster than this projected baseline, with the reasoning service hosted separately. 


We will progressively release the system code, models, training scripts and optimized infrastructure at \url{https://github.com/AutoLab-SAI-SJTU/A2MEAS}.  The release is intended to give researchers and developers an efficient, usable framework for training, evaluating and deploying embodied agents across large-scale simulation \mbox{platforms and real-world tasks.}

Our contributions are threefold. \(\spadesuit\) We introduce \system and its code interface, which learns from \metabrain-generated corrective supervision and connects online reasoning and grounded evidence to continuous control with deployment-time parameters fixed. \(\spadesuit\) We establish the performance of the simple \wla architecture across long-horizon BEHAVIOR and short-horizon RoboTwin evaluations, and test physical manipulation across diverse tabletop tasks and long-horizon coffee preparation. The complete system reaches 0.6411 Q-score and 45.40\% success on the BEHAVIOR 2026 public set. \(\spadesuit\) We provide infrastructure with a $4.03\times$ training-throughput gain under the stated hardware conditions and an evaluation speedup of approximately $8.8\times$ relative to the estimated naive eight-GPU baseline, making full-suite testing practical \mbox{within a daily research cycle.}

\FloatBarrier
\Needspace{8\baselineskip}
\section{The \texorpdfstring{A\textsuperscript{2}-MEAS}{A2-MEAS} framework}
\label{sec:system}

General embodied intelligence requires a system to connect its intentions to the physical consequences of its actions. Over a long activity, a robot must remember what remains to be done while responding to changes in objects, contacts, and its own configuration. A successful movement is useful only if the resulting state supports continued progress. \system addresses this problem by coupling learned action generation with task-level reasoning and evidence-based diagnosis.

The framework contains three cooperating groups of agents (\namedref{Figure}{fig:system}). \wla generates continuous actions from the task instruction and observed history. \metabrain monitors execution in relation to the full goal and determines when the next attempt should change. Diagnostic agents investigate the physical questions needed to make that decision. All three use a shared record of actions and observations. When execution requires intervention, a learned correction interface converts their instruction and evidence into changes inside \mbox{the action generator.}

\namedref{Section}{sec:coordination} explains how the shared record coordinates these agents. \namedref{Section}{sec:formulation} introduces the action model, its observation and prediction components, and the adapter that accepts corrective guidance. \namedref{Section}{sec:framework-recovery} follows a recovery example from diagnosis through corrected execution. \namedref{Sections}{sec:correction} and~\ref{sec:wla} then develop the interface and action model in detail.

\begin{figure}[htbp]
\centering
\includegraphics[width=\linewidth]{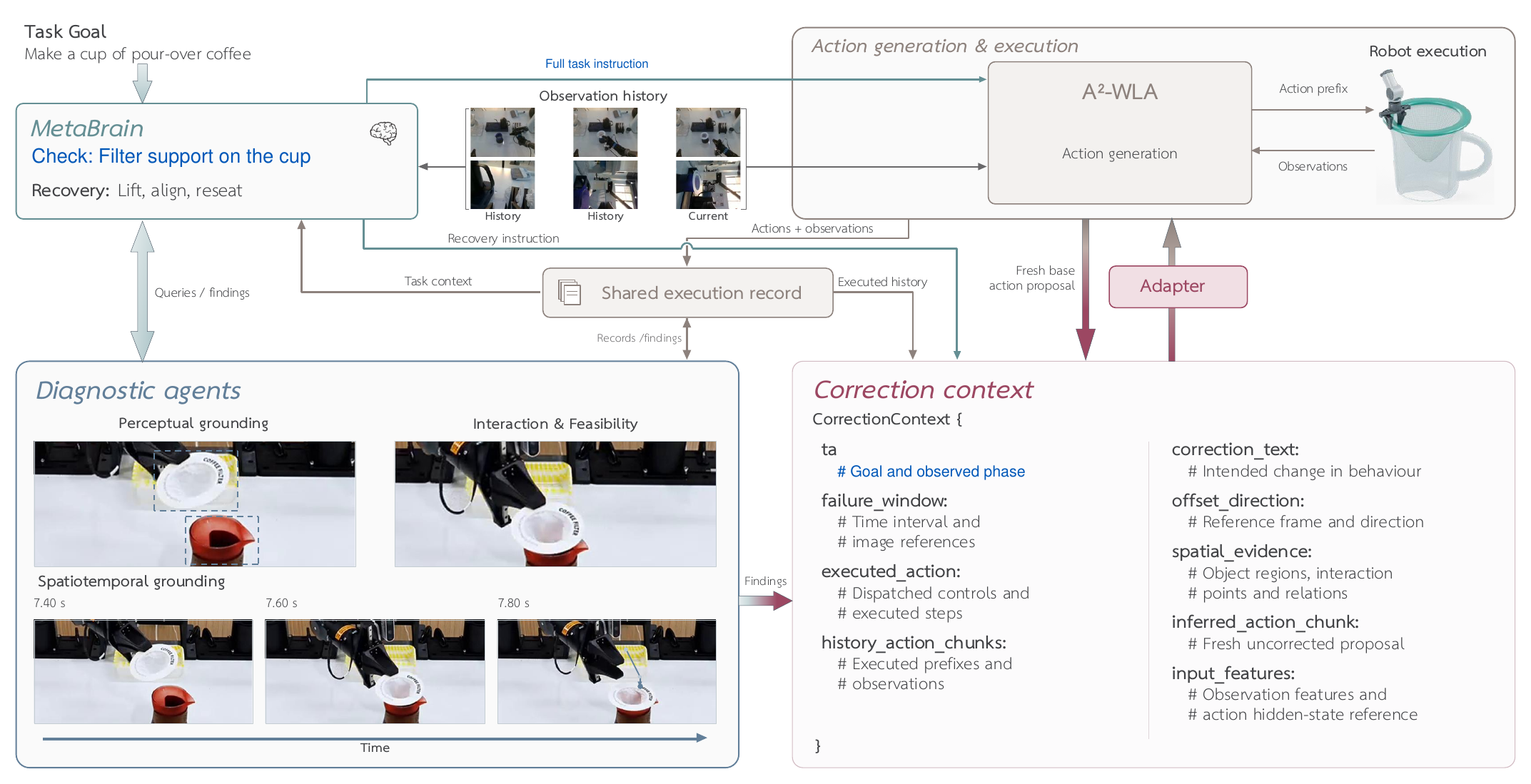}
\caption{The \system framework. The \metabrain coordinates task reasoning and recovery. Grounding agents provide specialized evidence. The correction interface combines recovery language, spatial evidence, executed history, and a fresh base-action proposal to modulate \wla. These agents cooperate to control one robot.}
\label{fig:system}
\end{figure}

\FloatBarrier
\Needspace{6\baselineskip}
\subsection{Framework architecture and coordination}
\label{sec:coordination}

\begin{table}[!t]
\centering
\small
\renewcommand{\arraystretch}{1.18}
\setlength{\tabcolsep}{6pt}

\caption{The three groups of agents in \system. Rows within each
group describe its responsibilities or internal components.
Diagnostic agents can share tools across functions. A common
execution layer dispatches controls, checks outcomes and records
the history used by all three groups.}
\label{tab:roles}

\begin{tabularx}{\linewidth}{
  @{}
  >{\raggedright\arraybackslash}p{2.3cm}
  >{\raggedright\arraybackslash}p{4.0cm}
  Y
  @{}
}
\toprule
\textbf{Agents}
  & \textbf{Responsibility}
  & \textbf{Function} \\
\midrule

\rowcolor{MetaBg}
  & Task verification
  & Maintain the task goal and verify execution progress \\
\rowcolor{MetaBg}
  & Failure diagnosis
  & Review attempted actions and request missing evidence \\
\rowcolor{MetaBg}
\multirow{-3}{2.3cm}{\metabrain}
  & Recovery instruction
  & Describe how the next attempt should change \\
\midrule

\rowcolor{DiagBg}
  & Perceptual grounding
  & Identify objects, regions and visible properties \\
\rowcolor{DiagBg}
  & Spatiotemporal grounding
  & Track identities, relative positions and changes over time \\
\rowcolor{DiagBg}
\multirow{-3}{2.3cm}{Diagnostic agents}
  & Interaction and feasibility
  & Assess contact, support and movement constraints \\
\midrule

\rowcolor{ActionBg}
  & Observation encoding
  & Represent the current scene and recent motion \\
\rowcolor{ActionBg}
  & World prediction
  & Predict scene features relevant to the next movement \\
\rowcolor{ActionBg}
  & Action generation
  & Generate continuous controls and estimate local progress \\
\rowcolor{ActionBg}
\multirow{-4}{2.3cm}{\wla}
  & Correction interface
  & Encode recovery context and adapt action generation \\
\bottomrule
\end{tabularx}
\end{table}

\system couples a local control loop with a task-level reasoning loop. The local loop repeatedly generates action chunks from the task instruction and recent observations. The task-level loop uses the resulting evidence to decide whether execution can continue, whether the goal has been achieved, or whether recovery is needed. A shared execution record stores the goal, dispatched controls, and observed outcomes. It gives all agents a common account of what has actually happened without requiring them to share a backbone or run at \mbox{the same frequency.}

The execution layer dispatches only a prefix of each predicted chunk, then obtains fresh observations before the next policy call. Thus, control is closed-loop across chunks even though an uninterrupted prefix executes without a new policy prediction. \wla handles ordinary changes through this repeated observation--action cycle. When monitoring detects an abnormal outcome, the runtime pauses pending controls and asks \metabrain to review the evidence. Verified recovery returns execution to the nominal loop, and verified completion ends the episode.

Diagnostic agents are invoked when a decision depends on an unresolved physical condition. \namedref{Table}{tab:roles} groups their functions into perceptual grounding, spatiotemporal grounding, and interaction assessment. Perception locates objects and regions. Temporal analysis tracks how they move relative to one another. Interaction assessment combines those observations with available robot-state, contact, and geometric evidence. For example, seeing an object between the gripper fingers does not establish a secure grasp. Tracking whether it moves with the hand, together with gripper-state evidence, supports a more informative assessment. Depth or reconstructed geometry can supply additional evidence when available. Each finding retains its observation source and timestamp so that \metabrain can distinguish a supported conclusion from a missing observation.


The result determines what happens next. Evidence that supports the required condition permits execution to continue. Evidence of a mismatch prompts a correction. Insufficient evidence prompts another observation within the sensing budget. A negative answer and an unknown answer therefore lead to different decisions. Conditional tool invocation keeps ordinary control inexpensive while allowing several diagnostic capabilities to contribute when a correction needs them.


\subsection{Action model and correction interface}
\label{sec:formulation}

Let $g$ denote the complete task instruction, retained throughout nominal execution. We index the physical control step at which the policy executes by $t$. The policy input $x_t$ aggregates synchronized multiview visual observations, a finite temporal window of past visual frames, and the current proprioceptive state $s_t$ comprising joint positions and end-effector configurations.

\wla combines three sources of action context. A vision-language backbone encodes the task instruction and visual observations. Its motion-aware history encoder represents changes across recent frames. A latent world model predicts future visual features from a transition representation inferred by the backbone. The action expert uses current context, predicted future context, and proprioception to generate controls. Future observations supervise training but are unavailable at deployment. \namedref{Section}{sec:wla} describes these components and the three stages used to train the base policy.

During standard nominal operation, \wla executes with the correction branch deactivated. We designate this default configuration as the base policy $\pi_\theta$, parameterized by $\theta$. At invocation step $t$, with sampling randomness suppressed in the notation, it jointly predicts a multi-step continuous action chunk and an auxiliary local-phase progress estimate
\begin{equation}
  (\hat A_t,\hat p_t)=\pi_\theta(x_t,g),
 \qquad \hat A_t\in\mathbb R^{H\times d_a},\quad
 \hat p_t\in[0,1]^H,
 \label{eq:policy}
\end{equation}
where $H$ is the prediction horizon and $d_a$ is the number of values in each robot control command. Row $\hat a_{t,j}\in\mathbb R^{d_a}$ proposes the control for step $t+j$, with $j\in\{0,\ldots,H-1\}$. The corresponding entry $\hat p_{t,j}$ estimates relative progress within the annotated local-phase interval. Its training target follows position within a demonstrated skill phase, from values near zero at the beginning to one at the final annotated step. This estimate supports execution monitoring, while observed physical outcomes determine whether to continue execution, request correction, or confirm task completion.

The correction branch adds a second way to influence action generation. Its condition encoder combines a recovery instruction, verified spatial evidence, relevant executed history, and an uncorrected proposal from the base policy. The resulting code conditions a residual adapter inside the action expert. The code is computed once per generated chunk, while the adapter recomputes its hidden-state adjustment at each flow-integration step. Later action layers turn the modified representation into a corrected trajectory. This mechanism supplements the policy's ordinary visual and predictive context without \mbox{updating its weights (\namedref{Section}{sec:correction}).}

The robot executes only an initial prefix of the generated chunk. Let $L_t$ be the number of controls actually dispatched at call $t$, with $1\leq L_t\leq H_{\mathrm{exec}}\leq H$ whenever execution starts. The BEHAVIOR reference configuration predicts $H=50$ controls, executes at most $H_{\mathrm{exec}}=32$, and uses $d_a=23$ control dimensions. After an uninterrupted prefix, the next observation is acquired at $t+32$, and a new prediction replaces the unused 18 controls. If monitoring interrupts the prefix, the next observation corresponds to $t+L_t$. A candidate rejected before dispatch produces no executed controls and is not logged as an action prefix.

The execution record $R_t$ stores the history available at physical step $t$. Each entry links observations before and after a dispatched prefix to its actual controls, executed length, and timestamps. Camera identifiers, coordinate frames, task identifiers, and available local-phase annotations preserve the meaning of those records. Unexecuted proposals are stored separately. The short visual window in $x_t$ serves immediate action prediction, whereas $R_t$ supplies a longer history for diagnosis and recovery.

\begin{figure}[t!]
\centering
\includegraphics[width=\linewidth]{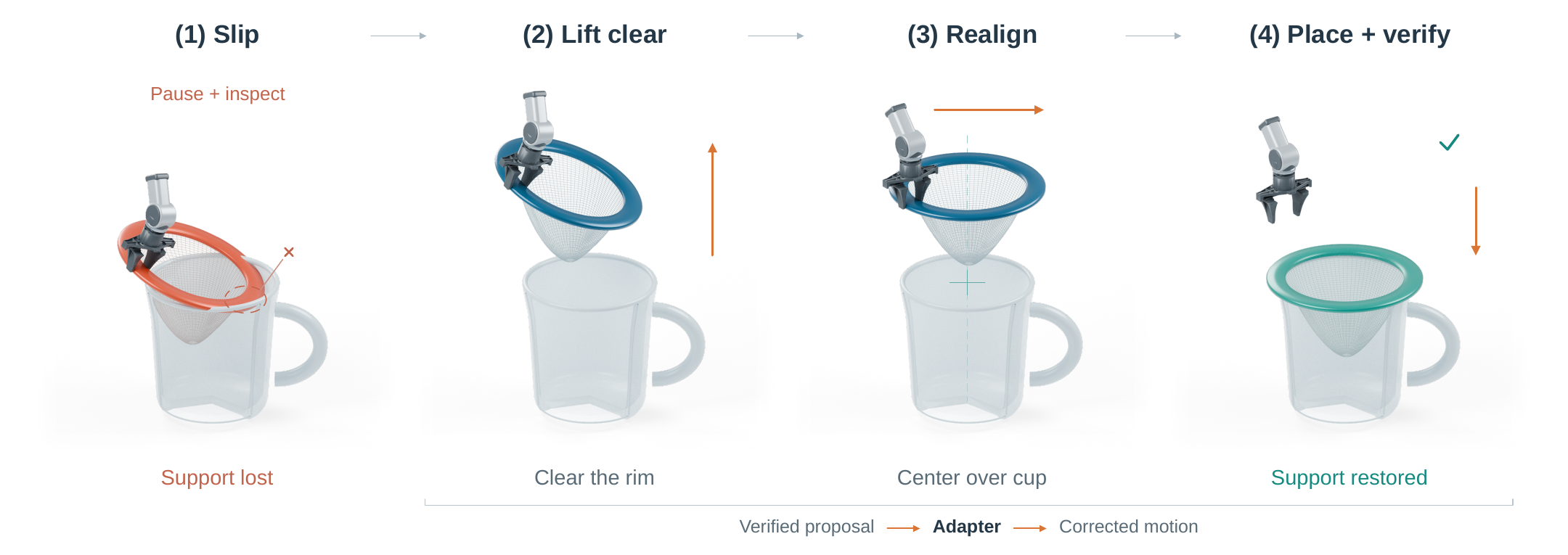}
\caption{State progression during the closed-loop recovery of a misaligned filter placement. Following a loss of contact support along one edge of the support ring, the execution layer pauses to verify spatial relationships and synthesizes a sequence of lifting, realigning, and reseating actions conditioned on grounded high-level guidance. Faded outlines trace the initial misaligned pose, orange arrows indicate the directional motion flow, vertical dashed lines mark the central axis of the cup opening, and green indicators highlight the restored planar support.}
\label{fig:filter-correction}
\end{figure}

After a dispatched prefix finishes or is interrupted at step $t+L_t$, the runtime returns a feedback bundle $b_{t+L_t}$ containing the updated execution records, diagnostic findings, progress estimates, and status. The status is \texttt{RUNNING} during nominal execution, \texttt{TASK\_COMPLETE} after completion checks succeed, or \texttt{ABNORMAL} when execution requires review. Progress predictions describe the proposed chunk and are interpreted together with the executed controls and observed outcomes. \metabrain uses this feedback to decide what evidence or corrective instruction is needed next.

\subsection{From failure diagnosis to corrected action}
\label{sec:framework-recovery}

The correction interface becomes useful when observing the changed scene is not enough to change an ineffective behavior. \namedref{Figure}{fig:filter-correction} illustrates a filter-placement example. The support ring should rest on the cup rim, with the filter hanging inside the opening. If one edge slips below the rim, continuing downward may wedge the ring or disturb the cup. Recovery instead calls for lifting the ring, aligning it above the opening, and lowering it again.

After detecting an anomaly, the framework pauses pending controls and establishes a recovery state. \metabrain examines the observations and executed controls between the last verified nominal state and the detected deviation. It asks the diagnostic agents to assess contact stability and the spatial conditions required for another placement attempt. Perceptual tools identify the filter and cup regions, while temporal tracking helps bound the interval in which support was lost. Available robot-state and contact evidence informs the recovery assessment. Geometric and kinematic checks assess whether the proposed lifting motion is feasible. The recovery instruction is finalized once its required premises have been resolved. If occlusion leaves a necessary condition unknown, the system requests another observation within its sensing budget. The instruction specifies the adjustment and reference frame, while \wla generates \mbox{the continuous trajectory.}

At the recovery state, the base policy first generates an uncorrected proposal. This proposal shows what the policy would attempt without intervention. The condition encoder combines it with the instruction, spatial evidence, and executed history. In the filter example, ``realign the filter'' is therefore interpreted in relation to a particular ring orientation, cup opening, and attempted movement. The adapter can then apply different residuals to different action positions, allowing a chunk to lift first and approach again later. These are learned adjustments rather than a prescribed geometric trajectory.

Before dispatch, the runtime checks that the evidence remains current and that the candidate satisfies the available joint-limit and obstacle-clearance checks. It executes a prefix, records the outcome, and checks whether the support condition has been restored. Successful recovery returns the robot to the unfinished task. Otherwise, the new observations and failed attempt inform another recovery pass. The reference configuration allows at most three physical retries and 60 seconds per anomaly, with separate sensing and reasoning limits specified in \namedref{Section}{sec:recovery-protocol}.

The methodological contribution is \emph{code as a scaling interface}, a learned connection between grounded recovery context and the action policy's internal representation. \metabrain can express what should change, diagnostic tools can establish where and under what conditions, and the action model can translate that context into movement. During training, the condition encoder and adapter learn from \metabrain-generated guidance, offline-mined candidate deviations, and expert continuations while the original \wla checkpoint remains frozen (\namedref{Section}{sec:correction-training}). At deployment, fresh guidance and evidence change the correction code, which shapes action generation with all model parameters fixed. The interface thus connects corrective knowledge learned during training with the physical context encountered during execution.

\FloatBarrier
\Needspace{8\baselineskip}
\section{Cooperation through grounded correction codes}
\label{sec:correction}

A diagnosis is useful to a robot only if it changes what the robot does next. Our organizing principle, \emph{code as a scaling interface}, makes the connection through a learned representation that carries reasoning and diagnostic evidence into action generation. \metabrain describes the intended adjustment, diagnostic agents ground it in the observed scene, and the execution record identifies what has already been tried. A condition encoder combines this context with the current action proposal into a correction code. The adapter uses the code inside \wla to shape the next movement. Language is one input to this interface, together with the spatial and temporal evidence needed to make the instruction actionable. Different agents can contribute through the same context schema while retaining their own architectures.

\subsection{\metabrain diagnosis and recovery instructions}
\label{sec:metabrain-diagnosis}

\metabrain formulates a recovery instruction $u_t$ from the task goal $g$ and feedback $b_t$, including relevant executed history and diagnostic findings. The public-set system uses Qwen3.8-27B~\cite{qwen38model} for this role, separately from the 2B vision-language backbone inside \wla. It compares the observed outcome with the intended operation and requests further evidence when a condition is unresolved. The instruction describes how the next attempt should change while preserving the original goal. The memory-limited hidden-set deployment uses a smaller coordinator, as described in \namedref{Section}{sec:public-results}.

A visible failure may occur several controls after the action that caused it. An unstable grasp, for example, may become apparent only when the object slips during transport. \metabrain therefore examines the interval from the last verified nominal state to the detected anomaly, including the controls actually executed. This history helps distinguish errors during approach, grasp closure, and transport. When $u_t$ requests a spatial adjustment, it also identifies the reference frame, such as the camera, robot base, or object. \metabrain specifies the intended change, while \wla determines the continuous controls.

\subsection{Diagnostic agents and evidence verification}
\label{sec:grounding}

Before a recovery instruction is executed, the framework checks the physical conditions on which it depends. The diagnostic functions summarized in \namedref{Table}{tab:roles} contribute complementary evidence. For grasp assessment, perception identifies the target, temporal tracking measures its motion relative to the hand, and available proprioceptive signals describe the gripper state. Together, these observations help assess whether the object moves consistently with the gripper. If the evidence is insufficient to answer the query, \mbox{the result remains unknown.}

\begin{table}[t!]
\centering
\caption{Structure of the correction context provided to the model
to guide recovery. It combines the task objective, failure evidence,
executed action history, and corrective guidance.
Spatial evidence includes camera, frame, and coordinate references.
$\hat A_t^{\mathrm{base}}$ denotes a fresh, uncorrected action proposal
generated from the current recovery state.}
\label{tab:context}

\begin{tcblisting}{
  enhanced,
  listing only,
  colback=contextbg,
  colframe=contextborder,
  boxrule=0.5pt,
  arc=3mm,
  outer arc=3mm,
  boxsep=0pt,
  left=10pt,
  right=10pt,
  top=9pt,
  bottom=9pt,
  width=\linewidth,
  listing options={
    basicstyle=\ttfamily\footnotesize\color{black},
    columns=fullflexible,
    keepspaces=true,
    showstringspaces=false,
    breaklines=true,
    breakatwhitespace=false,
    breakindent=1.5em,
    aboveskip=0pt,
    belowskip=0pt
  }
}
CorrectionContext {
  task: {
    goal, observed_phase
  },
  failure_window: {
    start_time, end_time, image_refs
  },
  executed_action: {
    chunk_id, executed_steps, actual_controls
  },
  history_action_chunks: [
    {executed_prefix, observation_before, observation_after}
  ],
  correction_text: "Intended change in behaviour",
  offset_direction: {
    reference_frame, direction
  },
  spatial_evidence: {
    camera_id, frame_id, coordinate_frame,
    image_refs, object_regions, interaction_points,
    spatial_relations
  },
  inferred_action_chunk: A_base,
  input_features: {
    observation_features, action_hidden_states_ref
  }
}
\end{tcblisting}

\end{table}


A request planner routes each question to tools that can supply the necessary evidence. DINO-based features support visual matching, while SAM-based segmentation supplies object masks~\cite{dinov2,sam}. Tracking, depth or 3D estimation, robot-state readings, and analytical kinematics provide additional evidence when available. These tools support the diagnostic functions in \namedref{Table}{tab:roles}. They are selected according to the query and sensor configuration, and their outputs retain the source image, timestamp, and coordinate frame.

The interface supports six kinds of questions. \textbf{(i)} State and attribute queries assess visible properties. \textbf{(ii)} Spatial queries assess relative positions and region membership. \textbf{(iii)} Interaction queries assess grasp, contact, or support. \textbf{(iv)} Temporal queries identify changes such as release or slippage. \textbf{(v)} Instance queries track identity and detect objects that have appeared or disappeared. \textbf{(vi)} Feasibility queries assess reachability and clearance under the available geometry and robot model. Multiple query types can use the same tool, and a query can use multiple tools.

Verification distinguishes a negative answer from insufficient evidence. An observed obstacle calls for a different path, while an occluded region may require another view. All conditions required by an instruction must be supported before that instruction is dispatched. Geometric evidence must also be internally consistent. A point and mask must refer to the same view and frame, or be transformed into a common coordinate system, and the selected interaction point must lie in the intended region. Available robot-model checks assess reachability, joint limits, and obstacle clearance. Ordinary deployment uses observations and robot-state readings, without ground-truth object poses or privileged simulator segmentation. The simulator-assisted scheduling controls in \namedref{Section}{sec:focus-scheduling-protocol} are a separate diagnostic experiment.

\subsection{Correction context and code construction}
\label{sec:correction-context}

An instruction such as ``realign toward the opening'' leaves several questions unanswered. The policy must know which opening is intended, where it is in the current observation, and what the previous attempt did. The \texttt{CorrectionContext} in \namedref{Table}{tab:context} keeps the instruction, spatial evidence, executed history, and uncorrected proposal in separate fields. Each field retains enough provenance to associate it with the correct recovery state.

\metabrain supplies the instruction after reviewing diagnostic findings. Grounding agents attach object regions, interaction points, and spatial relations. The runtime supplies executed controls and their observed outcomes. Feasibility assessments inform both instruction construction and the decision to dispatch a candidate. This division lets different tools contribute through a common schema while preserving the distinction between an observation, an inferred relation, and a proposed action.

\begin{figure}[!t]
\centering
\includegraphics[width=0.95\linewidth]{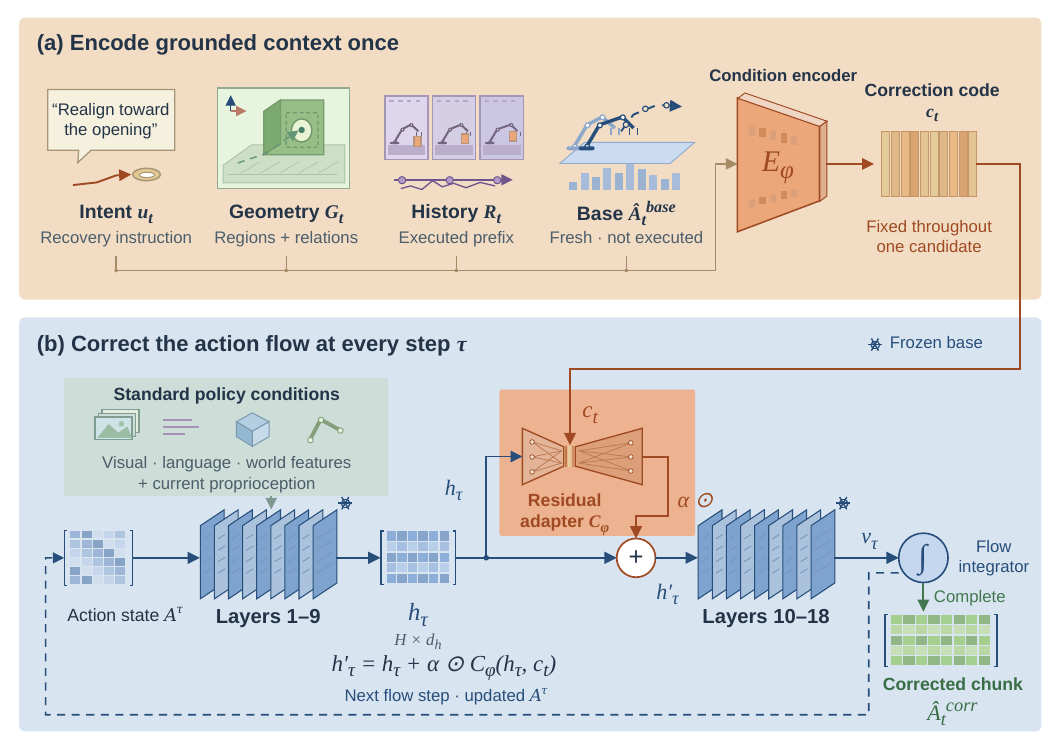}
\caption{The correction interface within action generation. Language, spatial evidence, executed history and a fresh base proposal are encoded once per candidate. At every flow step, the adapter uses the current action hidden states to produce position-dependent residuals. Injection after the ninth of 18 action layers allows subsequent layers to convert the modified state into a corrected flow velocity. Orange modules are trained during correction learning while the base model is frozen. All parameters remain fixed during deployment.}
\label{fig:correction}
\end{figure}

The recovery text and its referenced offset direction jointly form $u_t$. The failure window selects and links relevant executed records in $R_t$, while verified spatial evidence is encoded as $G_t$. These fields form the four code inputs. The task instruction $g$ enters the ordinary language-conditioning path. The \texttt{input\_features} field references observation features and action hidden states used by the policy and adapter. These runtime references are distinct from the four inputs encoded into the fixed correction code.

After recovery preparation, the runtime acquires fresh observations at step $t$. With correction disabled, \wla generates the unexecuted proposal $\hat A_t^{\mathrm{base}}$. Let $G_t$ encode the verified spatial evidence and let $R_t$ supply the relevant executed history. The condition encoder combines these inputs with $u_t$ to produce a latent correction representation
\begin{equation}
 c_t=E_\phi(u_t,G_t,R_t,\hat A_t^{\mathrm{base}}),
 \label{eq:code}
\end{equation}
where $\phi$ denotes all trainable correction-branch parameters, including the encoder, adapter, and residual scale below. The code $c_t$ is a continuous learned representation of the correction context. It stays fixed while one corrected chunk is generated. The same verbal instruction can therefore produce different corrections when the evidence, history, or base proposal changes. Current visual features and predicted future context still enter through the policy's ordinary conditioning paths.

\subsection{Correction adapter for action generation}
\label{sec:adapter}

The action expert generates a chunk by integrating a learned flow field. Physical time is indexed by $t$, while $\tau\in[0,1]$ indexes the internal noise-to-action process. For one policy call, let $h_\tau\in\mathbb R^{H\times d_h}$ be the action hidden states after layer nine, with hidden width $d_h$. We suppress the fixed physical index $t$ on these states. The residual adapter applies
\begin{equation}
 h'_\tau=h_\tau+\alpha\odot C_\phi(h_\tau,c_t),
 \label{eq:residual}
\end{equation}
where $\alpha\in\mathbb R^{d_h}$ is a learned channel scale, broadcast over the $H$ action positions, and $\odot$ denotes element-wise multiplication. The code $c_t$ is fixed for the chunk, but $h_\tau$ and the adapter output are recomputed at every integration step. The remaining nine transformer layers predict the corrected flow velocity (\namedref{Figure}{fig:correction}). The residual acts in hidden-feature space rather than directly specifying a joint-space offset.

Because the adapter reads each action position's hidden state, its adjustment can vary across the chunk. A change in approach direction need not impose the same offset on later transport actions. The adapter's output projection is initialized to zero and $\alpha$ to $0.1$, so the initial residual is zero and the base network output is unchanged. The adapter uses a bottleneck width of 256. The complete correction branch, including the condition encoder, has fewer than 100M trainable parameters.

\FloatBarrier
\begin{wrapfigure}[17]{R}{0.55\textwidth}
\centering
\captionsetup{font=footnotesize}
\includegraphics[width=\linewidth]{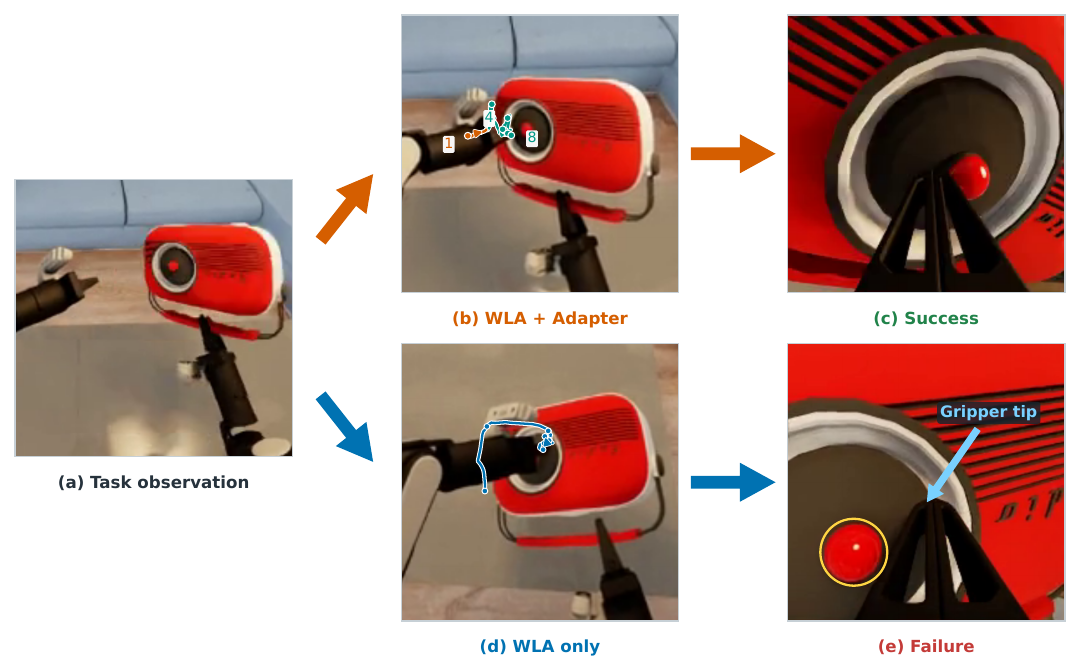}
\vspace{-2em}
\caption{Radio activation (Task 000). (a) Task context. (b--c) Adapter-guided motion (orange), base-policy continuation (teal), and success. (d--e) A separate \wla{}-only rollout misses the circled button. Trajectories project left-gripper motion. The branches show different rollout instances, with panel (a) providing context rather than a shared initial state.}
\label{fig:adapter-task000-execution}

\end{wrapfigure}

The task of turning on a radio (BEHAVIOR Task 000), shown in \namedref{Figure}{fig:adapter-task000-execution}, illustrates how this adapter translates a recovery instruction into a local motion adjustment. To activate the radio, the gripper must align with the small power button and then press toward the front panel. \metabrain expresses this adjustment in language. Combined with spatial evidence, executed history, and a fresh base-policy proposal, the instruction is encoded as $c_t$ (\equationref{eq:code}) and guides residual modulation of the action hidden states (\equationref{eq:residual}). The remaining action layers turn the modified representation into continuous controls while the base model parameters remain fixed.

\begin{figure}[!t]
\centering
\includegraphics[width=\linewidth]{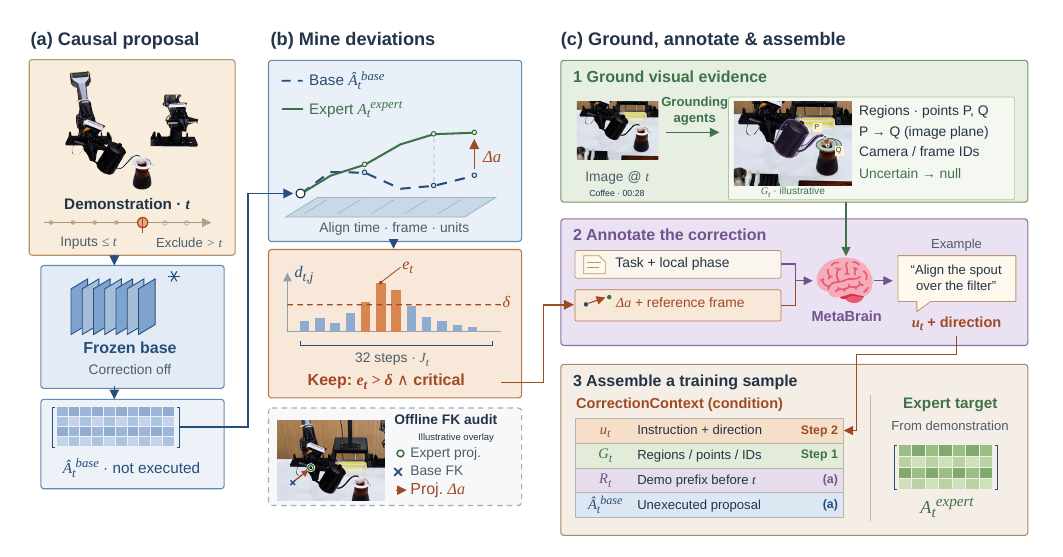}
\caption{Offline construction of correction supervision. A frozen policy predicts a candidate at a demonstration state. Candidate and expert actions are aligned in time, units and coordinates, and windows with a large deviation at critical manipulation steps are retained. Visual evidence and the candidate-to-expert direction support context construction. The expert continuation is a training target. The projected positions in the example are an offline quality check.}
\label{fig:mining}
\end{figure}

In the upper branch, the orange trajectory shows the adapter-guided adjustment. Base-policy execution then resumes in teal and completes the task. The lower branch shows a separate \wla{}-only rollout in which the gripper approaches beside the button. These examples illustrate the division of work between \metabrain's recovery instruction, adapter-guided motion, and subsequent base-policy execution.

\subsection{Correction data and adapter training}
\label{sec:correction-training}

We construct correction supervision offline from the demonstration corpus (\namedref{Figure}{fig:mining}). At a recorded state, the frozen base policy proposes an action chunk, which is compared with the expert continuation from that same state. Large deviations at critical manipulation steps provide candidate correction examples. These are unexecuted policy deviations, not observed physical failures. They teach the adapter how to change a proposal toward an expert continuation, while leaving a distribution gap to the failed execution states \mbox{encountered at deployment.}

Data mining focuses on skill transitions, pre-contact approach phases, and gripper actuation boundaries. At each sampled start $t$, the frozen base policy receives the task prompt, robot state, current images, and causal observation history. It generates one uncorrected candidate chunk using the standard inference procedure, including the configured flow-integration steps. Future observations and expert continuations do not \mbox{enter this candidate generation.}

To quantify divergence between the policy's proposal and the expert demonstration, both trajectories are mapped into an identical coordinate frame, kinematic unit system, and temporal indexing convention. A 32-step comparison window defines the directional residual and normalized error
\begin{equation}
 \Delta a_{t,j}=a_{t,j}-\hat a_{t,j}^{\mathrm{base}},\qquad
 e_t=\max_{j\in J_t} d(\hat a_{t,j}^{\mathrm{base}},a_{t,j}),
 \label{eq:mining}
\end{equation}
where $a_{t,j}$ is the expert action at local index $j$ in the segment starting at $t$, and $\hat a_{t,j}^{\mathrm{base}}$ is the corresponding base-policy prediction. The set $J_t$ contains positions within the 32-step comparison window that are both valid and marked as critical manipulation steps. Positions beyond a skill boundary or demonstration termination are excluded. The distance $d(\cdot,\cdot)$ normalizes differences using the configured scale of each action component. A window is retained when $e_t$ exceeds the threshold $\delta$. Windows without a valid critical position are skipped. Small perturbations of the sampled start around critical events provide coverage of \mbox{approach and timing variations.}

The annotation pipeline uses open-vocabulary object localization, SAM-based masks, and \metabrain to construct corrective instructions. The candidate-to-expert deviation indicates the required change, while recorded images locate the relevant objects and regions. The annotation prompt includes the task, local phase, deviation, and grounded entities with their camera and frame identifiers. Unresolved spatial fields remain null. Expert actions and future frames can inform offline labels, but they are not supplied to the base-policy proposal. The correction text describes the intended change and its reference frame, while the expert continuation supervises its metric realization.

Training and deployment use the same context schema with different evidence sources. Online, \texttt{failure\_window} identifies an observed execution anomaly. Offline, it marks the interval where an unexecuted proposal deviates from the demonstration. The historical fields contain only the demonstrated prefix before $t$, and the proposal is stored separately in \texttt{inferred\_action\_chunk}. This separation prevents an unexecuted candidate or a future target from being treated as experienced history.

\begin{figure}[!t]
\centering
\includegraphics[width=\linewidth]{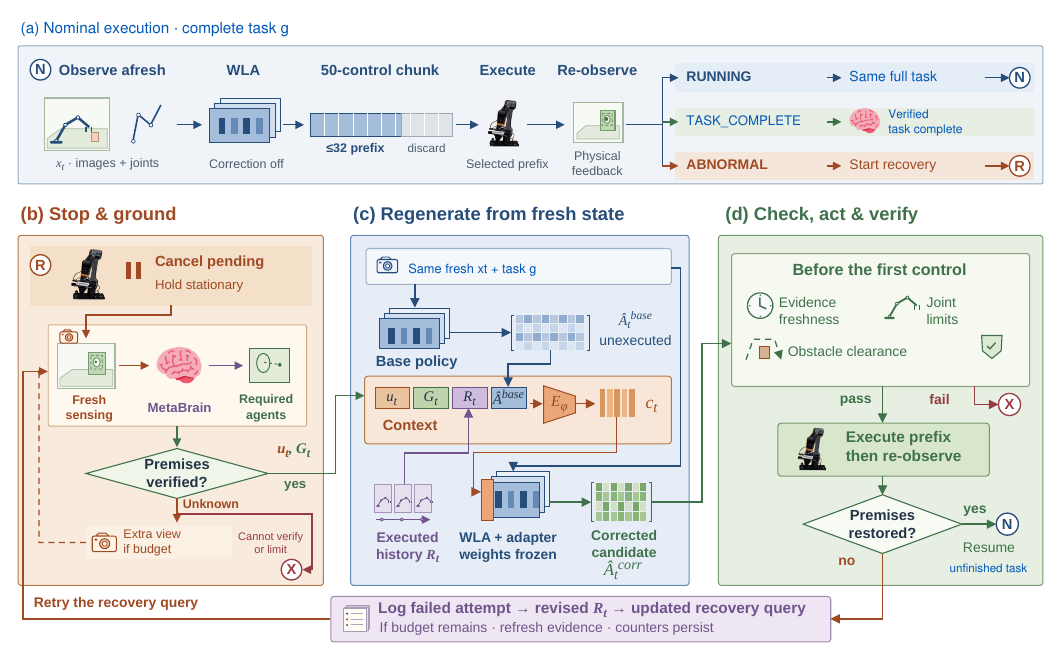}
\caption{Execution and recovery as one feedback loop. Ordinary action generation records the executed prefix and its observed outcome. An abnormal event invokes the \metabrain and the required grounding agents. A fresh recovery state conditions both the base proposal and corrected generation. Failed attempts update the evidence and consume the remaining event budget. Verified recovery resumes execution toward the unfinished task goal. A timeout, exhausted budget or failed feasibility check exits recovery.}
\label{fig:recovery}
\end{figure}

Each sample retains its trajectory identifier, timestamps, and base-checkpoint metadata. Overlapping windows are consolidated when measuring dataset duration. Mining makes one candidate-generation call at each sampled start, and previously mined samples can be reused or refreshed for another checkpoint. Retained supervision is capped at 2\% of the 1,950-hour corpus, or 39 hours, with priority given to critical steps that have complete spatial evidence. Schema checks and sampled manual inspection assess whether annotations agree with the images and trajectory deviations. Training freezes the base parameters $\theta$ and updates only the correction parameters $\phi$ against the expert continuation.


Correction training minimizes the masked flow-matching objective in \equationref{eq:flowloss}, with $c_t$ supplied as an additional condition. The base parameters $\theta$ remain frozen, and only the correction-branch parameters $\phi$ are updated. Expert continuations provide the action targets, using the same validity masks and flow-time convention as base-policy training. In this way, \metabrain's corrective knowledge enters action-policy learning through its generated instructions, and the encoder and adapter learn the mapping from that guidance to physical action. The knowledge is absorbed into the correction parameters $\phi$, while the pretrained action parameters $\theta$ remain unchanged. This is supervision of a learned correction interface, not direct distillation of \metabrain's internal representations. The mining offset $\Delta a_{t,j}$ is defined in control space, whereas the adapter predicts hidden-state residuals. The flow objective learns how those residuals should \mbox{change the generated actions.}

Correction training uses eight GPUs, BF16 precision, and a global batch size of 256. AdamW uses $\beta=(0.9,0.999)$, weight decay 0.01, and gradient-norm clipping at 1.0. Training runs for 6,000 steps, with a 1,000-step linear warm-up to $10^{-4}$ followed by cosine decay toward $10^{-5}$. Validation and checkpoint saving occur every 400 steps. Each evaluated adapter is paired with its corresponding base checkpoint.



\subsection{Closed-loop runtime orchestration}
\label{sec:recovery-protocol}

\namedref{Figure}{fig:recovery} summarizes the recovery loop. After detecting an anomaly, the runtime pauses pending controls and saves the observations, robot state, and executed history. Recovery preparation may include controlled stopping, posture adjustment, or contact release when required. The runtime then reacquires observations and robot state. Returning the robot to an earlier pose does not imply that the object or environment has been restored. \metabrain and the required diagnostic agents use this updated evidence to prepare the recovery instruction. Spatial evidence, the relevant observation history, the base proposal, and the corrected candidate must all correspond to the current recovery state. Before dispatch, the runtime checks evidence freshness, joint limits, and obstacle clearance. If the state changes during candidate generation, it refreshes the affected evidence and regenerates the candidate before execution.

After each corrective prefix, the runtime checks whether the required physical condition has been restored. If so, nominal execution resumes. Otherwise, the failed attempt and updated observations inform another recovery query. Each anomaly allows at most two additional viewpoints, each with a five-second allowance, and three physical retries. A \metabrain request times out after ten seconds, and the entire recovery event is limited to 60 seconds. Resubmitting a query does not reset these counters. The runtime aborts recovery when a required condition cannot be verified or an applicable budget is exhausted.

\metabrain determines how the intended operation should change, grounding agents assess the physical conditions needed for that change, and \wla generates the continuous controls. This division of labor allows reasoning and diagnostic capabilities to contribute through the shared \texttt{CorrectionContext}. New tools can be incorporated when their outputs preserve the field semantics, coordinate conventions, and representations expected by the correction encoder.

\FloatBarrier
\begin{figure}[!ht]
\centering
\includegraphics[width=\linewidth]{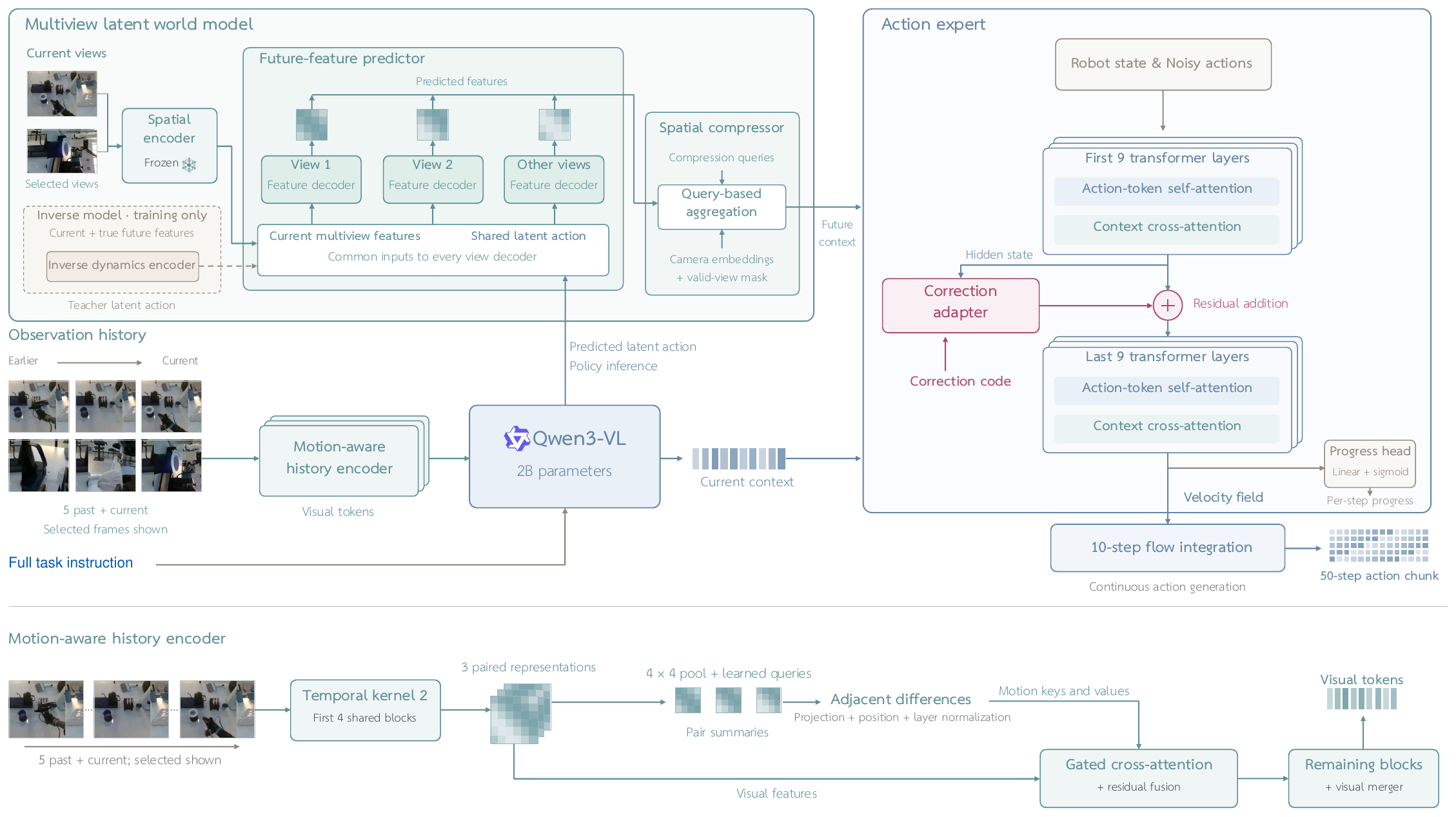}
\caption{\textbf{Architecture of \wla.} A motion-aware history encoder converts six-frame multiview histories into visual tokens. A 2B Qwen3-VL backbone combines these tokens with the task instruction to produce current context and a latent action. The world model predicts future visual features, which a spatial compressor aggregates into future context. An 18-layer action expert uses both contexts to generate a 50-step action chunk through ten-step flow integration, while a separate head estimates per-step progress. During recovery, a correction adapter injects a code-conditioned residual after the ninth layer. The lower panel shows how adjacent differences between paired-frame representations provide motion features for gated cross-attention and residual fusion before visual token merging.}
\label{fig:wla_framework}
\end{figure}

\Needspace{8\baselineskip}
\section{The \texorpdfstring{A\textsuperscript{2}-WLA}{A2-WLA} action model}
\label{sec:wla}

\wla supplies the continuous control capability on which grounded correction operates. It combines a 2B-parameter Qwen3-VL backbone~\cite{qwen3vl}, a multiview latent world model, and an approximately 300M-parameter Gemma-style action expert with 18 transformer layers (\namedref{Figure}{fig:wla_framework}). The backbone interprets the task and visual history. The world model predicts future scene features. The action expert uses both current and predicted context, together with proprioception, to generate continuous controls.
The history encoder, world model, and action expert are described first, followed by data construction and training. The architecture is shared across embodiments through a masked action representation, while camera inputs, state fields, and control semantics follow each robot's configuration.

The base policy is trained in three stages. World-model pretraining learns from multiview transitions. Vision-language pretraining adds action and embodied-language supervision. Joint optimization then trains the continuous policy and its auxiliary objectives. Correction learning is a separate fourth stage that freezes the base policy and trains the interface described in \namedref{Section}{sec:correction-training}.

\subsection{Motion-aware visual history encoding}
\label{sec:visual-history}

A single image often leaves an interaction ambiguous. An object between the gripper fingers may be securely held, slipping, or still resting on a surface. The history encoder supplies temporal evidence before visual tokens enter the language backbone. Each camera contributes five past frames and the current frame, arranged chronologically. A temporal patch kernel of size two groups them into three frame pairs. The first four visual blocks process the pairs with shared parameters.

Each paired representation is compressed with adaptive $4\times4$ pooling and learned queries. Let $f_k$ denote the compressed representation of pair $k\in\{0,1,2\}$. Adjacent differences provide temporal features
\begin{equation}
 d_k=\operatorname{LN}\bigl(W(f_{k+1}-f_k)+\operatorname{PE}(k)\bigr),\qquad k\in\{0,1\},
 \label{eq:history}
\end{equation}
where $W$ is a learned projection, $\operatorname{PE}(k)$ is a learned interval-position embedding, and $\operatorname{LN}$ denotes layer normalization. These features supply keys and values for gated cross-attention in later visual blocks. Residual fusion, the remaining visual blocks, and token merging produce the sequence passed to the language backbone. This design combines appearance with changes across the sampled observation window.

The history representation is recomputed from the sliding window at each policy call, without recurrent hidden-state accumulation. Its feature differences encode observed change rather than calibrated velocity, because the sampled frame intervals can vary. The longer execution record $R_t$ remains separate and is used for diagnosis. \namedref{Section}{sec:wla-data} specifies frame sampling and camera alignment.

\subsection{Multiview latent world model}
\label{sec:worldmodel}

Visual history describes recent changes, while the world model predicts the scene features relevant to the next action. The head camera captures workspace layout, and wrist cameras provide local views of the gripper and nearby objects. Occlusion and camera motion can make a transition ambiguous in any single view. Building on latent-action modeling in LaWAM~\cite{lawam} and Wave-0~\cite{wave-0}, we train a shared latent variable to explain changes across these complementary views. The representation is learned from visual transitions and is not tied to a fixed joint-vector dimensionality.

The world model contains a frozen visual encoder, an inverse model, and a future-feature predictor. For clarity, we suppress the physical step index $t$ within this subsection. Let $F=(F_1,\ldots,F_V)$ be the current patch features from $V$ camera views, and let $F^+$ contain features observed 50 control steps later. During world-model pretraining, the inverse encoder $E_{\mathrm{inv}}$ produces a shared transition latent $z^*$. The predictor then reconstructs future features for each view
\begin{equation}
 z^*=E_{\mathrm{inv}}(F,F^+),\qquad \hat F_v^+=D_v(F,z^*).
 \label{eq:teacher}
\end{equation}
Reconstructing all views from a shared latent variable encourages consistency across their predicted changes. This joint constraint can help resolve ambiguity in individual views.

At deployment, future frames are unavailable. Learned queries in the vision-language backbone instead predict a latent $\hat z$ from current and historical observations and the task instruction. The backbone also supplies current-context tokens $C$. The future predictor uses $\hat z$, and a spatial compressor $P$ aggregates its outputs into future-context tokens
\begin{equation}
 \hat F_v^+=D_v(F,\hat z),\qquad C^+=P(\hat F^+),
 \label{eq:futureconditions}
\end{equation}
where $\hat F^+=(\hat F_1^+,\ldots,\hat F_V^+)$. Camera embeddings preserve view identity, and validity masks exclude missing views. We write $C^+\in\mathbb R^{N_{\mathrm{fut}}\times d_c}$, where $N_{\mathrm{fut}}$ is the total number of compressed tokens across views and $d_c$ is their context width. Attention projections map these tokens to the action expert's hidden width $d_h$.


\subsection{Action expert with correction adapter}
\label{sec:action-expert}

The action expert receives current-context tokens $C$, future-context tokens $C^+$, and a projection of the proprioceptive state $s_t$. Its 18 transformer layers alternate between the two context streams. With zero-based indexing, even layers cross-attend to $C$ and odd layers cross-attend to $C^+$. Every layer also applies self-attention among the action tokens. This gives both observed and predicted scene information repeated opportunities to influence the generated trajectory.

An embodiment-specific input/output mapping handles each robot's state fields and control semantics. For shared tensor processing, an action chunk with $d_a$ active control dimensions is padded to width $d_{\max}$. A validity mask excludes padded dimensions and invalid timesteps from losses and dispatch. In the learning equations below, $A$ and $\hat A$ denote these normalized, padded chunks. At execution, the output mapping removes padding and restores native units. This keeps the learning representation distinct from the physical control interface in \equationref{eq:policy}.

The action expert uses conditional flow matching~\cite{flowmatching}. For one policy call, we omit the fixed physical index $t$ and write the noisy chunk as $A^\tau$. The velocity network $v_\theta(A^\tau,\tau\mid s,C,C^+)$ is conditioned on the current state $s=s_t$ and both context streams. Inference starts from Gaussian noise at $\tau=1$ and integrates toward $\tau=0$ in ten numerical steps. The integration rule is deterministic conditional on the initial noise, while the sampled noise makes the policy stochastic. Physical execution time and internal flow time are separate. \namedref{Section}{sec:training} defines the path and target velocity.

Condition tokens remain fixed during numerical integration, so their projected keys and values are computed once and cached across flow steps. The noisy action tokens and velocity estimates are updated at each step. During recovery, the continuous correction code $c_t$ conditions a residual adapter after the ninth transformer layer, corresponding to zero-based index 8. The adapter adds $\alpha\odot C_\phi(h_\tau,c_t)$ to the action hidden states, and the remaining nine layers predict the corrected flow velocity. The code stays fixed for one candidate, while the hidden states and residual are recomputed at every integration step. All policy and correction-branch parameters remain fixed during deployment.

A separate head predicts local-phase progress at each action position. The BEHAVIOR reference configuration predicts 50 controls and executes at most 32 before requesting a new chunk. Longer predictions provide context for the dispatched prefix, and the next observation allows the policy to revise the remaining movement. Predicted progress helps monitoring but does not establish physical completion. A detected abnormal outcome invokes the recovery interface.

\subsection{Training details}
\label{sec:training}

\subsubsection{Training sample construction}
\label{sec:wla-data}

Each training sample is anchored at one physical step. It contains the task instruction, robot state, synchronized multiview history, and a target action chunk. Recorded future frames and phase annotations provide additional targets when available. Input histories end at the anchor step. Future observations and actions appear only as supervised targets or offline annotation evidence.

For the BEHAVIOR 2026 configuration described here, training uses the official corpus of 20,000 teleoperated demonstrations, totaling approximately 1,950 hours across 100 household tasks and seven scenes~\cite{behavior2026}. Demonstrations are collected through the JoyLo whole-body interface, which coordinates the mobile base, torso, arms, and grippers. The average trajectory lasts 5.86 minutes, and the annotations cover 31 skill primitives, including grasping, pouring, wiping, and articulated-object manipulation~\cite{behavior2026dataset}. These extended trajectories require the policy to handle object-state changes and dependencies between successive operations.

Recorded timestamps align multiview images, proprioceptive states, and robot controls. Task instructions and skill intervals supply the corresponding language and phase annotations. Official head-camera streams recorded at $720\times720$ resolution and dual wrist-camera streams recorded at $480\times480$ resolution are bilinearly downsampled to $224\times224$ for policy training, matched with continuous robot states and controls at 30\,Hz. Self-supervised world model pretraining ingests visual inputs at $256\times256$ spatial resolution. Robot states and action dimensions use quantile normalization computed from the training split. The embodiment adapter instantiates 23-dimensional absolute continuous commands encompassing whole-body actuation channels.

Training windows retain the complete task instruction, while skill annotations define the temporal boundaries used for auxiliary progress supervision and transition sampling. When skill intervals overlap, the earlier-starting skill determines the phase label. Action supervision continues across skill boundaries wherever trajectory steps are valid. Separate masks identify valid phase labels, progress targets, and control dimensions, so a missing auxiliary label does not discard otherwise valid action supervision.

\paragraph{History sampling schedule.} Each training input contains five past frames and the current frame. We sample one positive look-back offset from each of the five intervals bounded by $[30,84,138,192,246,300]$, using a log-uniform distribution within each interval. The frame at offset $\Delta$ is taken from $t-\Delta$. We append the frame at $t$ and sort the six frames chronologically, using the same timestamps across cameras. Deployment uses fixed relative offsets $[-256,-205,-154,-60,-30,0]$. At 30\,Hz, training look-back offsets span one to ten seconds, and the oldest deployment frame is 8.53 seconds old. A lookup before episode start repeats the initial frame and carries a temporal-padding flag.

\paragraph{Transition sampling and regularization.} The boundary-sampling configuration increases sample weights fivefold within 60 frames on either side of a primitive transition. Proprioceptive dropout decreases linearly from 0.3 to 0.05 over the first 60\% of training. This encourages the policy to use visual context rather than rely only on robot state. Non-gripper action errors receive velocity-dependent weights in $[0.5,1.5]$, giving more emphasis to faster demonstrated movements. These weights are distinct from the binary validity masks. Optional pace conditioning appends a quantized demonstration-duration category to the task text. At inference, the category comes from a checkpoint-specific task mapping or an explicit request, not the unknown duration of the ongoing episode.

\subsubsection{World-model and vision-language pretraining}
\label{sec:pretraining}

The first two stages learn representations before continuous-policy optimization. Stage 1 trains the world model from multiview transitions. Stage 2 trains the vision-language backbone with latent-action targets, FAST action tokens, continuous action candidates, and embodied question answering.


We denote losses by $L$ and stop-gradient by $\sg(\cdot)$. The operators $\operatorname{MSE}_M$ and $\operatorname{MAE}_M$ average squared and absolute error over entries enabled by mask $M$. Masks refer to the domain of each loss. They may select valid action dimensions, timesteps, camera views, or annotations. A loss with no valid target entries is omitted.

\paragraph{World-model pretraining.} Stage 1 trains the inverse model and future-feature predictor while keeping the visual encoder frozen. Let $m_v\in\{0,1\}$ indicate a valid current--future pair for camera view $v$, and let $w_v>0$ be its reconstruction weight. Using the transition latent $z^*$, we minimize
\begin{equation}
 L_{\mathrm{world}}=
 \frac{\sum_{v=1}^{V}w_vm_v\,\operatorname{SmoothL1}(\hat F_v^+,F_v^+)}{\sum_{v=1}^{V}w_vm_v}
 +\lambda_{\mathrm{reg}}\operatorname{Reg}(z^*),
 \label{eq:lamloss}
\end{equation}
where $\operatorname{SmoothL1}$ compares predicted and observed feature tensors, $\operatorname{Reg}$ denotes latent regularization, and $\lambda_{\mathrm{reg}}$ controls its weight. Samples without a valid current--future view pair are excluded. After Stage 1, the inverse encoder is frozen to provide latent-action targets. The future predictor is transferred to the policy and updated again during joint continuous-action training.

\paragraph{Shared vision-language supervision.} Stage 2 trains the vision-language backbone and its prediction heads with four complementary objectives
\begin{equation}
 L_{\mathrm{pre}}=
 L_{\mathrm{FAST}}+
 \lambda_{\mathrm{AQ}}L_{\mathrm{AQ}}+
 \lambda_{\mathrm{LA}}L_{\mathrm{LA}}+
 \lambda_{\mathrm{VQA}}L_{\mathrm{VQA}}.
 \label{eq:pretraining}
\end{equation}
Each loss is evaluated only for samples with the corresponding targets. The shared backbone processes visual context and task language, while output queries and heads supply the different predictions. The following paragraphs define each supervision stream. During teacher-forced language or action-token training, causal masking prevents a prediction from seeing later target tokens. Policy-conditioning queries must likewise remain independent of answer and action targets.

\topicmarker{} \textit{Latent-action prediction.} A latent-action head maps learned query tokens to a predicted transition vector $\hat z$. It learns from the frozen inverse model through
\begin{equation}
 L_{\mathrm{LA}}=\|\hat z-\sg(z^*)\|_2^2.
 \label{eq:latent-target}
\end{equation}
The frozen teacher uses current and future observations to produce the target $z^*$. The student predicts $\hat z$ from current and past observations and language. Stop-gradient keeps the teacher fixed while the student learns to predict its transition representation. Future observations are therefore used for supervision, not as \mbox{deployed policy inputs.}

\topicmarker{} \textit{Discrete actions with FAST.} The FAST frequency-domain tokenizer~\cite{fast} converts an action chunk $A$ into discrete tokens $y_1,\ldots,y_T$. Predicting this sequence teaches the backbone to represent action trajectories. Training uses teacher-forced cross-entropy
\begin{equation}
 L_{\mathrm{FAST}}=-\sum_{r=1}^{T}\log p_\theta(y_r\mid x,g,y_{<r}),
 \label{eq:fast}
\end{equation}
where $T$ is the token-sequence length, $p_\theta$ is the conditional distribution over the discrete vocabulary, and $y_{<r}$ is the target token prefix. The FAST configuration uses a vocabulary of 2,048 tokens and a sequence budget of 256 tokens. FAST supplies action-sequence supervision to the backbone, while the flow-matching action expert generates continuous controls at deployment.

\topicmarker{} \textit{Continuous candidates and scores.} Parallel action queries predict $N$ continuous trajectory candidates $\hat A^{(n)}$ and their scalar errors $\hat e_n$. For candidate $n$, the supervised error is $e_n=\operatorname{MAE}_M(\hat A^{(n)},A)$, with mask $M$ excluding inactive control dimensions and padded steps. We train the closest candidate against the demonstrated chunk and train the error head against detached candidate errors
\begin{equation}
 n^*=\arg\min_n e_n,\qquad
 L_{\mathrm{AQ}}=e_{n^*}+\frac{\lambda_{\mathrm{score}}}{N}
 \sum_{n=1}^{N}(\hat e_n-\sg(e_n))^2.
 \label{eq:aq}
\end{equation}
Selecting one winning candidate for the entire horizon encourages distinct trajectory hypotheses rather than requiring every candidate to approximate the same demonstrated path. Detaching the target error $e_n$ prevents the score-regression loss from updating trajectory predictions through that target. Gradients through the predicted score $\hat e_n$ can still update the score head and any shared representations on its branch.

\topicmarker{} \textit{Embodied question answering and local goals.} Stage 2 also trains the backbone to answer questions about scene layout, task progress, and the next local goal. Inputs contain multiview frames, the task instruction, and a question. Demonstration-derived answers describe milestones such as repositioning an object or orienting it toward a receptacle. These goals use the standard language-output head. Target answers enter only the teacher-forced answer sequence and remain hidden from policy-conditioning queries.

\subsubsection{Joint policy optimization}
\label{sec:wla-joint-training}

Stage 3 jointly trains the VLM backbone, future predictor, action expert, and enabled auxiliary heads. Flow matching supplies the main continuous-action objective. Action-query supervision continues on samples with valid trajectories, and auxiliary losses constrain predicted features, action representations, geometry, and task information. FAST token prediction is disabled during this stage.

\noindent\topicmarker{} \textbf{Conditional flow matching.} The continuous action expert models whole-body trajectories by learning a time-dependent velocity field along a probability path connecting pure Gaussian noise to demonstration segments. For a demonstrated trajectory $A$, we construct the \mbox{intermediate noisy segment}
\begin{equation}
 A^\tau=(1-\tau)A+\tau\epsilon,\qquad \epsilon\sim\mathcal{N}(0,I),
 \label{eq:flowpath}
\end{equation}
where $\tau\in[0,1]$ denotes continuous flow time and $I$ represents the identity covariance. This linear interpolation between demonstration actions and Gaussian noise has target velocity $\epsilon-A$. The action expert network $v_\theta$ is supervised to regress this velocity conditioned on the noisy segment, proprioceptive state $s$, causal context embeddings $C$, and predicted prospective visual features $C^+$ through
\begin{equation}
 L_{\mathrm{FM}}=\mathbb{E}_{\tau,\epsilon,A}\left[
 \operatorname{MSE}_M\bigl(v_\theta(A^\tau,\tau\mid s,C,C^+),\,\epsilon-A\bigr)
 \right].
 \label{eq:flowloss}
\end{equation}
The expectation uses demonstrated action chunks, standard Gaussian noise, and uniformly sampled $\tau\in[0,1]$. The mask $M$ excludes padded steps and inactive control dimensions. Since the target is the derivative with respect to increasing $\tau$, inference integrates the learned field with negative time increments from one to zero. This matches the noise-to-action convention in \namedref{Section}{sec:action-expert}.

\noindent\topicmarker{} \textbf{Prospective feature consistency.} To retain predictive visual information during supervised policy updates, joint optimization aligns causal world predictions with observed future features through
\begin{equation}
 L_{\mathrm{future}}=
 \frac{\sum_{v=1}^{V}m_v\|\hat F_v^+-F_v^+\|_F^2}{\sum_{v=1}^{V}m_v},
 \label{eq:futureloss}
\end{equation}
where the squared Frobenius norm sums feature errors within a view. Here $m_v=1$ requires a valid current--future feature pair, and samples without one are excluded. This equation expresses the feature-consistency objective at a conceptual level. The training recipe records separate distillation and perceptual components for this purpose, as distinguished below.

\noindent\topicmarker{} \textbf{Action-chunk representation alignment.} We pool the action expert's tokens over the prediction horizon, project them, and normalize the resulting vector to obtain $\hat z_a$. A fixed linear projection of the flattened target chunk, followed by the same normalization, produces $z_a$. The alignment objective is
\begin{equation}
 L_{\mathrm{align}}=\operatorname{MSE}(\hat z_a,\sg(z_a)).
 \label{eq:actionalignment}
\end{equation}
The target $z_a$ summarizes the current sample's complete action chunk, rather than the chunk from a later policy call. It is used only for supervision and is not supplied to the deployed policy. The objective encourages chunk-level representation consistency, while flow matching determines the individual control values.

\noindent\topicmarker{} \textbf{End-effector geometry.} Joint-space errors can have different Cartesian consequences across poses. We therefore compare predicted and target end-effector motion through a common whole-body kinematic mapping $K_r$. Let $\hat A$ denote the predicted clean action chunk supplied to this auxiliary objective. Both chunks are converted to the same physical units and evaluated from the same initial state $s$
\begin{equation}
 L_r=\operatorname{MSE}_{M_r}\bigl(K_r(s,\hat A),K_r(s,A)\bigr),
 \qquad r\in\{\mathrm{pos},\mathrm{rot}\},
 \label{eq:kinematic}
\end{equation}
where $r$ selects position or rotation increments of the active end-effectors. The mask $M_r$ selects valid end-effectors, timesteps, and geometric components. Both branches use the same reference frame and rotation convention. These losses encourage agreement with demonstrated motion. They do not test collision avoidance, contact stability, or dynamic feasibility.

\noindent\topicmarker{} \textbf{Semantic identity and temporal progress.} For samples with a valid task identifier, an auxiliary classification head averages the final-layer action representations and predicts task identity through cross-entropy $L_{\mathrm{task}}=-\log p_{\mathrm{task}}(y_{\mathrm{task}})$, where $y_{\mathrm{task}}$ is the task label. This objective preserves task information in the action features without replacing the language instruction. It is separate from the local-phase and progress objectives \mbox{grouped in $L_{\mathrm{status}}$.}

A separate linear head with sigmoid activation predicts relative skill-phase progress $\hat p_{t,j}$ at each action position. For the annotated skill phase $[t_0,t_1)$ active at the anchor step $t$, the target is the fraction of the interval elapsed after the corresponding action step
\begin{equation}
 p_{t,j}=\operatorname{clip}\left(\frac{t+j-t_0+1}{t_1-t_0},0,1\right).
 \label{eq:progress}
\end{equation}
Only positions satisfying $t_0\leq t+j<t_1$ contribute to the masked mean squared progress loss. Progress masks are defined separately from action-validity masks. The status objective $L_{\mathrm{status}}$ aggregates the enabled local-phase and progress objectives and excludes the task-identity loss $L_{\mathrm{task}}$.

Progress is learned from relative positions within demonstrated skill phases and is not a probability of physical completion. The predicted curve describes a proposed action chunk, including its unexecuted suffix. Feedback to \metabrain refers to the observed local phase and is interpreted together with the executed prefix and observed outcomes. Whole-episode progress is not substituted for skill-phase progress. A value near one does not independently establish physical completion, which requires verification of \mbox{the physical completion conditions.}

The total training objective combines these terms
\begin{equation}
 \begin{aligned}
 L_{\mathrm{train}}={}&L_{\mathrm{FM}}+L_{\mathrm{prior}}+\lambda_{\mathrm{AQ}}L_{\mathrm{AQ}}
 +L_{\mathrm{aux}}+0.1(L_{\mathrm{distill}}+L_{\mathrm{perc}}),\\
 L_{\mathrm{aux}}={}&L_{\mathrm{task}}+L_{\mathrm{status}}+L_{\mathrm{align}}
 +5\times10^{-6}L_{\mathrm{pos}}+10^{-3}L_{\mathrm{rot}},
 \end{aligned}
 \label{eq:fullloss}
\end{equation}
where $L_{\mathrm{distill}}$ and $L_{\mathrm{perc}}$ implement the future-feature consistency objective in \equationref{eq:futureloss}. The conceptual $L_{\mathrm{future}}$ is not added again. The Gaussian-prior loss $L_{\mathrm{prior}}$, task-identity loss $L_{\mathrm{task}}$, and aggregate status loss $L_{\mathrm{status}}$ each enter the total objective with unit weight. End-effector position and rotation losses have weights $5\times10^{-6}$ and $10^{-3}$, respectively. Each objective is evaluated only where its required observations \mbox{and labels are valid.}

Following the idea of knowledge insulation~\cite{knowledgeinsulation}, we attenuate gradients through the current-context token path from the VLM to the action expert. For a representation $h$ on this path, we apply
\begin{equation}
 \operatorname{KI}(h)=\gamma h+(1-\gamma)\sg(h).
 \label{eq:ki}
\end{equation}
This operator preserves the forward activation and scales gradients on this path by $\gamma\in[0,1]$. We apply it to current-context tokens supplied to the action expert. It does not freeze the VLM or scale every gradient entering it. Future-context prediction and auxiliary supervision remain separate paths through \mbox{the training graph.}

\begin{table}[!t]
\centering\small\setlength{\tabcolsep}{4pt}
\caption{Recorded schedules for the three base-model stages. The correction schedule is given in \namedref{Section}{sec:correction-training}. Training configuration and systems throughput measurements have different hardware scales and are reported separately.}
\label{tab:training}
\begin{tabularx}{\linewidth}{@{}p{3.15cm}YYY@{}}
\toprule
Setting & World model & VLM & Continuous action \\
\midrule
Global batch & 4,608 & 2,048 & 2,048 \\
Per-GPU batch & 72 & 32 & 32 \\
Training plan & Up to 10 epochs & 150,000 steps & 500,000 steps \\
Save interval & 5,000 steps & 20,000 steps & 20,000 steps \\
Initial learning rate & $3\times10^{-4}$ & $7.5\times10^{-5}$ & $7.5\times10^{-5}$ \\
Warm-up & 10,000 steps & 2,000 steps & 2,000 steps \\
Later schedule & Constant & Cosine to $2.5\times10^{-6}$ & Cosine to $2.5\times10^{-6}$ \\
Optimizer & AdamW & AdamW & Local Muon and AdamW \\
Weight decay & 0.01 & 0.001 & 0.001 \\
Gradient clipping & 1.0 & 1.0 & 1.0 \\
Precision & BF16 & BF16 & BF16 \\
Image resolution & $256\times256$ & $224\times224$ & $224\times224$ \\
Updates & Inverse dynamics model and future predictor & VLM and enabled heads & VLM, future predictor, action expert and enabled heads \\
\bottomrule
\end{tabularx}
\end{table}

\subsubsection{Sampling schedules, optimization, and hyperparameters}
\label{sec:training-config}

The objectives above specify supervision. The following schedules specify how training data, flow evaluations, and parameter updates are allocated.

\noindent\topicmarker{} \textbf{Data mixture and adaptive flow sampling.} Training batches sample heterogeneous datasets according to configured mixture probabilities. For each demonstration $i$, the action expert first evaluates $N_0=2$ noise-time pairs. Their masked flow losses provide a difficulty score
\begin{equation}
 r_i=\operatorname{Mean}_{1\leq r\leq N_0}(L_{i,r})
 +\beta_{\mathrm{sample}}\operatorname{Std}_{1\leq r\leq N_0}(L_{i,r}),
 \label{eq:flow-difficulty}
\end{equation}
where $\beta_{\mathrm{sample}}$ controls the contribution of loss variation. More difficult demonstrations receive additional flow evaluations, with at most eight pairs per demonstration and a target batch average of four.

To separate the sampling budget from each demonstration's explicit loss weight, we average flow losses within each demonstration and then across the batch
\begin{equation}
 L_{\mathrm{batch}}=\frac1B\sum_{i=1}^{B}\left(\frac1{N_i}\sum_{r=1}^{N_i}L_{i,r}\right),
 \label{eq:adaptive-flow}
\end{equation}
where $B$ is the batch size, $N_i\in\{2,\ldots,8\}$ is the number of flow evaluations for demonstration $i$, and $L_{i,r}$ is its masked flow loss at sample $r$. This reduction gives every demonstration the same explicit outer weight while assigning more computation to difficult examples. The multimodal backbone runs once per demonstration, and its representations are shared across flow branches. Backbone objectives independent of flow time are evaluated once. Auxiliary objectives that depend on action-suffix states use one designated branch, so their effective weights do not grow with $N_i$.

\noindent\topicmarker{} \textbf{Optimization schedules.} \namedref{Table}{tab:training} summarizes the three base-model stages. Each uses 64 GPUs. The world-model checkpoint transferred to later stages is taken at step 45,000. The final 1\% of recorded sequence intervals forms an internal validation partition for monitoring world-model training, separate from the downstream benchmark evaluation.

Optimizer dynamics are specialized according to layer types across training phases. Stage~1 applies the AdamW optimizer with hyperparameter settings $\beta=(0.9, 0.999)$, numerical stabilization constant $\epsilon_{\mathrm{opt}}=10^{-8}$, and unit gradient accumulation, excluding normalization gains and linear biases from weight decay regularization. Stage~2 combines FAST and VQA cross-entropy, continuous action-query regression, and latent-action distillation. The continuous action expert is not evaluated in this stage. During Stage~3 continuous-action training, structured weight matrices within attention projections and feed-forward layers are optimized via Local Muon, while remaining embedding and auxiliary layers are updated via AdamW. Muon configurations maintain a momentum factor of 0.95, Nesterov momentum acceleration, and five internal \mbox{Newton--Schulz orthogonalization iterations.}

\noindent\topicmarker{} \textbf{Loss weights.} Stage 2 uses $\lambda_{\mathrm{AQ}}=1.0$, $\lambda_{\mathrm{LA}}=0.1$, and $\lambda_{\mathrm{VQA}}=1.0$. Stage 3 reduces the action-query weight to $\lambda_{\mathrm{AQ}}=0.1$ and uses knowledge-insulation scale $\gamma=0.1$ on the current-context path. Task and status supervision follow \equationref{eq:fullloss}. The phase-classification component has an implementation default of 0.1 and must be explicitly configured for the reported status-supervision recipe.

\FloatBarrier
\Needspace{8\baselineskip}
\section{Infrastructure for Efficient Training and Deployment}
\label{sec:efficiency}

Long-horizon embodied learning is limited by the cost of collecting, processing, and evaluating extended interaction sequences. For \wla, multiview history encoding and repeated flow evaluations make training expensive. In BEHAVIOR, simulation, rendering, policy inference, and variable episode lengths also make full-suite evaluation costly. Our infrastructure addresses these costs through efficient data access, shared model computation, distributed optimization, and parallel simulation scheduling.

We report training and evaluation comparisons separately. Training comparisons measure demonstration throughput under the stated hardware and batch configurations. For evaluation, we compare the measured optimized worker pool with a naive eight-GPU baseline projected from recorded unoptimized serial-worker costs. The projection normalizes the GPU count by assuming ideal linear parallelism, while preserving the reference per-worker cost.

\subsection{Training Optimization and Distributed Scaling}
\label{sec:training-infra}

\begin{figure}[!t]
    \centering
    \includegraphics[width=\textwidth]{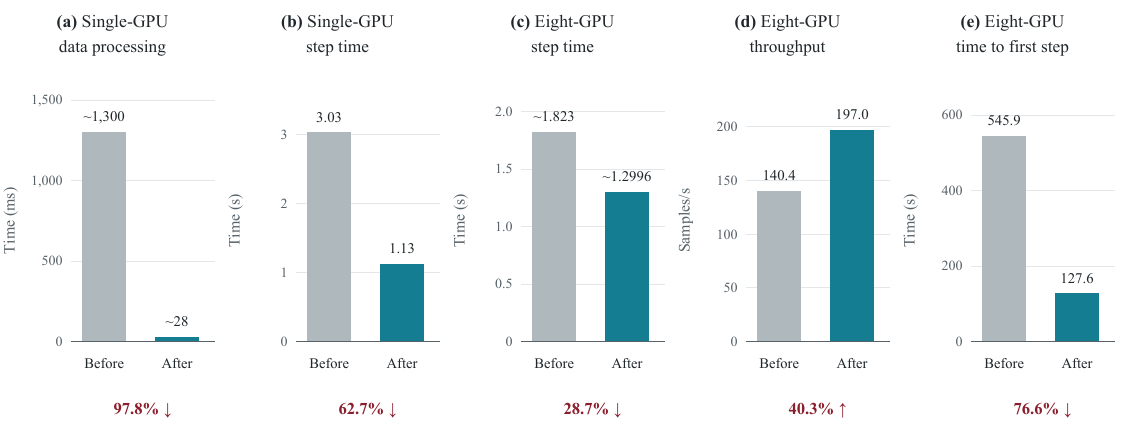}
    \caption{Data processing and distributed execution performance before
    and after optimization. \textbf{(a)} single-GPU data processing time,
    \textbf{(b)} single-GPU step time, \textbf{(c)} eight-GPU step time,
    \textbf{(d)} eight-GPU throughput, and \textbf{(e)} eight-GPU time to first step.
    Approximate values are
    marked with $\sim$. Percentages below the panels indicate relative
    decreases ($\downarrow$) or increases ($\uparrow$), calculated from
    the reported values. Single-GPU and eight-GPU comparisons use their respective fixed configurations and do not form one continuous ablation sequence.}
    \label{fig:data-optimization}
\end{figure}

\noindent\topicmarker{} \textit{I/O and Data Ingestion.} Continuous action optimization via flow matching constitutes the primary computational bottleneck within the multi-stage training pipeline. 
In the baseline single-device implementation, each optimization step requires approximately 3.03 seconds, with input processing alone accounting for 1.30 seconds. 
This delay stems from redundant visual decoding, where identical temporal video frames are repeatedly decompressed across overlapping observation windows. 
To eliminate this bottleneck, we consolidate frame extraction through batch JPEG decompression, map demonstration datasets into Apache Arrow shared-memory caches, and streamline inter-dataset indexing. 
These adjustments reduce per-step input preparation time from 1.30 seconds to 28 milliseconds. 
Together with asynchronous background data prefetching, single-GPU step latency drops to 1.13 seconds, representing a $2.7\times$ training acceleration. 
Furthermore, expanding dataset diversity from ten to 100 manipulation tasks incurs only a 3\% throughput penalty, confirming that data routing scales with task coverage without introducing proportional preprocessing overhead. \namedref{Figure}{fig:data-optimization} summarizes separate measurements of data preparation, step latency, throughput, and startup time.

\noindent\topicmarker{} \textit{Distributed Startup and Synchronization.} Distributed training across multi-GPU clusters frequently suffers from slow initial startup and hidden communication stalls. 
During job initialization, dozens of worker processes often attempt to parse the same raw dataset files simultaneously, causing severe storage contention and prolonged idle waiting. 
We parallelize dataset preparation, share sampler weights from rank zero, and delay parameter initialization on non-primary ranks to reduce repeated loading.
During active training steps, device synchronization is further streamlined by locking the backward communication order into a fixed static schedule. 
This configuration groups scattered parameter gradients into contiguous memory buffers for unified all-reduce synchronization, while eliminating redundant broadcast transfers for static model buffers that do \mbox{not change during training.}

\noindent\topicmarker{} \textit{Prefix Computation Sharing.} A substantial computational redundancy in continuous flow matching arises from sampling multiple noise-time pairs $(\tau, \epsilon)$ for each demonstration trajectory. 
Although these stochastic pairs evaluate distinct flow targets, they share identical visual histories, language instructions, and proprioceptive states. 
Rather than evaluating the complete model repeatedly, our implementation computes the multimodal prefix once through the vision-language backbone. 
The resulting representations are shared across the action expert's noise-time evaluations, which are batched instead of run serially. The reference configuration uses four evaluations per demonstration on average, while the adaptive schedule varies the number by sample.
Each flow sample processes its own noisy continuous action and time coordinate, whereas resulting backpropagation gradients accumulate into the shared prefix representation. 
For the same samples and masks, shared-prefix execution preserves the per-demonstration reduction in \equationref{eq:adaptive-flow}. Gradients from the action branches accumulate into the shared prefix. This equivalence concerns the shared-prefix computation and does not establish equivalence across complete training trajectories.
Together, shared-prefix execution and Arrow caching achieve 261.9 samples per second \mbox{on eight GPUs.}

\noindent\topicmarker{} \textit{Kernel Stabilization and Optimizer Sharding.} Additional latency in deep transformer training arises from dynamic sequence lengths and kernel launch overhead. 
We mitigate these delays by caching static tokenizer outputs, rotary position embeddings, and task metadata whenever input sequences permit reuse. 
Stable sequence-length buckets and padding upper bounds that increase only when necessary reduce repeated Triton compilation and autotuning.
These optimizations reduce static host-side overhead from 33 milliseconds to 0.8 milliseconds and forward latency from 243.7 to 210.7 milliseconds. 
At the optimization boundary, we coordinate AdamW with the matrix-oriented Local Muon optimizer. 
We distribute 126 independent matrix updates across eight local ranks according to their computational cost. Each update uses the configured Newton--Schulz iterations, and the results are gathered in BF16.
Sharded Local Muon reduces optimizer latency from 67.0 to 19.7 milliseconds. Bitwise equivalence was verified for the compared sharded optimizer updates.

The final SDPA-based implementation reaches 427.1 demonstration samples per second on eight GPUs with global batch size 256. Multiple flow samples from one demonstration count as one demonstration sample.
The $\pi_{0.5}$ reference reaches 106 demonstration samples per second on the same GPU hardware with matching image resolution and action horizon. Each implementation uses a batch size chosen to utilize the available resources. The resulting $4.03\times$ ratio compares implementation throughput under these conditions.
At 64 Pro6000 GPUs, the reported scaling efficiency is approximately 98\% relative to ideal linear scaling for the tested configuration. \namedref{Figure}{fig:optimization-gains} illustrates the aforementioned optimization gains.


\begin{figure*}[h]
    \centering
    \includegraphics[width=\textwidth]{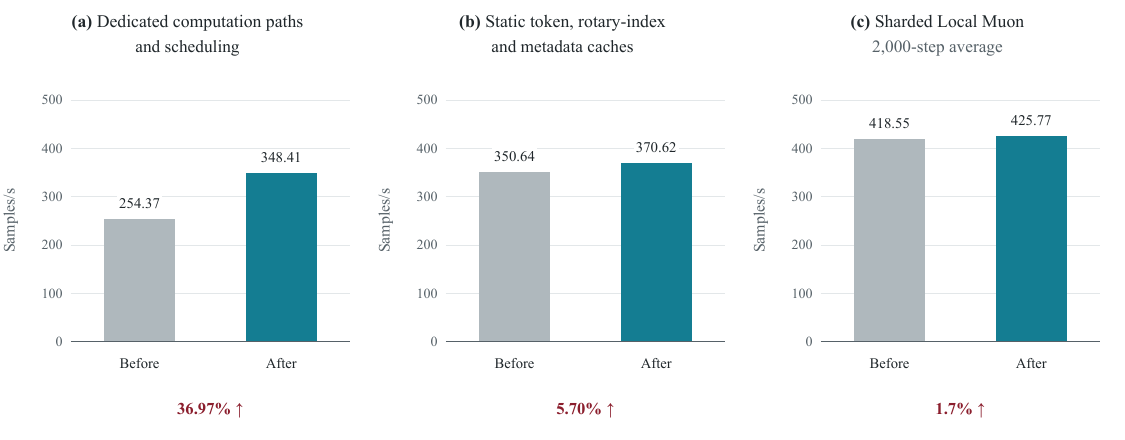}
    \caption{Throughput gains from
    \textbf{(a)} dedicated computation paths and scheduling,
    \textbf{(b)} static token, rotary-index and metadata caches, and
    \textbf{(c)} Sharded Local Muon.
    Each panel compares its own fixed configuration. Gains from different panels \mbox{cannot be added.}}
    \label{fig:optimization-gains}
\end{figure*}

\subsection{Heterogeneous Model Serving and Robot Execution}
\label{sec:serving}

The reported BEHAVIOR deployment separates high-level reasoning from action generation. The Qwen3.8-27B model used by \metabrain runs in BF16 with tensor parallelism across two GPUs. The complete \wla policy, including its 2B vision-language backbone, world-model components, action expert, and correction adapter, runs in BF16 on a separate GPU. Base and corrected action generation both \mbox{use ten flow-integration steps.}

The simulator and policy service communicate through synchronous WebSocket requests. After executing up to 32 controls from a 50-step chunk, the runtime acquires fresh observations and requests the next chunk. At 30\,Hz, an uninterrupted 32-step prefix spans approximately 1.07 seconds. The policy-call interval also includes inference and communication time. Observation buffers and execution records are cleared at each episode boundary to prevent information from carrying over between episodes.

\subsection{Scalable Simulation Infrastructure for Long-Horizon Benchmarks}
\label{sec:evaluation-acceleration}
\begin{table}[h]
\centering\small
\caption{Evaluation runtime for 100 BEHAVIOR 2026 tasks and 1,000 episodes. The naive eight-GPU baseline is projected as $1{,}275/8\approx159.4$ hours under ideal linear scaling of the estimated serial-worker runtime. The optimized result is measured. GPU counts cover simulation and policy workers, with the reasoning service hosted separately.}
\label{tab:evaluation-speed}
\begin{tabularx}{\linewidth}{@{}p{4.0cm}YY@{}}
\toprule
\textbf{Setting} & \textbf{Naive eight-GPU baseline} & \textbf{Optimized evaluation} \\
\midrule
Worker-pool GPUs & 8 (projected) & 8 Pro6000 GPUs \\
Workers per GPU & 1 & 3 \\
Concurrent workers & 8 & 24 \\
Evaluation episodes & 1,000 & 1,000 \\
Wall-clock duration & $\approx159.4$ hours & $18.1$ hours \\
Episode throughput & $\approx6.27$ per hour & $\approx55.2$ per hour \\
Basis & Ideal parallel projection & Full-suite measurement \\
Relative speedup & $1.0\times$ (reference) & $\approx8.8\times$ \\
\bottomrule
\end{tabularx}
\end{table}

The complete BEHAVIOR 2026 evaluation comprises 100 tasks with ten evaluation instances per task, yielding 1,000 episodes across the benchmark's seven scenes.
The measured run contains 15,113,902 simulator control steps. At 30\,Hz, this is approximately 140 hours of simulated manipulation, averaging 8.40 minutes per episode. This rollout duration differs from the 5.86-minute mean of the training demonstrations reported in \namedref{Section}{sec:wla-data}.
Photorealistic rendering, rigid-body contact dynamics, and neural network inference add substantial computational work to these trajectories. Recorded unoptimized single-worker costs give an estimated serial runtime of 1,275 hours for the suite. We define a naive eight-GPU baseline by running one unchanged worker per GPU and distributing episodes evenly. Assuming ideal linear scaling gives $1{,}275/8=159.375$ hours, reported as 159.4 hours, or approximately 6.27 episodes per hour (\namedref{Table}{tab:evaluation-speed}). This projection assumes unchanged per-worker performance, balanced workloads, and no added startup, scheduling, communication, or shared-service contention. Actual episode lengths vary, so our scheduler allows workers to take on new instances as they finish \mbox{their current tasks.}

\begin{figure}[!t]
\centering
\includegraphics[width=.93\linewidth]{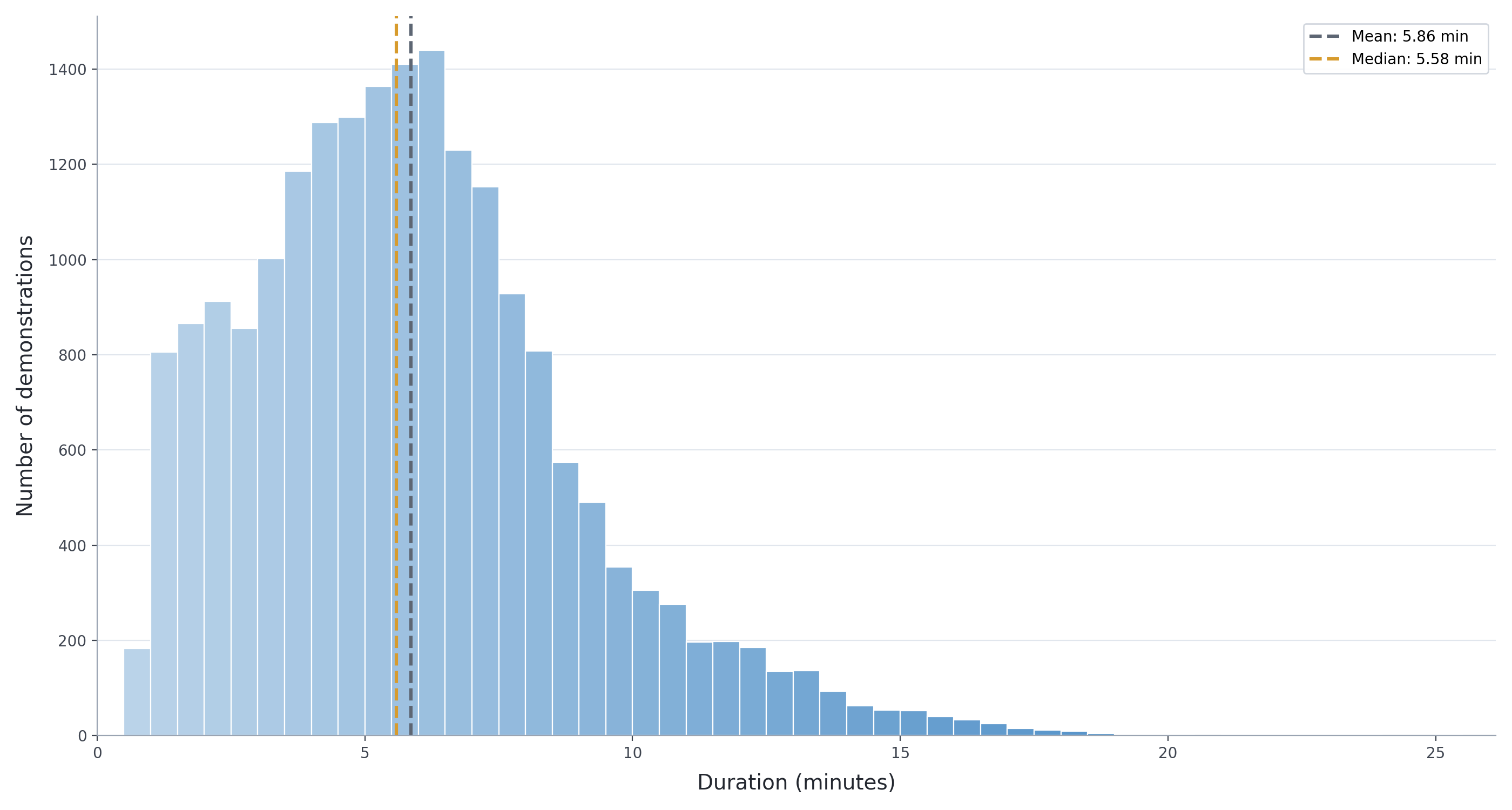}
\caption{Distribution of demonstration durations in the BEHAVIOR 2026 dataset. The dashed lines indicate the mean (5.86 minutes) and median (5.58 minutes). Reproduced from the official dataset statistics~\cite{behavior2026dataset}.}
\label{fig:behavior-demo-duration}
\end{figure}

Our evaluation infrastructure combines cache reuse, faster simulator startup, and coordinated CPU and GPU resource allocation.
These changes reduce repeated work in the evaluation pipeline while keeping task definitions, simulation settings, and model execution logic unchanged.
The optimized configuration supports three concurrent workers per Pro6000 GPU, including simulation and policy inference.
On an eight-GPU node, this arrangement sustains 24 active simulation workers in parallel. 
A dynamic scheduler assigns the next task to each available worker, reducing idle time \mbox{between unequal-length episodes.}

The optimized evaluation node completes the 1,000 episodes in 18.1 hours, corresponding to approximately 55.2 episodes per hour and an aggregate throughput of $7.7\times$ simulated real time. The eight-GPU allocation in \namedref{Table}{tab:evaluation-speed} describes the simulation and policy workers. The separately hosted reasoning service is described in \namedref{Section}{sec:serving}, so this node allocation should not be read as the total hardware requirement of every full-system deployment.
Relative to the projected naive eight-GPU runtime, the measured reduction is $(1{,}275/8)/18.1\approx8.8\times$. This comparison includes cache reuse, startup optimization, dynamic scheduling, and the increase from one to three workers per GPU. It removes the direct one-versus-eight GPU-count difference, but remains conditional on the reference per-worker performance and ideal-scaling assumptions. A measured baseline on the same GPU model and model-serving configuration is still needed to isolate these contributions experimentally.
The distributed scheduler further supports multi-node deployment, supporting up to 96 concurrent simulator workers across a four-machine cluster. 
This infrastructure makes a complete evaluation practical within a day under the measured configuration, supporting more frequent model and system comparisons.

\setcounter{topnumber}{4}
\setcounter{bottomnumber}{2}
\setcounter{totalnumber}{6}
\renewcommand{\topfraction}{0.95}
\renewcommand{\bottomfraction}{0.85}
\renewcommand{\textfraction}{0.05}
\renewcommand{\floatpagefraction}{0.85}
\setlength{\textfloatsep}{10pt plus 2pt minus 2pt}
\setlength{\floatsep}{10pt plus 2pt minus 2pt}
\setlength{\intextsep}{10pt plus 2pt minus 2pt}
\newcolumntype{R}{>{\raggedright\arraybackslash}X}
\newcolumntype{C}{>{\centering\arraybackslash}X}
\newcolumntype{P}[1]{>{\raggedright\arraybackslash}p{#1}}
\newcolumntype{N}[1]{>{\centering\arraybackslash}p{#1}}
\Needspace{8\baselineskip}

\section{Experimental Evaluation}

\label{sec:evaluation}

We evaluate \system across long-horizon household activities, bimanual manipulation, and physical robot tasks. The experiments examine how model specialization and agent coordination affect task completion, whether the learned action policy transfers across scene variations, and how the complete system executes sequential manipulation in the real world.

\subsection{Experimental Setup}

BEHAVIOR 2026 evaluates long-horizon whole-body control over 100 household tasks spanning multiple scenes. The official dataset contains 20,000 human demonstrations with a mean duration of 5.86 minutes (351.54 seconds)~\cite{behavior2026dataset}. The distribution has a long tail, with some demonstrations lasting much longer than the mean (\namedref{Figure}{fig:behavior-demo-duration}). Our public-set analysis uses records selected from evaluation batches, with ten instances per task and 1,000 episodes per configuration. The single-model baseline uses one \wla checkpoint across all tasks. The multi-model baseline selects an action model and execution configuration for each task. Building on this action capability, \system coordinates \metabrain, grounding agents, and \wla through learned correction codes that connect task reasoning and spatial evidence to continuous action generation. Official hidden-set results are reported separately in \namedref{Section}{sec:public-results}.


The reference action interface uses six frames per view, 23-dimensional actions, and a 50-step prediction horizon, with the first 32 steps executed before the next observation. Integration steps and executed prefixes follow the corresponding configuration. All model parameters remain fixed at test time, and human takeover, timeouts, and exhausted episode budgets count as failures.

We report task-macro Q-score, complete-task success rate (SR), and successful task coverage. Q-score measures the fraction of goal conditions satisfied, while SR requires completion of the entire task. Let $Q_i$ and $S_i$ denote the mean goal-satisfaction score and success fraction for task $i$. Over $K$ tasks,

\begin{equation}
Q=\frac{1}{K}\sum_{i=1}^{K}Q_i,\qquad
\mathrm{SR}=\frac{100}{K}\sum_{i=1}^{K}S_i.
\label{eq:metrics}
\end{equation}

Successful task coverage counts tasks with at least one complete success among their ten evaluation instances. Equal episode counts per task make task-macro and episode-micro success rates coincide in \mbox{these benchmark evaluations.}

\subsection{Long-Horizon Household Manipulation on BEHAVIOR}

\label{sec:behavior}

\subsubsection{Overall Performance}

In our public-set evaluation records (\namedref{Table}{tab:2026}), \system obtains a Q-score of 0.6411 and a complete-task success rate of 45.40\%, with 454 successes in 1,000 recorded episodes and at least one success on 88 tasks. The corresponding multi-model and single-model results achieve 0.6040 and 0.4597 Q-score, respectively. The recorded task coverage increases from 70 to 80 to 88 tasks across the three configurations, while the number of tasks with mean Q-score of at least 0.8 rises from 17 to 30 to 34. These patterns motivate \mbox{the task-level analysis below.}

\begin{table}[!htbp]
\centering\small
\setlength{\tabcolsep}{5pt}
\renewcommand{\arraystretch}{1.08}
\caption{Public-set evaluation results for BEHAVIOR 2026, with 100 tasks and 1,000 recorded episodes per configuration. Coverage counts tasks with at least one complete success.}
\label{tab:2026}
\begin{tabularx}{\linewidth}{@{}P{4.8cm}CCCC@{}}
\toprule
{\bfseries Configuration} & {\bfseries Q-score} & {\bfseries Success (\%)} & {\bfseries Successes} & {\bfseries Coverage} \\
\midrule
\wla single model & 0.4597 & 27.20 & 272/1,000 & 70/100 \\
\wla multiple models & \underline{0.6040} & \underline{40.60} & 406/1,000 & 80/100 \\
\system complete system & {\bfseries 0.6411} & {\bfseries 45.40} & 454/1,000 & 88/100 \\
\bottomrule
\end{tabularx}
\end{table}

\subsubsection{Task-Level Performance and Completion Reliability}

\namedref{Figure}{fig:behavior-task-profile} shows the performance of all 100 tasks in the public-set records. Eleven tasks are completed in every recorded instance, spanning activities such as turning on a radio, cooking hot dogs, installing a modem, and laying tile floors. At the other end of the distribution, eight tasks have Q-scores below 0.2, including pizza preparation and patio cleaning, both with zero goal satisfaction. This spread shows that broad benchmark coverage coexists with sharply concentrated weaknesses.

\begin{figure}[p]
\centering
\includegraphics[width=\linewidth]{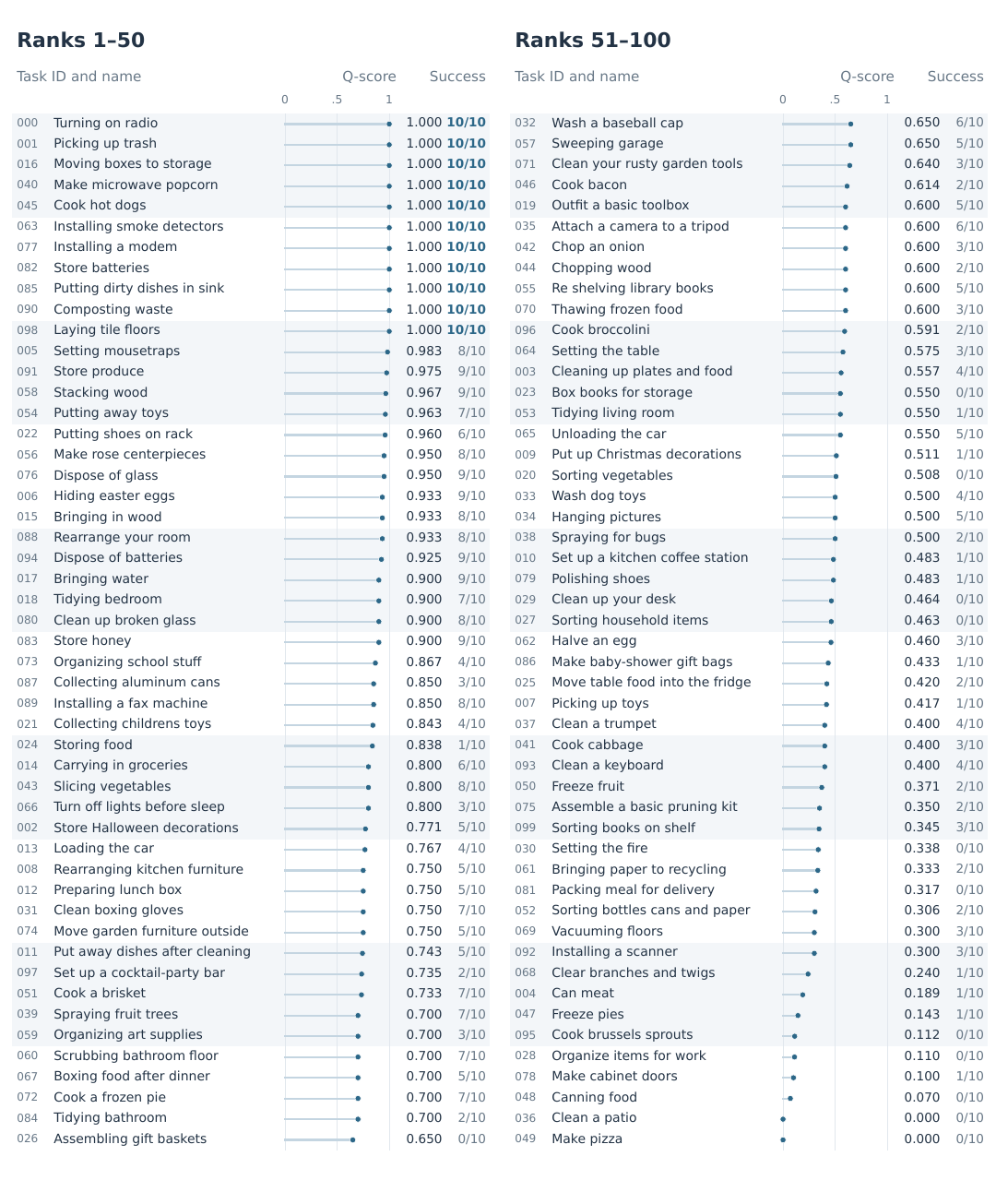}
\caption{Task-level performance in the public-set evaluation records, ordered by decreasing mean Q-score, with ties resolved by task ID. Blue dots show Q-score, with labels rounded to three decimals for display. The final column reports complete successes out of ten recorded instances. The two panels show ranks 1--50 and 51--100. Task names are shortened for display. IDs identify the benchmark tasks.}
\label{fig:behavior-task-profile}
\end{figure}

\begin{table}[!t]
\centering\footnotesize
\setlength{\tabcolsep}{3.5pt}
\renewcommand{\arraystretch}{1.05}
\caption{Public-set results for BEHAVIOR 2026. We include our self-reported result and the 15 highest-scoring entries in the published self-report list, ordered by increasing Q-score. These records follow the participating teams' reported protocols. Institution names follow leaderboard and archived participant records. Scores were accessed 31 October 2026~\cite{behavior2026publicreports,behavior2026leaderboard}.}
\label{tab:behavior2026-comparison}
\begin{tabularx}{\linewidth}{@{}P{3.7cm}Y N{1.45cm} N{1.35cm}@{}}
\toprule
\textbf{Team or submission} & \textbf{Institution} & \textbf{Q-score} & \textbf{SR (\%)} \\
\midrule
George Albrecht & CAU / GMU / EWHA & 0.3862 & 20.00 \\
Mirua & Xiaomi Robotics & 0.3971 & 19.80 \\
VERA & AXXX & 0.4000 & 25.00 \\
Newbie & morphic \& 4dvlab & 0.4040 & 21.70 \\
Huawei AI Algorithm Dept & Huawei Technologies & 0.4076 & 25.20 \\
Lion & Lionrock AI Lab & 0.4082 & 21.80 \\
The Southern Cross & Independent & 0.4099 & 23.00 \\
MXI & Beijing 1011 Technology & 0.4187 & 22.70 \\
Sora & Independent & 0.4195 & 22.50 \\
Multi & mor & 0.4302 & 19.80 \\
Anyverse A2R & Anyverse-A2R & 0.4321 & 24.70 \\
VGI-SKT & SNU / SKT & 0.4449 & 22.30 \\
Robot Learning Collective & Independent & 0.4458 & 27.30 \\
tasman3001 & Ambarella, Inc. & 0.4808 & 30.40 \\
asteroid & CLab & 0.5143 & 30.80 \\
\midrule
\rowcolor{OurMethodBg}
\system (ours) & Anyverse--AutoLab & \textbf{0.6411} & \textbf{45.40} \\
\bottomrule
\end{tabularx}
\end{table}

Performance varies substantially even within familiar activity families. Cooking hot dogs and making microwave popcorn both achieve 100.00\% success, whereas cooking bacon reaches 20.00\% and making pizza has no successful instances. Installation tasks exhibit a similar spread. Smoke-detector and modem installation succeed in all ten instances, while scanner installation succeeds in three of ten.

High goal satisfaction also does not guarantee reliable completion. Of the 34 tasks with Q-score at least 0.8, five have complete-task success rates of 40.00\% or lower. Food storage is a pronounced example, with a Q-score of 0.8375 but only one successful instance out of ten. Moreover, ten of the twelve tasks with no complete successes still achieve positive Q-scores. Gift-basket assembly and boxing books for storage reach 0.6500 and 0.5500, respectively. These results expose a gap between satisfying a substantial subset of goal conditions and consistently satisfying the entire task specification.

The task-wise profile identifies two complementary targets for improvement. For tasks with substantial goal satisfaction but few complete successes, the target is reliable satisfaction of all required conditions within the same episode. For tasks with near-zero scores, the target is to establish effective execution of the activity itself. Reporting Q-score together with complete-success counts makes these distinct performance regimes visible and provides a concrete basis for selecting tasks for further model and system development.

\begingroup
\captionsetup{hypcap=false}
\begin{table}[!t]
\centering\small
\setlength{\tabcolsep}{5pt}
\renewcommand{\arraystretch}{1.08}
\caption{Public-set results on task IDs 0--49 of BEHAVIOR 2026, with 500 recorded episodes per configuration, and BEHAVIOR 2025 public-dataset results grouped by evaluation setting. A dash denotes an unreported success rate.}
\label{tab:2025}
\begin{tabularx}{\linewidth}{@{}P{4.6cm}N{1.8cm}N{1.8cm}R@{}}
\toprule
{\bfseries Method} & {\bfseries Q-score} & {\bfseries SR (\%)} & {\bfseries Evaluation context} \\
\midrule
\multicolumn{4}{@{}l@{}}{\rule[-3pt]{0pt}{15pt}\hspace{4pt}{\bfseries Competition public-dataset evaluation~\cite{behavior2025board}}} \\
\addlinespace[2pt]
RLC~\cite{rlc2025} & 0.2605 & 11.20 & 4 checkpoints \\
Comet~\cite{behavior2025board} & 0.1830 & 14.40 & Competition submission \\
SimpleAI Robot~\cite{behavior2025board} & 0.1943 & 14.00 & Competition submission \\
The North Star~\cite{behavior2025board} & 0.1702 & 12.80 & Competition submission \\
\midrule
\multicolumn{4}{@{}l@{}}{\rule[-3pt]{0pt}{15pt}\hspace{4pt}{\bfseries Post-competition and subsequent reports}} \\
\addlinespace[2pt]
Openpi Comet\cite{comet} & 0.3450 & --- & 2 checkpoints, reinforcement fine-tuning \\
$\pi_{0.5}$\cite{g05} & 0.2626 & --- & 4 epochs, single checkpoint \\
G0.5\cite{g05} & 0.2904 & --- & 1 epoch, single checkpoint \\
G0.5\cite{g05} & 0.3136 & --- & 4 epochs, single checkpoint \\
\midrule
\rowcolor{OurMethodBg}
\multicolumn{4}{@{}l@{}}{\rule[-3pt]{0pt}{17pt}\hspace{4pt}{\bfseries Our 2026 public-set subset, 50 tasks and 500 episodes}} \\
\rowcolor{OurMethodBg}
\wla single model & 0.4522 & 25.60 & 4 epochs, single checkpoint \\
\rowcolor{OurMethodBg}
\wla multiple models & 0.5755 & 37.40 & 25 checkpoints \\
\rowcolor{OurMethodBg}
\system complete system & 0.6191 & 41.40 & Including the full system \\
\bottomrule
\end{tabularx}
\par\vspace{10pt}
\centering\small
\setlength{\tabcolsep}{5pt}
\renewcommand{\arraystretch}{1.08}
\caption{Public references and our results on RoboTwin 2.0. Public entries follow their respective source protocols. Average is the arithmetic mean of Clean and Randomized success rates.}
\label{tab:pending-evaluations}
\begin{tabularx}{\linewidth}{@{}P{5.7cm}CCC@{}}
\toprule
\textbf{Method or configuration} & \textbf{Clean (\%)} & \textbf{Randomized (\%)} & \textbf{Average (\%)} \\
\midrule
$\pi_{0.5}$~\cite{pi05,lingbotvla} & 82.74 & 76.76 & 79.75 \\
LingBot-VLA (without depth)~\cite{lingbotvla} & 86.50 & 85.34 & 85.92 \\
LingBot-VLA (with depth)~\cite{lingbotvla} & 88.56 & 86.68 & 87.62 \\
Motus~\cite{motus} & 88.70 & 87.00 & 87.85 \\
StarVLA~\cite{starvla} & 88.70 & 87.80 & 88.25 \\
LingBot-VA~\cite{lingbotva} & 92.90 & 91.50 & 92.20 \\
Fast-WAM~\cite{fastwam} & 91.90 & 91.80 & 91.85 \\
G0.5~\cite{g05} & 93.70 & 92.80 & 93.25 \\
LingBot-VLA 2.0~\cite{lingbotvla2} & 93.52 & 92.80 & 93.16 \\
MotuBrain~\cite{motubrain} & 95.80 & 96.10 & 95.95 \\
\midrule
\rowcolor{OurMethodBg}
\multicolumn{4}{@{}l@{}}{\textit{Our configurations}} \\
\rowcolor{OurMethodBg}
\wla & 94.26 & 93.70 & 93.98 \\
\rowcolor{OurMethodBg}
\system & \textbf{95.08} & \textbf{95.02} & \textbf{95.05} \\
\bottomrule
\end{tabularx}
\end{table}
\endgroup

\begingroup
\captionsetup{hypcap=false}
\begin{table}[!t]
\centering\small
\setlength{\tabcolsep}{3pt}
\renewcommand{\arraystretch}{1.15}
\caption{Base-policy configurations and results. ``+'' highlights an added component. $\checkmark$ and $\times$ denote enabled and disabled options. ``---'' in the settings columns denotes an unrecorded setting. FAST initialization applies to the vision-language backbone and history encoder. Evaluation conditions and qualifications are given in \namedref{Table}{tab:physical-tasks}.}
\label{tab:physical-short}
\begin{tabularx}{\linewidth}{@{}R N{1.25cm} N{1.65cm} N{1.45cm} N{1.65cm}@{}}
\toprule
\textbf{Configuration} & \textbf{History frames} & \textbf{Latent world model} & \textbf{FAST init.} & \textbf{Macro SR (\%)} \\
\midrule
$\pi_{0.5}$ (baseline) & --- & --- & --- & 26.67 \\
\midrule
\multicolumn{5}{@{}l}{\emph{\wla configurations}} \\
Config. 1 (base fine-tuning) & 2 & $\times$ & $\times$ & 39.58 \\
Config. 2 (+ latent world model) & 2 & $\checkmark$ & $\times$ & 47.08 \\
Config. 3 (+ six-frame history) & 6 & $\checkmark$ & $\times$ & 54.17 \\
\rowcolor{OurMethodBg}
\wla (full configuration) & 6 & $\checkmark$ & $\checkmark$ & 57.92 \\
\bottomrule
\end{tabularx}
\par\vspace{10pt}
\centering\fontsize{8}{10}\selectfont
\setlength{\tabcolsep}{2.5pt}
\renewcommand{\arraystretch}{1.08}
\caption{Per-task successes / attempts. Success is defined as whole-task completion.}
\label{tab:physical-tasks}
\begin{tabularx}{\linewidth}{@{}P{1.2cm}P{3.0cm}R*{5}{N{0.96cm}}@{}}
\toprule
\textbf{Context} & \textbf{Task} & \textbf{Evaluation focus} & $\pi_{0.5}$ & \textbf{Cfg. 1} & \textbf{Cfg. 2} & \textbf{Cfg. 3} & \textbf{Full} \\
\midrule
Study & USB drive storage & Small-object grasping, opening, closing, and storage & 6/20 & 8/20 & 11/20 & 13/20 & 13/20 \\
 & Keyboard pushback with obstacle avoidance & Action sequencing, pushing with obstacle avoidance & 6/20 & 15/20 & 17/20 & 15/20 & 17/20 \\
 & Bookshelf organization & Placement in confined spaces, pose alignment & 4/20 & 7/20 & 9/20 & 12/20 & 13/20 \\

\addlinespace[3pt]
Bedroom & Skincare and cosmetics organization & Pose adjustment, stable placement & 12/20 & 14/20 & 14/20 & 15/20 & 16/20 \\

 & Eye-drop/medicine retrieval and return & Retrieval and placement in confined spaces, delivery and return & 0/20 & 1/20 & 4/20 & 5/20 & 4/20 \\
 & Bedtime item preparation & Multi-object manipulation, action sequencing & 7/20 & 8/20 & 10/20 & 13/20 & 14/20 \\
 & Object retrieval from a drawer & Target selection, retrieval and placement in confined spaces & 2/20 & 4/20 & 5/20 & 8/20 & 9/20 \\
\addlinespace[3pt]
Living room & Watch/wristband and earbud placement for charging & Target alignment, stable placement & 12/20 & 8/20 & 11/20 & 12/20 & 13/20 \\

\addlinespace[3pt]

Kitchen & Food-waste sorting & Transfer by pouring, sorted disposal & 2/20 & 13/20 & 13/20 & 15/20 & 16/20 \\
 & Cucumber preparation & Tool manipulation, cutting contact & 10/20 & 9/20 & 10/20 & 12/20 & 13/20 \\
\addlinespace[3pt]

Entryway & Parcel unpacking & Tool manipulation, cutting and box opening & 1/20 & 3/20 & 3/20 & 5/20 & 5/20 \\
 & Parcel packing and sealing & Ordered packing, sequencing lid closure & 2/20 & 5/20 & 6/20 & 5/20 & 6/20 \\
\bottomrule
\end{tabularx}
\end{table}
\endgroup

\subsubsection{Comparison with Public Results}

\label{sec:public-results}
\namedref{Table}{tab:behavior2026-comparison} compares public-set results. Our public evaluation used a remotely hosted Qwen3.8-27B \metabrain. For the official hidden-set submission, limited bandwidth between the organizers' evaluation site and our server in China required a self-contained Docker deployment. The 24 GB GPU-memory limit forced us to replace the 27B coordinator with Qwen3.5-2B~\cite{qwen35model}. This substantial reduction in coordinator capacity limits the submitted system, so the hidden-set results do not fully represent the capability of the 27B-based system.

On the official hidden set, \system obtains a verified Q-score of 0.4975 and 32.05\% complete-task success~\cite{behavior2026verified}, compared with 0.4951 and 30.95\% for the separate \wla submission. The 2B coordinator thus provides a small gain. In our tests, the 2B coordinator was less accurate than Qwen3.8-27B at deciding when to intervene and sometimes interrupted execution unnecessarily. The reported hidden-set results were obtained with the submitted Docker configuration using the 2B coordinator. We will release the Qwen3.8-27B version of \metabrain for further evaluation and development.

\namedref{Table}{tab:2025} places our public-set results on the 50-task BEHAVIOR 2026 subset alongside methods evaluated on the public dataset of BEHAVIOR 2025, including RLC, Comet, Openpi Comet, $\pi_{0.5}$, and G0.5. The BEHAVIOR 2025 entries report public-dataset tests and are grouped by their reported evaluation settings. On task IDs 0--49, \system achieves 0.6191 Q-score and 41.40\% success across \mbox{500 recorded episodes.}

\subsection{Bimanual Manipulation on RoboTwin 2.0}

RoboTwin 2.0 evaluates 50 bimanual tasks under Clean and Randomized settings~\cite{robotwin2}. The randomized scenes vary background, clutter, lighting, table height, and language. \namedref{Table}{tab:pending-evaluations} summarizes public reference results alongside our configurations, which use \wla as the action policy and add agent-guided correction in the complete system. The \wla configuration completes 100 episodes per task in each setting, for 10,000 episodes in total. Public references retain the evaluation protocols reported by their respective sources. We report complete-task success for both settings, with Average defined as their arithmetic mean.

\wla achieves 94.26\% success in Clean scenes and 93.70\% under randomization, averaging 93.98\%. This is 14.23 percentage points above the reported $\pi_{0.5}$ result of 79.75\% and 0.73 percentage points above G0.5 at 93.25\%. Agent coordination raises the average to 95.05\%, a further 1.07-point gain, while MotuBrain reports 95.95\% under its own protocol. Our simple \wla architecture is thus competitive with leading bimanual policies without requiring a separate large-scale pretraining strategy tailored to this benchmark. At the same time, success rates near 95.00\% leave limited room to distinguish systems on these short tasks. We include RoboTwin as a reference for local manipulation competence, while the longer BEHAVIOR and real-robot tasks more directly test sustained coordination and recovery.

\Needspace{6\baselineskip}
\subsection{Real-World Manipulation}

\subsubsection{Short-Horizon Tasks}

\begin{figure}[!t] \centering \includegraphics[width=\linewidth,height=0.85\textheight,keepaspectratio]{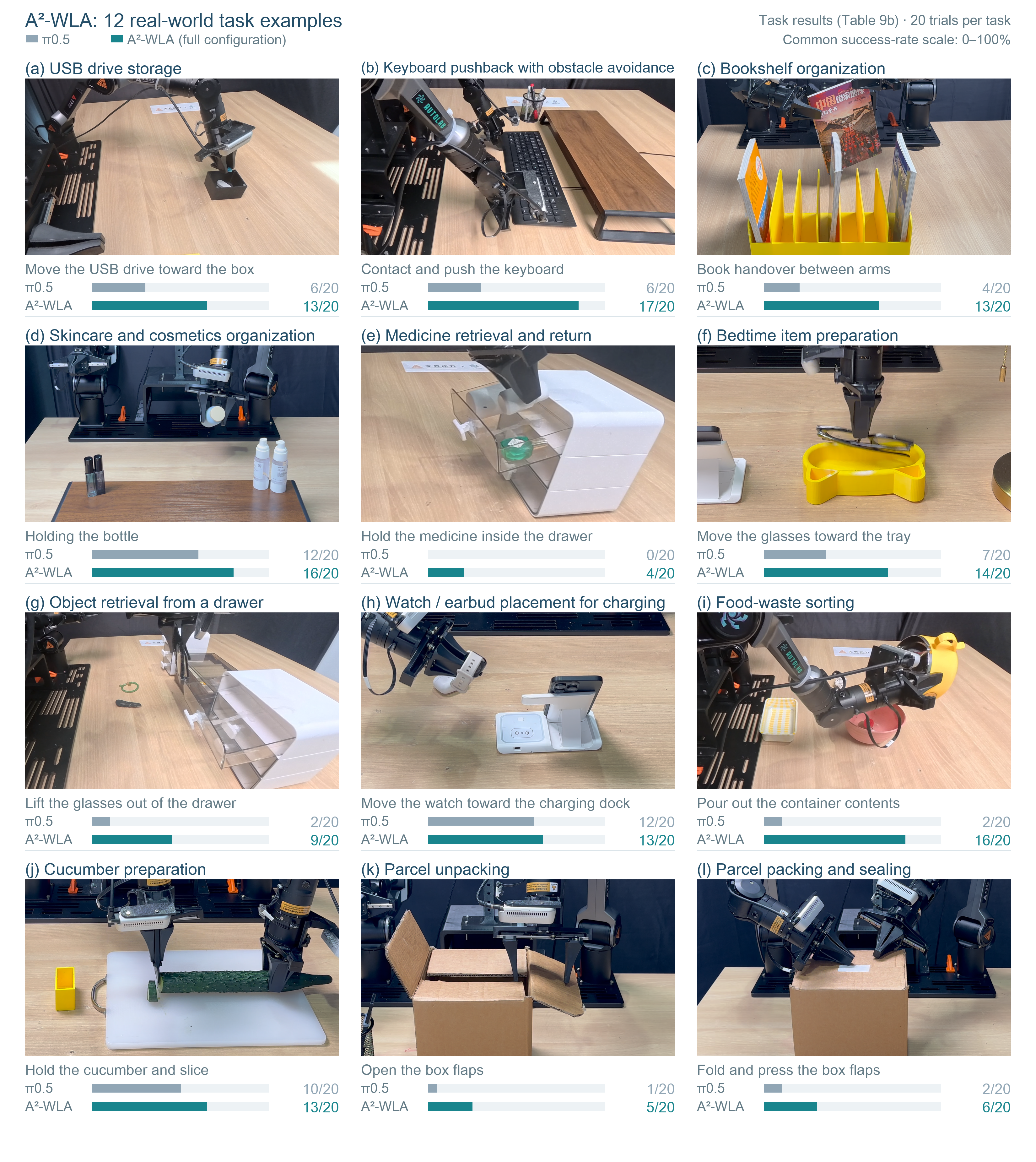} \caption{Execution examples and results across 12 real-world tasks. Panels follow the task order in \namedref{Table}{tab:physical-tasks} and show keyframes from \wla demonstration videos, labeled with the visible actions. The bars below each frame compare successes / attempts for $\pi_{0.5}$ and the full \wla configuration, with 20 trials per task and a common success-rate scale of 0--100\%.} \label{fig:real-robot-comparison} \end{figure}

We evaluate 12 tabletop tasks across five household settings, comprising \emph{the study, bedroom, living room, kitchen, and entryway}. The tasks include storage and retrieval in confined spaces, pushing with obstacle avoidance, food preparation, and parcel handling, testing grasping, placement, action sequencing, and contact control under varied object sizes, spatial constraints, and contact conditions. We use $\pi_{0.5}$ as the principal generalist baseline because of its strong manipulation performance across diverse real-world environments~\cite{pi05}, and compare it with four fine-tuned \wla configurations. For a fair real-robot comparison, high-level feedback and test-time correction are disabled.

Each task uses 200 training trajectories, and a separate policy is trained for each task and configuration for 60,000 steps from the specified initialization. The \wla configurations include base fine-tuning, the multiview latent world model, a six-frame history window, and FAST action-token initialization (\namedref{Table}{tab:physical-short}). For the full configuration, the vision-language backbone and motion-aware history encoder are initialized from weights jointly trained with FAST on 12 tasks for 60,000 steps, while the action expert is initialized from pretrained action-policy weights. The FAST classification loss is disabled during subsequent \wla training. \namedref{Sections}{sec:wla} and~\ref{sec:base-design} describe the model components and simulation ablations. These configurations use different complete training recipes. We evaluate each task and configuration over 20 trials and compute macro success as the arithmetic mean of the per-task success rates across all 12 tasks (\namedref{Table}{tab:physical-tasks}).

\namedref{Table}{tab:physical-short} summarizes the policy configurations and their macro success rates across the 12 tasks. The full \wla configuration achieves 57.92\% macro success, compared with 39.58\% for base fine-tuning and 26.67\% for $\pi_{0.5}$. The full configuration exceeds base fine-tuning by 18.33 percentage points, computed from the underlying counts.

The per-task results in \namedref{Table}{tab:physical-tasks} show that the full configuration outperforms $\pi_{0.5}$ on all 12 tasks, with gains spanning several types of manipulation. Food-waste sorting improves from 2/20 to 16/20 successful trials, keyboard pushback from 6/20 to 17/20, and bookshelf organization from 4/20 to 13/20. These tasks involve pouring, pushing around obstacles, and placing objects in confined spaces, respectively. The improvements therefore extend beyond object pickup to operations that require different action sequences and contact conditions. \namedref{Figure}{fig:real-robot-comparison} illustrates these execution sequences and compares the two policies on each task.

The gains are nevertheless uneven. The full configuration completes only 4/20 eye-drop/medicine retrieval-and-return trials, 5/20 parcel-unpacking trials, and 6/20 parcel-packing trials. Its result on medicine retrieval is also below the 5/20 of Configuration 3, so the aggregate improvement does not imply a gain on every task from every component. These remaining difficulties include small-object localization, retrieval in confined spaces, tool use, and ordered closure operations. Because these experiments disable high-level feedback and test-time correction, they measure the learned action policy's ability to complete each task unaided. We next examine recovery during a longer coffee-preparation task.

\FloatBarrier

\begin{figure}[!ht]
\centering
\includegraphics[width=\linewidth]{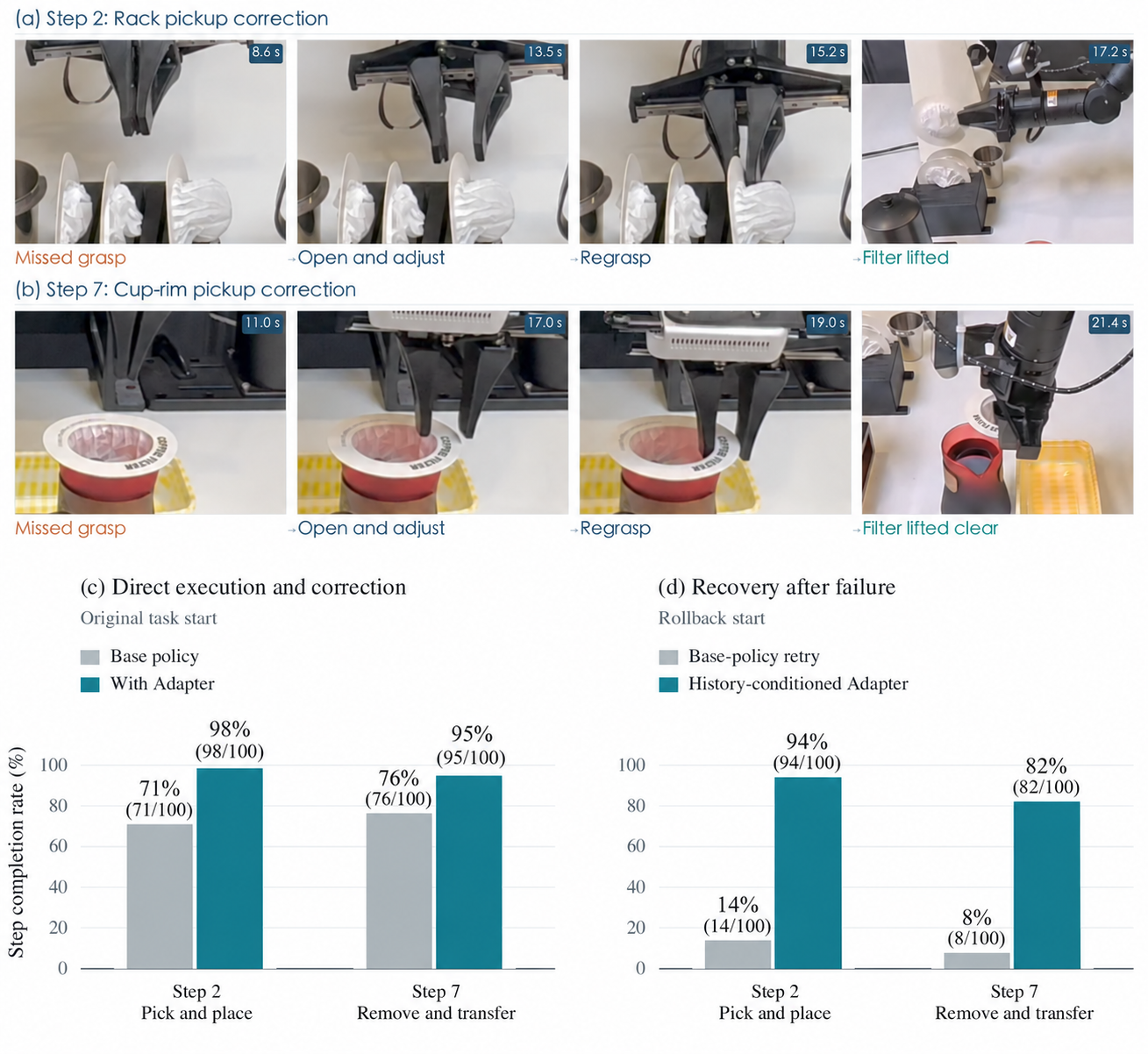}

\caption{Grasp correction and step completion during pour-over coffee preparation. (a) Rack pickup in Step 2. (b) cup-rim pickup in Step 7. Each row shows a missed grasp, gripper opening and adjustment, regrasping, and lifting. Timestamps are playback positions in the respective clips. Stages are labeled from visible actions. (c) Base-policy execution with and without the Adapter from the original task start. (d) base-policy retry and the failure-history-conditioned Adapter after rollback. Each measured condition has 100 trials per step. Measured bars show completions / attempts.}
\label{fig:filter-recovery}
\end{figure}

\subsubsection{Long-Horizon Pour-Over Coffee Preparation}
\label{sec:deployment}

Pour-over coffee preparation requires the robot to handle objects, position a filter, and complete a sequence of operations under changing contact and support conditions. In deployment, \system completes the drink in approximately four minutes per run, with more than one hundred cups completed across deployment. To examine recovery within this longer activity, we evaluate two steps where a failed grasp can interrupt progress. Step 2 retrieves the filter from the rack and places it stably on the cup rim. Step 7 removes the filter from the cup and transfers it to a designated container.

For each step, we evaluate four experimental conditions with 100 trials per condition: direct base-policy execution, base-policy execution with the Adapter, base-policy retry, and Adapter recovery. The first two conditions start from the original task starting point, while the latter two start after a failed grasp and rollback. A trial succeeds when the robot completes the placement or transfer, releases the filter, and withdraws the gripper. Step 2 also requires stable support on the cup rim. Step 7 with a used filter requires \mbox{removal of the grounds.}

During recovery, the runtime pauses pending actions after an anomaly and updates the images and robot state. Using execution history and diagnostic evidence, \metabrain proposes an adjustment, and \wla generates corrective actions under Adapter modulation. After a short execution segment, the system observes the scene again and checks the recovery condition before deciding whether to continue. For a missed filter grasp, the correction aims to adjust the grasp position, regrasp, and lift the filter clear. Subsequent step completion measures whether execution can continue (\namedref{Figure}{fig:filter-recovery}).

\begin{table}[!t]
\centering\small
\setlength{\tabcolsep}{3pt}
\renewcommand{\arraystretch}{1.14}
\caption{Base-model ablations on Selected12, averaged over three evaluation rounds of 120 episodes each. The upper block compares observation inputs and initialization. The lower block adds model and data components cumulatively to the action-pretrained baseline. Check marks denote enabled components and dashes denote their absence. History-input configurations retain conventional history encoding when motion-aware encoding is disabled.}
\label{tab:base-design}
\begin{tabularx}{\linewidth}{@{}Ycccc rr@{}}
\toprule
\textbf{Setting} & \multicolumn{2}{c}{\textbf{Model components}} & \multicolumn{2}{c}{\textbf{Training data}} & \multicolumn{2}{c}{\textbf{Performance}} \\
\cmidrule(lr){2-3}\cmidrule(lr){4-5}\cmidrule(l){6-7}
 & \shortstack{World\\model} & \shortstack{Motion\\encoding} & \shortstack{Skill\\sampling} & \shortstack{Successful\\rollouts} & Q-score & SR (\%) \\
\midrule
\multicolumn{7}{@{}l}{\textit{Observation and initialization baselines}} \\
Current frame only & -- & -- & -- & -- & 0.1582 & 8.33 \\
Six-frame history & -- & -- & -- & -- & 0.2167 & 13.33 \\
FAST pretraining & -- & -- & -- & -- & 0.3453 & 20.83 \\
Action-policy initialization & -- & -- & -- & -- & 0.4846 & 39.72 \\
\midrule
\multicolumn{7}{@{}l}{\textit{Cumulative additions to action-policy initialization}} \\
Configuration A & $\checkmark$ & -- & -- & -- & 0.5183 & 42.50 \\
Configuration B & $\checkmark$ & $\checkmark$ & -- & -- & 0.5279 & 45.56 \\
Configuration C & $\checkmark$ & $\checkmark$ & $\checkmark$ & -- & 0.5546 & 46.67 \\
\rowcolor{OurMethodBg}
\textbf{Full \wla} & $\checkmark$ & $\checkmark$ & $\checkmark$ & $\checkmark$ & \textbf{0.5905} & \textbf{50.28} \\
\bottomrule
\end{tabularx}
\end{table}

The real-robot experiments use a failure-history-conditioned Adapter. It takes current observations and up to 64 historical state transitions, reads the history through cross-attention, and adds a gated residual before the action output head. This variant conditions its residual on observations and failure history. The full interface in \namedref{Figure}{fig:correction} instead encodes all four CorrectionContext inputs and injects its residual at an intermediate action layer. The real-robot recovery results evaluate this history-conditioned Adapter. We freeze the base model and train approximately 860,000 Adapter parameters with a flow-matching objective, using measured state sequences from correction segments as supervision. The gate is initialized to zero, and all parameters remain fixed at deployment. After rollback, the Adapter completes Step 2 in 94/100 trials (94.00\%) and Step 7 in 82/100 trials (82.00\%), while base-policy retry succeeds in 14/100 (14.00\%) and \mbox{8/100 (8.00\%), respectively.}

%


\Needspace{8\baselineskip}
\begin{revision}
\section{Ablation Studies and Long-Horizon Execution Analysis}
\label{sec:ablations}

We analyze how action-model design and grounded correction contribute to long-horizon manipulation, then compare full-task conditioning with subtask scheduling and examine whether local recovery leads to continued progress and task completion. To make repeated training and evaluation feasible, we use two 12-task subsets with different purposes. Selected12 covers object transport, storage, cleaning, cooking, and assembly, and supports the action-model and task-conditioning studies. A separate Adapter subset tests how corrective context changes execution, including button pressing in turning on a radio, object alignment in hanging pictures and attaching a camera to a tripod, and successive object-handling operations in bringing water and washing a baseball cap. These activities require control to account for target-relative position, object orientation, and the current manipulation stage, allowing us to examine the contributions of corrective language, spatial grounding, and execution history. Adapter comparisons keep the same action model frozen to hold its underlying control capability fixed, and a disjoint set of instances tests whether correction gains extend beyond the instances used in the ablations. Each evaluation round covers ten instances per task, or 120 episodes, and reports task-macro Q-score and complete-task success. The full 100-task system evaluation is reported separately in \namedref{Section}{sec:evaluation}.


\Needspace{6\baselineskip}
\subsection{Base Action Model Ablations}
\label{sec:base-design}

\namedref{Table}{tab:base-design} shows that the full \wla achieves a Q-score of 0.5905 and a complete-task success rate of 50.28\% on Selected12, exceeding the action-pretrained baseline by 0.1059 and 10.56 percentage points, respectively. All variants use a single policy without high-level correction or the correction Adapter.

\Needspace{3\baselineskip}
\noindent\topicmarker{} \textit{Action Pretraining and Initialization.} The pretraining stages in \namedref{Section}{sec:pretraining} introduce action supervision before continuous-policy optimization. We compare how much of this learned capability is transferred into the Selected12 policy. A six-frame model initialized from a general-purpose VLM reaches 0.2167 Q-score and 13.33\% success. FAST initialization first teaches the backbone to predict discrete action tokens, raising these scores to 0.3453 and 20.83\%. Action-policy initialization additionally transfers the action expert together with the vision-language backbone pretrained on all 100 tasks, reaching 0.4846 and 39.72\% after downstream training. We therefore use action-policy initialization as the starting point for \mbox{all subsequent cumulative ablations.}

\noindent\topicmarker{} \textit{Temporal and Predictive Context.} \wla uses visual history to interpret recent interaction and a latent world model to anticipate scene changes (\namedref{Sections}{sec:visual-history} and~\ref{sec:worldmodel}). We first test whether the historical observations themselves help. With the same weights trained on six-frame inputs, replacing the current frame alone with the full history raises Q-score from 0.1582 to 0.2167 and success from 8.33\% to 13.33\%. This measures sensitivity to history removal at inference, including the change from the model's six-frame training input. We then add predictive and motion representations to the action-pretrained baseline. Conditioning the action expert on predicted future features raises Q-score from 0.4846 to 0.5183 and success from 39.72\% to 42.50\% (A). Adding the motion-aware history encoder over the same six-frame window further improves these results to 0.5279 and 45.56\% (B). Thus, retaining past observations and learning explicit representations of motion and future change provide complementary gains.

\noindent\topicmarker{} \textit{Training-Data Composition.} The sampling design in \namedref{Section}{sec:wla-data} gives additional weight to skill transitions, where the policy must change its behavior as an operation ends. With the architecture and auxiliary objectives fixed, we jointly apply skill-boundary oversampling and primitive balancing. This raises Q-score from 0.5279 to 0.5546 and success from 45.56\% to 46.67\% (C), using the same trajectories and adaptive flow-sampling settings. We then broaden the training experience by perturbing initial robot and target-object positions within fixed training-scene layouts and retaining rollouts with terminal Q-score equal to one. Mixing these 2,400 successful trajectories, 200 per task, with the expert demonstrations brings the full model to 0.5905 Q-score and 50.28\% success. The final comparison measures the combined effect of position perturbation, success filtering, and additional data. Together, the results show gains from both redistributing training toward skill transitions and expanding the successful execution examples. We next examine what task specialization reveals about the design of generalist action models.

\Needspace{5\baselineskip}
\subsection{Task Specialization and Model Design}
\label{sec:model-routing}

A generalist action model must share knowledge across activities while retaining the control precision needed for specific interactions. In our development experiments, a model trained specifically on a 12-task subset outperformed the shared model trained on all 100 tasks when evaluated on that subset. This observation raises a model-design question about how to preserve the benefits of broad training without diluting the representations needed for particular scene layouts, object interactions, and action sequences.

We examine one form of specialization using a bank of \wla policies trained on task subsets. Prior evaluations determine the task-to-model mapping and execution configuration, both fixed before the reported evaluation. Each episode uses one selected policy conditioned on the complete task instruction. Across all 100 tasks, this configuration reaches 0.6040 Q-score and 40.60\% success, compared with 0.4597 and 27.20\% for the shared model (\namedref{Section}{sec:evaluation}). The gains combine specialization, checkpoint selection, and execution configuration.

These results motivate architectures that combine shared representations with specialized control capacity. One direction is to share the visual, temporal, and world-model backbone while allowing smaller action experts to adapt to different interaction patterns. A learned router conditioned on observations and task language could choose the appropriate action expert without a fixed mapping from benchmark task IDs. Another direction is to distill the specialized policies into one model, retaining their complementary skills while simplifying deployment. These possibilities remain to be tested under matched parameter and compute budgets. Grounded correction is complementary to this architectural question, and the following experiments isolate its contribution by keeping \wla fixed.

\subsection{Grounded Action Correction}
\label{sec:grounded-correction-analysis}

The correction experiments use the 12-task Adapter evaluation subset described above and keep the \wla weights fixed. The first ten instances per task form the condition-ablation set, and the last ten form a disjoint generalization set, each containing 120 episodes. Missing episode results contribute zero Q-score and count as failures. We first examine the information supplied to the Adapter, then compare correction interfaces \mbox{and test additional instances.}

\subsubsection{Contribution of Correction Conditions}
\label{sec:adapter-condition-ablation}

The condition encoder combines correction language $u_t$, spatial evidence $G_t$, executed history $R_t$, and the current base proposal $\hat A_t^{\mathrm{base}}$ into $c_t$ (\equationref{eq:code}). We remove these inputs to evaluate their contribution while retaining the same action model and upstream reasoning modules. The language--grounding variant removes correction language, task text, and structured grounding together. It is therefore a joint semantic ablation rather than an isolated test of the recovery instruction. The history variant retains only the latest state--action record. \namedref{Table}{tab:adapter-condition-ablation} reports the resulting task performance.

\begin{table}[!htbp]
\centering\small
\setlength{\tabcolsep}{4pt}
\renewcommand{\arraystretch}{1.12}
\caption{Adapter condition ablation on the 12-task correction subset, with 120 episodes per configuration. All variants share \mbox{the \wla weights.}}
\label{tab:adapter-condition-ablation}
\begin{tabularx}{\linewidth}{@{}P{4.8cm}R N{1.25cm}N{1.55cm}@{}}
\toprule
\textbf{Model} & \textbf{Condition change} & \textbf{Mean Q} & \textbf{Success (\%)} \\
\midrule
\wla (w/o Adapter) & No Adapter & 0.5310 & 41.67 \\
Adapter w/o Language \& Grounding & Remove correction language, task text and structured grounding & 0.5344 & 43.33 \\
Adapter w/o Long History & Retain only the latest state--action record & 0.5356 & 44.17 \\
Adapter w/o Candidate & Remove base-action candidate encoding & 0.5542 & 47.50 \\
Adapter w/o Grounding & Remove structured grounding input & 0.5570 & 49.17 \\
\rowcolor{OurMethodBg}
\textbf{Full Adapter} & All conditions retained & \textbf{0.6108} & \textbf{52.50} \\
\bottomrule
\end{tabularx}
\end{table}

The full Adapter achieves 0.6108 Q-score and 52.50\% success, improving over the no-Adapter baseline by 0.0798 and 10.83 percentage points. Joint removal of language and grounding reduces performance to 0.5344 and 43.33\%. Restricting history to one record yields 0.5356 and 44.17\%. Both ablations bring performance close to the base policy, supporting the contribution of semantic--spatial context and extended \mbox{execution history in this evaluation.}

Removing the base candidate lowers Q-score by 0.0566 and success by 5.00 percentage points. Removing structured grounding lowers them by 0.0538 and 3.33 percentage points. Every tested deletion reduces performance relative to the complete condition set. The candidate anchors the adjustment to the action currently proposed by the policy, while grounding locates the correction in the physical scene.

\begin{table}[!t]
\centering\small
\setlength{\tabcolsep}{5pt}
\renewcommand{\arraystretch}{1.15}
\caption{Recovery and continuation with three correction interfaces on the 12-task Adapter evaluation subset.}
\label{tab:controlled-correction-interfaces}
\begin{tabularx}{\linewidth}{@{}P{5.1cm}CCCC@{}}
\toprule
{\bfseries Interface}
& {\bfseries Q-score}
& {\bfseries Physical recovery rate}
& {\bfseries Effective continuation rate}
& {\bfseries Final task completion rate} \\
\midrule
Instruction restatement \mbox{(frozen \wla)}
& 0.5668 & 46.27\% & 25.37\% & 38.81\% \\

End-effector geometric correction (VLM + IK)
& 0.6083 & 52.05\% & 28.77\% & 45.21\% \\

\rowcolor{OurMethodBg}
Latent-space residual adapter
& 0.6108 & 54.93\% & 33.80\% & 43.66\% \\
\bottomrule
\end{tabularx}
\end{table}

\subsubsection{Comparison of Correction Interfaces}

We compare instruction restatement, end-effector geometric correction, and the latent residual Adapter on the 12-task Adapter evaluation subset (\namedref{Table}{tab:controlled-correction-interfaces}). Each interface uses high-level advice, spatial evidence, executed history, and a fresh base proposal. Trainable correction components use the same correction examples and source-trajectory split. These comparisons evaluate different intervention mechanisms; the interfaces are not matched for parameter count or computational cost. Physical recovery measures whether the required physical condition is re-established. Effective continuation additionally requires the next predefined meaningful task advance within a fixed window, while final completion records whether the original task goal is achieved within the remaining episode budget. For each method, the three recovery rates use episodes with at least one correction trigger, counting each episode once from its first trigger and including cases where no correction is executed. Q-score uses all 120 evaluation episodes. Triggered-episode cohorts can differ across interfaces. Effective continuation is measured within a fixed window, whereas final completion uses the remaining episode budget, so final completion need not be lower than the time-limited continuation rate.

\begin{figure}[!b]
\centering
\includegraphics[width=\linewidth]{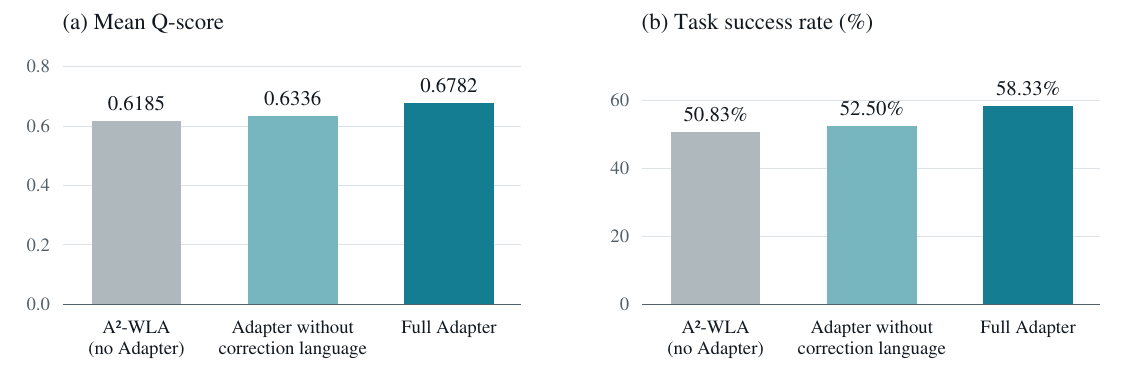}
\caption{Adapter generalization on the last ten instances of each of the 12 correction tasks, with 120 episodes per configuration. Both axes start at zero.}
\label{fig:adapter-instance-generalization}
\end{figure}

Compared with VLM+IK, the Adapter achieves a higher physical recovery rate, 54.93\% versus 52.05\%, and a slightly higher Q-score, 0.6108 versus 0.6083. Its final completion rate among triggered episodes is 43.66\%, compared with 45.21\% for VLM+IK. These rates describe different triggered-episode cohorts, so they are not paired comparisons on identical failure states. VLM+IK applies geometric corrections through inverse kinematics, whereas the Adapter injects a correction code into the action expert's hidden states during trajectory generation. Conditioned on the recovery instruction, scene evidence, execution history, and base proposal, the Adapter can coordinate the adjustment with the policy's learned whole-body behavior across the action chunk. The observed improvement in physical recovery therefore does not translate into higher final task \mbox{completion in this comparison.}



The condition ablations vary the evidence supplied to one Adapter, whereas the interface comparison varies how the same corrective evidence changes behavior. The execution cases in \namedref{Section}{sec:recovery-analysis} illustrate the distinction between local recovery and completing the original task.

\begingroup
\captionsetup{hypcap=false}
\begin{figure}[!t]
\centering
\includegraphics[width=0.94\linewidth]{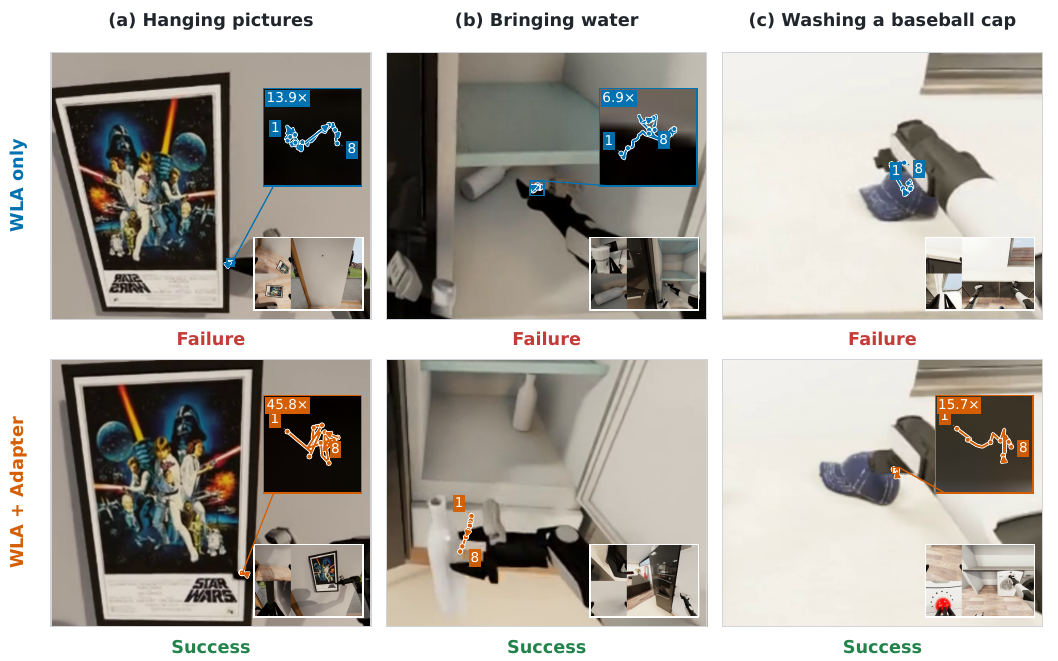}
\caption{Local correction and complete-episode outcomes for hanging pictures, bringing water, and washing a baseball cap. Blue trajectories show \wla{}-only execution. Orange trajectories show Adapter-controlled intervals. Small photographic insets show subsequent observations from failed \wla{}-only rollouts in the upper row and terminal observations from successful Adapter-enabled rollouts in the lower row. Colored boxes magnify the local trajectories. Each pair comprises independently executed rollouts on the same task instance.}
\label{fig:adapter-qualitative}
\end{figure}
\endgroup

\subsubsection{Instance Generalization and Execution Cases}
\label{sec:adapter-instance-generalization}

On the disjoint set of ten additional instances per task, the full Adapter reaches 0.6782 Q-score and 58.33\% success, compared with 0.6185 and 50.83\% for the base policy (\namedref{Figure}{fig:adapter-instance-generalization}). The gains of 0.0597 and 7.50 percentage points correspond to nine additional completed episodes, increasing the success count from 61 to 70. The improvement therefore holds on both disjoint instance sets of the 12 correction tasks.

The correction-language ablation retains task text and structured grounding but removes corrective instructions and direction. It obtains 0.6336 Q-score and 52.50\% success, below the full Adapter by 0.0446 and 5.83 percentage points. These results support the contribution of correction language alongside spatial evidence and history on additional instances of the same tasks.

\namedref{Figure}{fig:adapter-qualitative} relates the quantitative gains to hanging pictures, bringing water, and washing a baseball cap. Each column uses the same task instance and base-action checkpoint, with ten flow-integration steps and six input frames. Blue trajectories show \wla{}-only motion, and orange trajectories show Adapter-controlled intervals. The outcome insets show subsequent observations and terminal task completion. In hanging pictures, the base rollout drops the frame after approaching the wall, while the Adapter-enabled rollout completes hanging. In bringing water, the correction occurs during bottle acquisition and is followed by task completion. In cap washing, repeated baseline grasp attempts end in a dropped cap, whereas the corrected rollout finishes the task. Across these examples, local alignment and object acquisition form the link between corrective intervention and continued execution.

We next compare execution with explicit subtask prompts and \mbox{different scheduling strategies.}

\begin{table}[!t]
\centering\small
\setlength{\tabcolsep}{3.5pt}
\renewcommand{\arraystretch}{1.08}
\caption{Complete-task conditioning (B) and VLM-selected Focus (F) on Selected12, both trained for 180k steps. Each configuration reports one round of ten instances per task. The baseline results are from its first evaluation round.}
\label{tab:focus-preliminary}
\begin{tabularx}{\linewidth}{@{}lYrrrr@{}}
\toprule
{\bfseries Task} & {\bfseries Task name} & {\bfseries B: Q} & {\bfseries F: Q} & {\bfseries B: success} & {\bfseries F: success} \\
\midrule
000 & Turn on radio & 0.7000 & 0.7000 & 7/10 & 7/10 \\
001 & Pick up trash & 0.6000 & 0.4667 & 5/10 & 3/10 \\
006 & Hide Easter eggs & 0.3667 & 0.3778 & 3/10 & 3/10 \\
017 & Bring water & 0.8333 & 0.3000 & 8/10 & 3/10 \\
018 & Tidy bedroom & 0.7000 & 0.7333 & 5/10 & 3/10 \\
022 & Put shoes on rack & 0.8500 & 0.7900 & 7/10 & 5/10 \\
030 & Set the fire & 0.5000 & 0.4750 & 2/10 & 0/10 \\
032 & Wash baseball caps & 0.1000 & 0.1000 & 0/10 & 0/10 \\
034 & Hang pictures & 0.3000 & 0.0000 & 3/10 & 0/10 \\
035 & Attach camera to tripod & 0.1000 & 0.0000 & 1/10 & 0/10 \\
040 & Make microwave popcorn & 0.5000 & 0.6000 & 5/10 & 6/10 \\
045 & Cook hot dogs & 0.9500 & 0.7500 & 9/10 & 7/10 \\
\midrule
\rowcolor{OurMethodBg}
\multicolumn{2}{l}{\bfseries Overall (task-macro Q)} & \textbf{0.5417} & \textbf{0.4411} & \textbf{55/120} & \textbf{37/120} \\
\bottomrule
\end{tabularx}
\end{table}

\begin{figure}[!b]
\centering
\includegraphics[width=\linewidth]{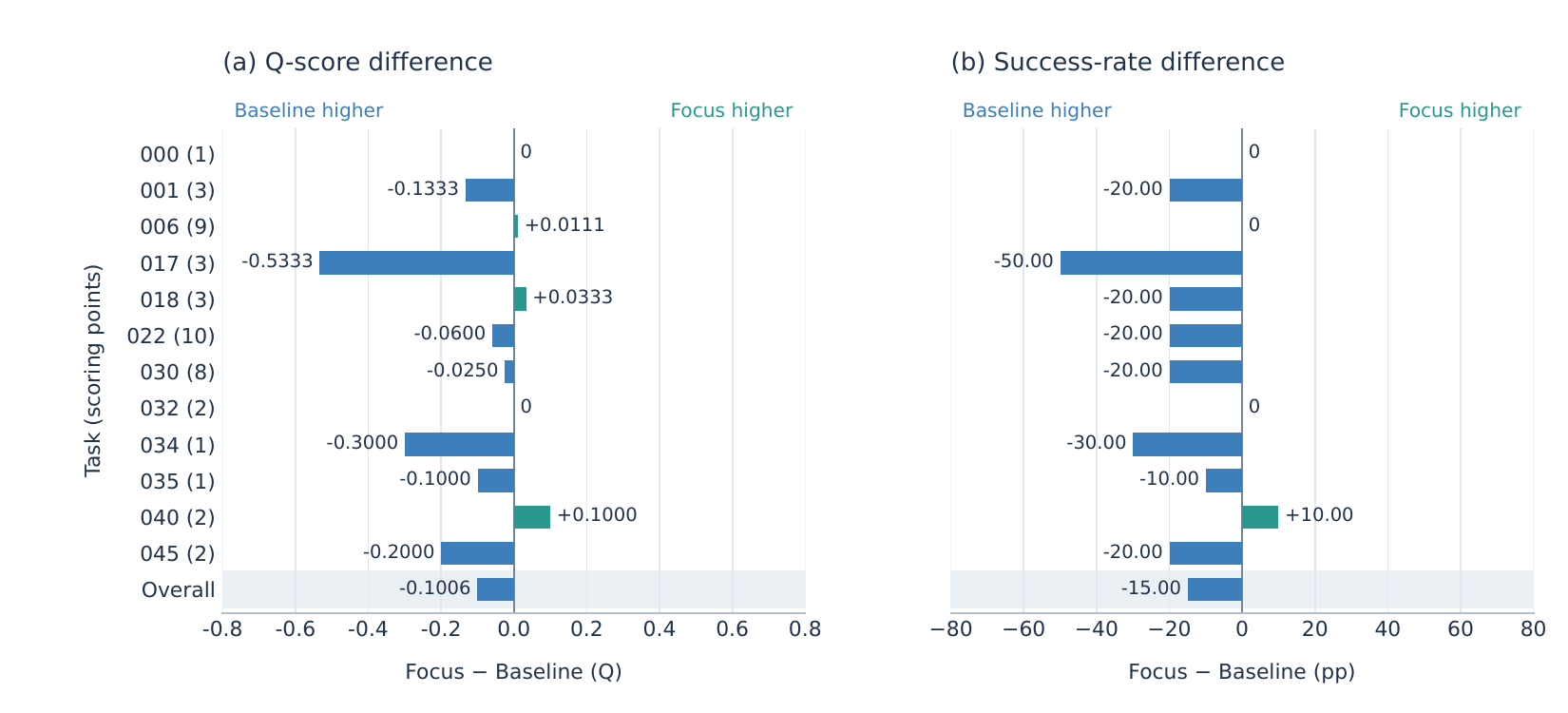}
\caption{Task-level changes from complete-task conditioning to Focus on Selected12. The left panel shows Q-score differences and the right panel shows success-rate differences in percentage points.}
\label{fig:focus-preliminary}
\end{figure}

\subsection{Full-Task Conditioning and Subtask Scheduling}
\label{sec:instruction-routes}

\subsubsection{Full-Task Conditioning versus Focus}
\label{sec:focus-preliminary}

\namedref{Table}{tab:focus-preliminary} compares two training-and-execution configurations on Selected12. Both use the same pretrained initialization and demonstrations from the same 12 tasks. The baseline row is the first round of the action-pretrained configuration in \namedref{Table}{tab:base-design}, whose reported result averages three rounds. The single-round value here therefore differs from that three-round mean. The baseline uses the complete task instruction. Focus retains that goal and adds a \texttt{Current focus:} instruction selected by an asynchronous VLM.


On Selected12, the baseline achieves a task-macro Q-score of 0.5417 and a complete-task success rate of 45.83\%, while Focus reaches 0.4411 and 30.83\%, respectively. The corresponding differences are 0.1006 Q-score and 15.00 percentage points in favor of the baseline. As shown in \namedref{Figure}{fig:focus-preliminary}, this aggregate gap masks variation across tasks. Focus improves Q-score on three tasks, decreases it on seven, and ties on two. The largest gain is 0.1000 on making microwave popcorn, whereas the largest declines are 0.5333 on bringing water, 0.3000 on hanging pictures, and 0.2000 on cooking hot dogs. The task-specific gains therefore do not translate into an overall improvement for this Focus configuration.

The comparison evaluates the two training-and-execution configurations as a whole. To examine scheduling itself, the following control fixes the action model across all modes.

\subsubsection{Scheduling with a Fixed Action Model}
\label{sec:focus-scheduling-protocol}

We use the same frozen Focus action model to compare four scheduling modes on the same 120 episodes (\namedref{Table}{tab:focus-scheduling-protocol}). All modes use the same history-window length, action-prefix length, and task budgets. Focus dropout during training supports input without a Focus prompt in F-Global. F-Linear and F-Adaptive use the same simulator-assisted completion and executability tests. The former enforces a fixed order, while the latter \mbox{permits reordering and revisitation.}
\begin{table}[!htbp]
\centering\small
\setlength{\tabcolsep}{4pt}
\caption{Scheduling ablation with one frozen Focus action model and 120 scheduled episodes per mode. Q-score is reported on a 0--1 scale. Missing or failed evaluation runs are assigned zero Q-score and counted as unsuccessful. Valid incomplete trajectories retain their measured partial Q-score.}
\label{tab:focus-scheduling-protocol}
\begin{tabularx}{\linewidth}{@{}lRCC@{}}
\toprule
\textbf{Mode} & \textbf{Conditioning and scheduling} & \textbf{Q-score} & \textbf{SR (\%)} \\
\midrule
F-VLM & Full task and online VLM-selected Focus & 0.4411 & 30.83 \\
F-Global & Full task only & 0.4201 & 30.00 \\
F-Linear & Simulator-assisted assessment, fixed order & 0.4115 & 25.00 \\
F-Adaptive & Same assessment, reordering and revisitation & 0.3785 & 27.50 \\
\bottomrule
\end{tabularx}
\end{table}


F-VLM achieves the highest Q-score and complete-task success rate among the four modes, reaching 0.4411 and 30.83\%, respectively. Compared with F-Global, it gains 0.0210 Q-score and completes one additional episode out of 120, corresponding to a 0.83 percentage-point difference in success rate. This comparison evaluates explicit Focus conditioning together with online scheduling. The simulator-assisted modes show a different pattern. Under the same completion and executability tests, F-Adaptive achieves a higher success rate than F-Linear, 27.50\% versus 25.00\%, but a lower Q-score, 0.3785 versus 0.4115. Thus, flexible ordering and revisitation do not improve both measures in this diagnostic comparison.

\subsubsection{Task Transitions and Demonstration Coverage}
\label{sec:transition-analysis}

\begingroup
\captionsetup{hypcap=false}
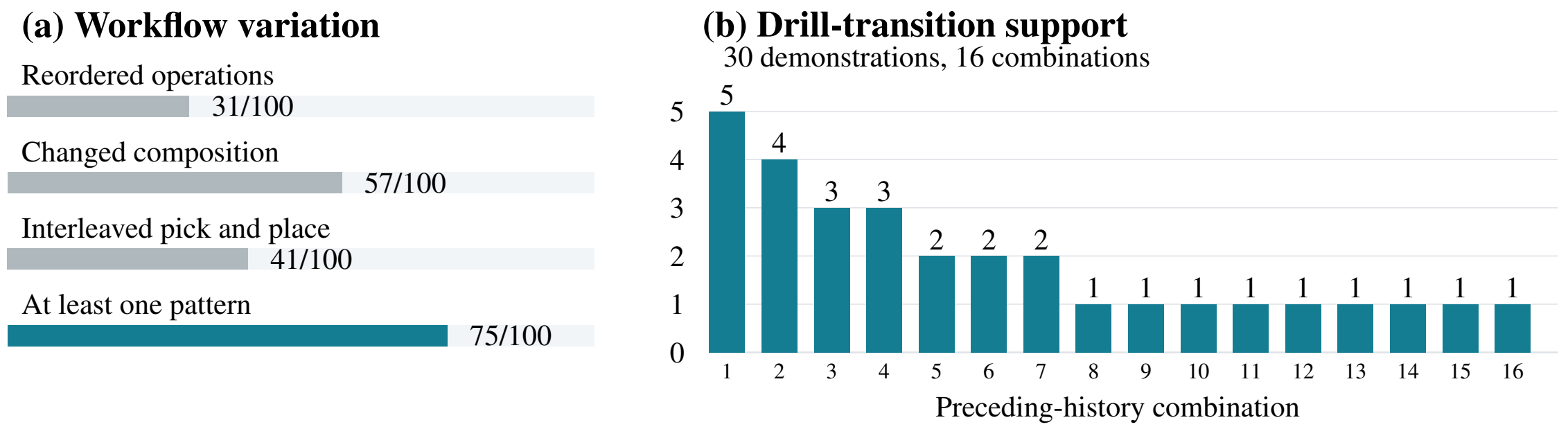
\begin{figure}[!t]
\centering
\begingroup
\definecolor{ChartGray}{HTML}{AFB8BD}
\definecolor{ChartTeal}{HTML}{147D92}
\definecolor{ChartGrid}{HTML}{DEE3E7}
\begin{tikzpicture}[x=1cm,y=1cm,every node/.style={font=\scriptsize\rmfamily,text=black}]
\node[anchor=west,font=\small\rmfamily\bfseries] at (0,3.1) {(a) Workflow variation};
\foreach \y/\n/\label in {2.35/31/Reordered operations,1.62/57/Changed composition,0.89/41/Interleaved pick and place,0.16/75/At least one pattern} {
 \node[anchor=west] at (0,\y+.27) {\label};
 \fill[Apale] (0,\y-.1) rectangle (5.6,\y+.1);
 \ifnum\n=75\relax\def\BarColor{ChartTeal}\else\def\BarColor{ChartGray}\fi
 \fill[\BarColor] (0,\y-.1) rectangle ({5.6*\n/100},\y+.1);
 \node[anchor=west] at ({5.6*\n/100+.08},\y) {\n/100};
}
\node[anchor=west,font=\small\rmfamily\bfseries] at (6.5,3.1) {(b) Drill-transition support};
\draw[ChartGrid] (6.7,0) -- (14.8,0);
\foreach \y in {0,1,2,3,4,5} {
 \draw[ChartGrid,line width=.3pt] (6.7,{\y*.46}) -- (14.8,{\y*.46});
 \node[anchor=east] at (6.6,{\y*.46}) {\y};
}
\foreach \n [count=\i] in {5,4,3,3,2,2,2,1,1,1,1,1,1,1,1,1} {
 \fill[ChartTeal] ({6.7+(\i-1)*.5},0) rectangle ({7.04+(\i-1)*.5},{\n*.46});
 \node at ({6.87+(\i-1)*.5},{\n*.46+.16}) {\n};
 \node[font=\tiny\rmfamily] at ({6.87+(\i-1)*.5},-.18) {\i};
}
\node at (10.6,-.55) {Preceding-history combination};
\node[anchor=west] at (6.7,2.8) {30 demonstrations, 16 combinations};
\end{tikzpicture}

\endgroup
\caption{Workflow variation and sparse transition histories. Common variation covers 75 of 100 tasks. The 30 drill-placement demonstrations cover 16 preceding-history combinations, with \mbox{1--5 examples per combination.}}
\label{fig:workflow-coverage}
\end{figure}
\endgroup

The scheduling comparisons highlight the need to determine whether the physical state at the end of one operation is suitable for the next. Retrieving a bottle establishes possession, but pouring or placement also requires an appropriate grasp and orientation. Closing a door depends on whether retrieval is complete and a hand is available. A direct handoff therefore requires both completion of the preceding operation and satisfaction of the next operation's physical preconditions. The relevant state includes the objects and contacts in the scene, beyond the robot proprioception denoted by $s_t$.

This dependence on preceding interactions is also visible in demonstration coverage. We analyze 20,000 original demonstrations across 100 BEHAVIOR tasks, with 200 demonstrations per task. The analysis uses the original \texttt{skill\_annotation}, removes navigation, normalizes object categories, and counts \mbox{support by independent demonstration.} Workflow variation appears even within a single task. For toolbox organization, 161 demonstrations open the lid before collecting tools. The other 39 pick up and replace a tool before opening the lid. Fixing tool categories, actions, and repetition counts still leaves 67 distinct orders among 126 demonstrations. In art-supply organization, 199 demonstrations with the same four items and pick-and-place counts contain 70 workflows, including interleaved operations on different objects.

These variations change the history preceding an otherwise identical transition. Picking up a drill and placing it in the toolbox occurs in 30 demonstrations. Distinguishing the two preceding operations divides these examples into 16 history combinations, each supported by only one to five demonstrations. Histories involving a screwdriver and then the drill have five examples, while flashlight-and-drill histories have four. The next operation therefore encounters different preceding actions and held-object states. Across all 100 tasks, reordered identical operations occur in 31 tasks, changes in operation composition or repetition in 57, and interleaved picking and placing in 41. Their union covers 75 tasks (\namedref{Figure}{fig:workflow-coverage}). The first two categories require at least ten demonstrations of each alternative. Interleaving requires at least ten demonstrations containing pick A, pick B, place A, place B. Categories overlap.

Across 97 tasks with sufficient history and the same 203,245 operation windows, task-wise counting yields 2,851 adjacent transition combinations. Conditioning on the two preceding operations increases this number to 6,849. Of these history-conditioned combinations, 62.8\% have support from at most five demonstrations, while accounting for only 3.7\% of window occurrences. Repeated individual operations coexist with sparse coverage of the long tail of preceding contexts. These statistics characterize variation in workflow and coverage of preceding contexts. Together with the physical handoff requirements, they identify the unfinished goals, available support and grasp relations, and required physical states that a scheduler must represent.

\begin{figure}[!t]
\centering
\begin{minipage}{0.80\linewidth}\begingroup
\setlength{\tabcolsep}{2pt}
\noindent{\small\sffamily\bfseries a\quad Grip adaptation and handover; episode ends in timeout}
\par\smallskip
\begin{tabular}{@{}ccc@{}}
\includegraphics[width=.326\linewidth]{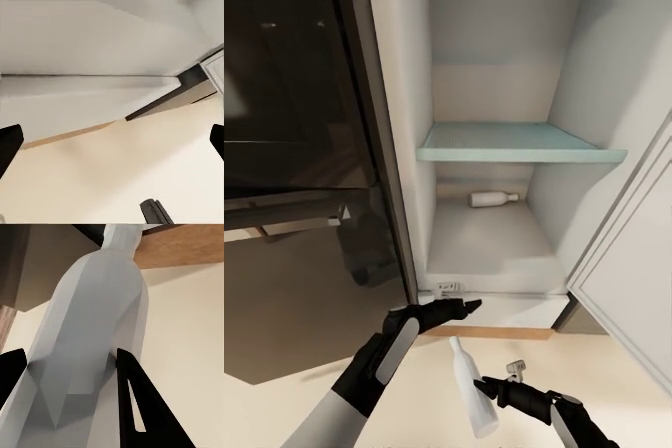} &
\includegraphics[width=.326\linewidth]{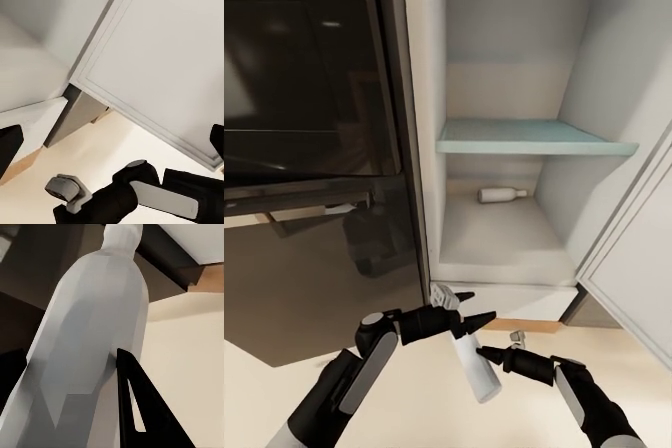} &
\includegraphics[width=.326\linewidth]{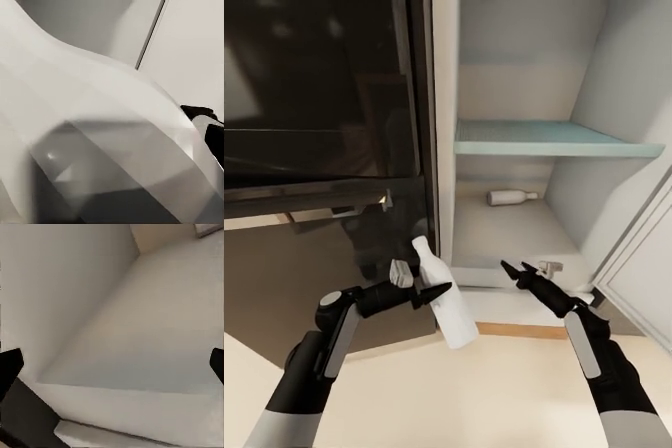} \\
\scriptsize 03:50 & \scriptsize 04:40 & \scriptsize 05:00
\end{tabular}
\par\medskip
\noindent{\small\sffamily\bfseries b\quad Corn recovery in an episode ending in timeout}
\par\smallskip
\begin{tabular}{@{}ccc@{}}
\includegraphics[width=.326\linewidth]{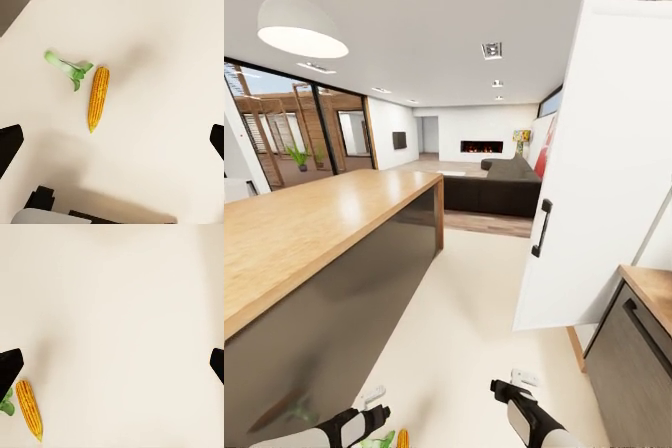} &
\includegraphics[width=.326\linewidth]{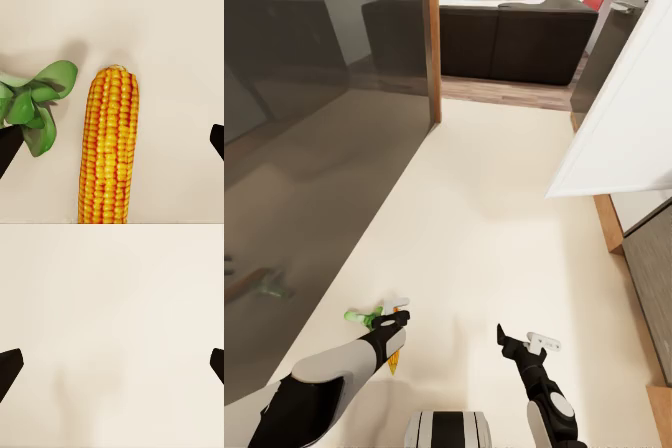} &
\includegraphics[width=.326\linewidth]{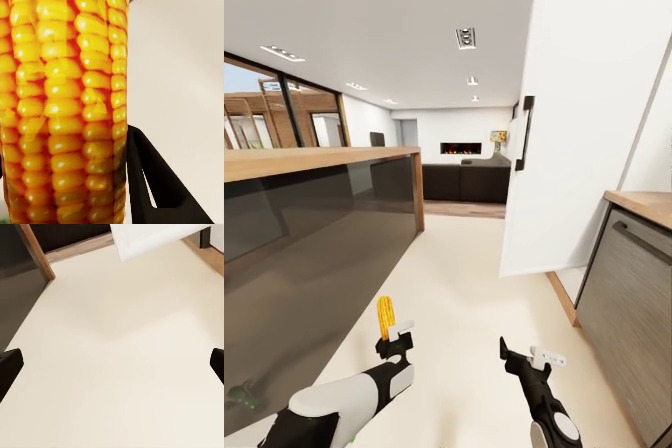} \\
\scriptsize 09:00 & \scriptsize 09:10 & \scriptsize 09:20
\end{tabular}
\par\medskip
\noindent{\small\sffamily\bfseries c\quad Glove recovery followed by task completion}
\par\smallskip
\begin{tabular}{@{}ccc@{}}
\includegraphics[width=.326\linewidth]{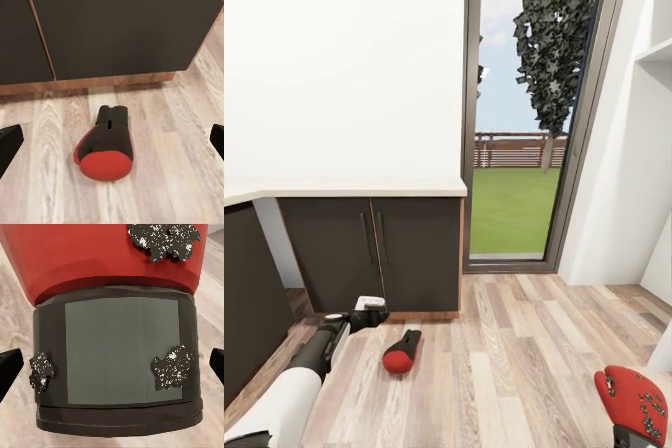} &
\includegraphics[width=.326\linewidth]{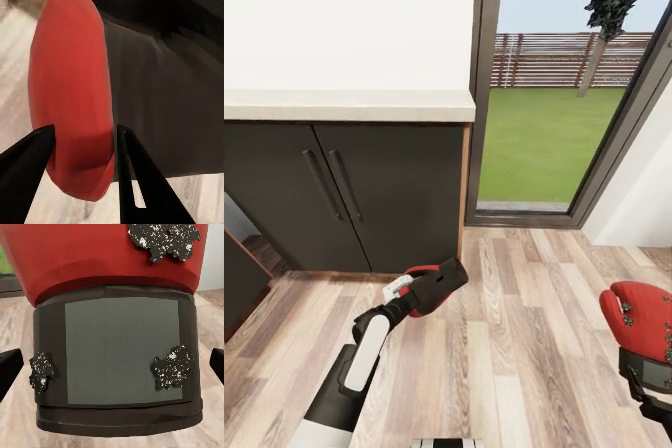} &
\includegraphics[width=.326\linewidth]{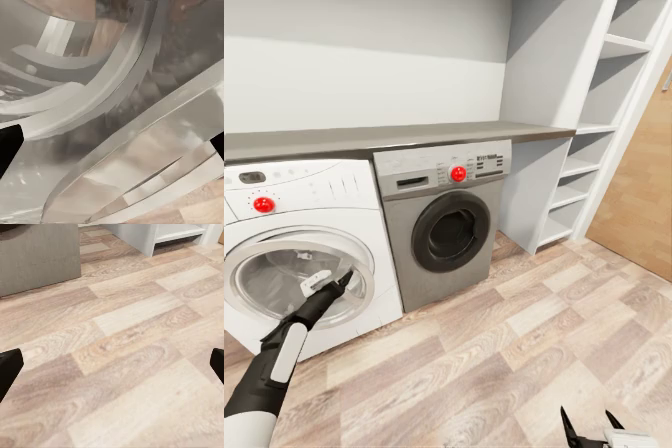} \\
\scriptsize 02:45 & \scriptsize 03:35 & \scriptsize 04:35
\end{tabular}
\par\medskip
\endgroup
\end{minipage}
\caption{Base-policy recovery without the Adapter, beyond the corresponding task demonstrations. Bottle reorientation and corn regrasping occur in episodes that ultimately time out after prolonged retries. Glove retrieval is followed by continued washing and task completion. Each row follows one episode chronologically.}
\label{fig:episodes}
\end{figure}

\subsection{Qualitative Analysis of Cross-Task Skill Transfer and Task Continuation}
\label{sec:recovery-analysis}

Following joint training across multiple long-horizon tasks, \wla autonomously reuses and adapts learned manipulation skills in execution states not covered by the corresponding task demonstrations. These behaviors arise from the base policy without the Adapter. We examine execution cases to assess how this capability supports local recovery and whether recovery enables the robot to resume and complete the original task.

\subsubsection{Emergent Skill Transfer in the Base Policy}
\label{sec:qualitative}

\namedref{Figure}{fig:episodes} presents three representative cases. The boxing-glove washing task provides the clearest example. While carrying a glove toward the washing machine, the robot accidentally drops it, bends down to retrieve it, and resumes the washing sequence, ultimately completing the task. Although the demonstrations do not include recovery from such a drop, the policy adapts learned object-retrieval and transport skills to the unexpected state and incorporates the recovery into the remaining sequence.

Similar adaptations occur in the other two tasks. During water fetching, the robot adjusts its grasp on an inverted bottle and completes a handover. During vegetable sorting, it regrasps a fallen ear of corn and resumes transport. These cases suggest that manipulation skills acquired through joint training can be reused and adapted across tasks, enabling autonomous recovery in states not covered by the demonstrations. However, this qualitative evidence does not identify the specific training tasks from which these skills originate.

Recovery capability alone does not ensure efficient correction. Without targeted guidance, the base policy may spend extended periods on repeated attempts. In the water-fetching and vegetable-sorting episodes, prolonged retries exhaust the available time budget, resulting in timeout failures despite the observed adaptive behaviors. These cases motivate targeted guidance during recovery. Within \system, \metabrain combines foundation-model reasoning with current observations and execution history to generate corrective instructions. The Adapter translates this guidance into action adjustments, aiming to reduce unproductive retries and help the policy resume progress toward the original goal within the remaining episode budget.

\subsubsection{Physical State Changes and Embodiment Constraints}
\label{sec:embodiment-constraints}

Failures can lead to prolonged retries and directly disrupt the scene conditions required for task execution (\namedref{Figure}{fig:physical-state-changes}). Examples include a cart tipping over, a lamp falling into a gap between furniture, and food pieces scattering into the sink. The robot must then re-establish suitable grasping, support, or contact conditions. However, hardware and workspace constraints may place the required recovery actions beyond the robot's reach, joint range of motion, or gripper capabilities. Such scene disruptions can therefore prevent task continuation when recovery demands exceed the robot's physical capabilities.

These constraints define the scope of reasoning, diagnosis, and action correction in \system. Reliable long-horizon embodied autonomy requires learned policies, system orchestration, motion control, and hardware design to be developed and validated as an integrated system.

\begin{figure}[!t]
\centering
\includegraphics[width=.98\linewidth,height=.80\textheight,keepaspectratio]{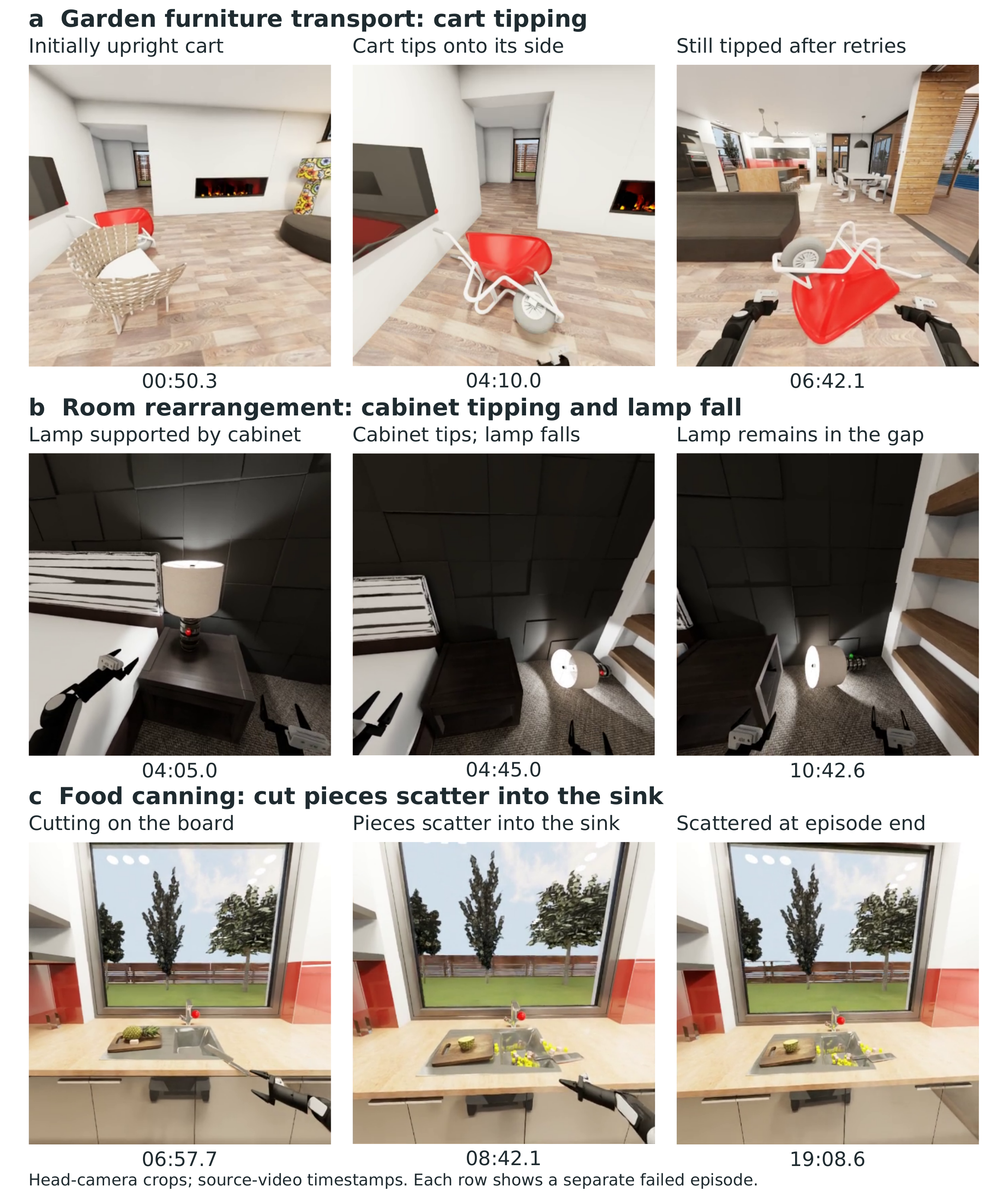}
\caption{Physical state changes during failed task execution. (a) During garden furniture transport, a cart tips over and subsequent manipulation fails to right it. (b) During room rearrangement, a bedside cabinet tips over and the lamp falls into a gap between furniture. (c) During food canning, cut food pieces scatter into the sink and remain scattered until the episode ends. Columns show the initial conditions, state changes, and subsequent observations within the same episode. Each row represents an independent failed episode. Timestamps indicate playback positions in the source videos; images show cropped head-camera views.}
\label{fig:physical-state-changes}
\end{figure}

The base-model study evaluates a cumulative representation and training recipe, and the fixed-policy Adapter experiments show gains from corrective context on two instance sets. Scheduling comparisons and demonstration statistics expose the demands of task transitions. The execution cases illustrate both local recovery by the base policy and its limits. Together, these studies distinguish improvements in action generation, intervention, and continued progress toward the task goal.

\end{revision}

\Needspace{6\baselineskip}

\begin{revision}

\section{Discussion and Conclusion}

\label{sec:discussion}

We treat long-horizon execution as a capability to be learned by the action model, with the agent system providing verification and correction at critical points. The experiments in \namedref{Sections}{sec:evaluation} and~\ref{sec:ablations} examine this division of roles through task performance, action-model design, and grounded correction.

\subsection{Simple Action Models and Room for Improvement}

A striking result is how far a relatively simple action architecture can go. \wla augments a vision-language-action policy with motion-aware history and a latent world model that predicts future visual features. Before adding agent-driven correction, its task-specialized configuration reaches 0.5755 Q-score on the 50-task BEHAVIOR 2026 public subset, approximately $2.21\times$ the 0.2605 public score of RLC, the 2025 competition winner (\namedref{Table}{tab:2025}).  Even the shared \wla policy reaches 0.4522. RLC also uses explicit stage-transition logic and recovery heuristics~\cite{rlc2025}. The strong performance of \wla before agent-driven correction therefore motivates learning more of the required behavior within the action model itself, while retaining the comparison's limits.

The useful lesson is to strengthen a small set of complementary capabilities by retaining recent interaction history, predicting how the scene will change, and translating that context into continuous control. The component study on Selected12 supports this direction. Adding the latent world model to the action-pretrained baseline raises Q-score from 0.4846 to 0.5183, while motion-aware history and the subsequent data improvements bring the full recipe to 0.5905. The remaining gap between partial progress and complete-task success indicates considerable room to improve temporal representations, predictive objectives, demonstration coverage, and training efficiency within this compact design. Better action policies would also give the multi-agent framework a stronger foundation for targeted correction and reliable task continuation.

\subsection{Learning Long-Horizon Capability in the Action Model}

Long-horizon tasks involve nonlinear composition because the feasibility of a later operation depends on the physical state and interaction history produced by earlier operations. A bottle grasp that completes retrieval may be unsuitable for pouring, and retrieving one object can change hand availability for the next operation. As analyzed in \namedref{Section}{sec:transition-analysis}, independently successful subtasks do not automatically compose into a successful activity. Their ordering and transition conditions must be represented by the policy, \mbox{the scheduler, or both.}

This motivates training the action policy directly on long-horizon task trajectories, rather than relying on a scheduler to assemble independently learned subtasks. The policy can learn recurring dependencies between operations from these trajectories. Its online decisions still depend on a finite observation window, so long-horizon training does not give it complete memory of the episode. The shared execution record and task-level reasoning supply complementary context. The sparse coverage of preceding contexts in the demonstration analysis further emphasizes the need for training data that preserve cross-step dependencies and cover different ways of reaching a task transition.

Stronger foundation models offer another source of knowledge for action-policy learning. In our implementation, \metabrain generates corrective instructions that train the correction encoder and adapter, while the original \wla parameters remain frozen. The current results evaluate the learned correction interface. A further direction is to distill broader task reasoning and successful corrected executions into the main action-policy weights, which could reduce the need for repeated deliberation.

The agent system supports this learned capability through verification and correction at critical execution points. It checks progress and physical feasibility, diagnoses deviations, and supplies grounded corrective context through the Adapter. The action model continues to generate the actions needed for the remaining activity. The fixed-policy Adapter experiments demonstrate the value of this complementary role. System-level feedback adjusts execution while the underlying model weights remain unchanged. Recovery is useful when the corrected state allows the policy to continue the unfinished activity.

The 100-task analysis identifies where this division of roles must improve reliability. The system succeeds at least once on 88 tasks, but only 11 tasks succeed in all ten evaluated instances. High Q-scores also coexist with low complete-task success, as in food storage. Further development therefore requires both broader long-horizon competence in the action model and effective verification and correction at points where execution departs from the task goal.

\subsection{Limitations and Future Directions}

The current correction process pauses execution for diagnosis, which limits its use for events that demand immediate reactions. We prioritize scaling and diversifying interaction data so that the action model learns how contact states evolve and how to respond to deviations from visual observations and execution history. Training trajectories should cover varied object properties, contact configurations, and both successful and failed attempts, including recovery and subsequent task continuation. Together with faster interaction between reasoning and control, this data-driven approach aims to strengthen timely responses within the learned policy.

Correction training uses deviations generated at demonstration states, whereas deployed policies encounter failures that alter object arrangements and grasp configurations. Collecting these execution states and their subsequent outcomes would extend supervision to the situations in which correction is actually needed. This calls for a data pipeline that links failure detection, corrective instructions, and the success or failure \mbox{of the remaining task.}

Current quantitative evaluations focus on household simulation. Additional Adapter tests examine cross-task generalization, while real-robot studies cover short-horizon manipulation and sequential coffee preparation. Future work will expand coverage to more task families, robot embodiments, and physical environments, and assess the consistency of performance gains across independent training and evaluation runs. Evaluation will continue to report both goal satisfaction and complete-task success to assess how improvements in individual operations contribute to reliable end-to-end execution.

\subsection{Conclusion}

We present \system, an embodied multi-agent framework that connects \metabrain, grounding agents, and \wla through \emph{code as a scaling interface}. The learned correction encoder and adapter absorb \metabrain-generated guidance during training and convert fresh corrective intent and spatial evidence into action adjustments while keeping model parameters fixed during deployment. On 100 BEHAVIOR 2026 tasks, the system achieves a Q-score of 0.6411 and a complete-task success rate of 45.40\%, with at least one successful episode on 88 tasks. The accompanying training and evaluation infrastructure supports iteration across this broad task set, and coffee preparation tests a history-conditioned recovery variant on a real robot. Our approach combines long-horizon capability learned within the action model with system-level verification and correction at critical points. Further progress depends on strengthening both the learned continuity of execution and the effectiveness of targeted intervention. We will release models, training scripts, and evaluation tools to support \mbox{research on this approach.}

\end{revision}

\FloatBarrier
\section*{Data availability}
The BEHAVIOR-1K challenge demonstrations used in this work are publicly available from the official benchmark project\cite{behavior2026dataset}. Correction supervision is constructed from these demonstrations using the procedure in \namedref{Section}{sec:correction-training}. The planned code release includes scripts for constructing correction supervision and training the models on these data. The source dataset remains available through its official distribution.
\section*{Code availability}
We will progressively open-source the complete \system solution at \url{https://github.com/AutoLab-SAI-SJTU/A2MEAS}. The planned releases cover multi-agent orchestration, the \wla model and trained checkpoints, training scripts for the action model and correction interface, data-processing code for the official BEHAVIOR demonstrations, and the optimized training, execution and evaluation infrastructure. Each release will specify its version, configuration and dependencies.
\Needspace{12\baselineskip}
\section*{Authors and contributions}
\noindent\textbf{Core Contributors.} Chao Liang, Ke Chen, Fengxiaoxiao Li, Shibo Huang and Yufeng Sun.

\noindent\textbf{Other Contributors.} Zhaokai Yin, Jianhuan Lai, Haozhe Feng and Chenghuan Jia.

\noindent\textbf{Project Leaders.} Chao Liang and Zhipeng Zhang.

\noindent\textbf{Corresponding Author.} Zhipeng Zhang (\href{mailto:zhipengzhang@sjtu.edu.cn}{zhipengzhang@sjtu.edu.cn}).


\FloatBarrier
\Needspace{8\baselineskip}
\setlength{\bibsep}{3pt plus 1pt minus 1pt}
\bibliographystyle{unsrtnat}
\bibliography{main}
\end{document}